%% file: main.tex
\documentclass{article}
\usepackage[T1]{fontenc}

\PassOptionsToPackage{numbers}{natbib}
\usepackage[eandd,final]{neurips_2026}

\usepackage{graphicx}
\usepackage{caption}
\usepackage{subcaption}
\usepackage{multirow}
\usepackage{soul}
\usepackage{xspace}
\usepackage{xcolor}
\usepackage{placeins}
\usepackage{wrapfig}
\usepackage{hyperref}
\hypersetup{hidelinks}
\usepackage{array}
\usepackage{amsmath}
\usepackage{amssymb}
\usepackage{booktabs}
\usepackage{tikz}
\usetikzlibrary{positioning}
\usepackage{float}
\newtheorem{proposition}{Proposition}
\newtheorem{definition}{Definition}

\newcommand{\stepcircled}[1]{
  \tikz[baseline=(char.base)]{
    \node[shape=circle,draw=black,fill=black,inner sep=1pt] (char) {\textcolor{white}{\textbf{#1}}};
  }
}

\newcolumntype{C}[1]{>{\centering\arraybackslash}m{#1}}

\newcommand*{\diabetes}{\texttt{Diabetes}\xspace}
\newcommand*{\german}{\texttt{German Credit}\xspace}
\newcommand*{\acs}{\texttt{ACS Income}\xspace}
\newcommand*{\gmsc}{\texttt{Give Me Some Credit}\xspace}
\newcommand*{\gmscabbr}{\texttt{GMSC}\xspace}

\title{Explanation Multiplicity in SHAP: Characterization and Assessment}

\newcommand{\codeurl}{\url{https://github.com/Aguno/SHAP_Multiplicity}}
\input{authors}

\begin{document}

\maketitle

\begin{abstract}

\input{abstract}

\end{abstract}

\noindent\textbf{Code and experiment scripts:} \codeurl

\input{introduction}

\FloatBarrier
\input{related}
\FloatBarrier
\input{prelims}
\FloatBarrier
\input{multiplicity}
\FloatBarrier
\input{metrics_short}
\FloatBarrier
\input{experiments_abbrev}
\FloatBarrier
\input{disc_conc}
\input{acknowledgments}

\bibliographystyle{unsrtnat}
\bibliography{main}

\clearpage
\appendix
\input{appendix}
\clearpage

\end{document}

%% file: authors.tex
\author{%
  Hyunseung Hwang \\
  New York University \\
  \href{mailto:hh3884@nyu.edu}{\texttt{hh3884@nyu.edu}} \\
  \And
  Seungeun Lee \\
  New York University \\
  \href{mailto:seungeun.lee@nyu.edu}{\texttt{seungeun.lee@nyu.edu}} \\
  \And
  Lucas Rosenblatt \\
  Williams College \\
  \href{mailto:lr21@williams.edu}{\texttt{lr21@williams.edu}} \\
  \AND
  Steven Euijong Whang \\
  KAIST \\
  \href{mailto:swhang@kaist.ac.kr}{\texttt{swhang@kaist.ac.kr}} \\
  \And
  Julia Stoyanovich \\
  New York University \\
  \href{mailto:stoyanovich@nyu.edu}{\texttt{stoyanovich@nyu.edu}} \\
}

%% file: abstract.tex
SHAP explanations are widely used in high-stakes settings to justify decisions, yet they can differ substantially across repeated runs, even when the model, the input instance, and the prediction are held fixed. Prior work has documented disagreement \emph{between} explanation methods; we show that substantial disagreement arises even \emph{within} SHAP across reruns of the same estimator on the same trained model and instance. We call this phenomenon \emph{explanation multiplicity} and develop an evaluation methodology for characterizing it under deployment-realistic computational budgets, combining a dual-seed protocol that compares model-induced and explainer-induced variability, a hierarchy of magnitude-based, rank-based, and set-based metrics, and randomized Dirichlet and Mallows null models that provide reference scales for observed disagreement. Across multiple datasets, models, and sampling strategies, we find that explanation multiplicity is pervasive and persists even for high-confidence predictions. The relative contribution of each source depends on the data regime and model: model-induced disagreement is generally greater on smaller datasets, while explainer-induced disagreement is greater on larger datasets. Commonly used $\ell_2$ distance can understate this instability, while rank-based metrics reveal substantial changes in top-ranked features, including the leading feature. Improved sampling methods such as CTE do not eliminate rank-level multiplicity, and K-Means reduces run-to-run variation while its compressed-background explanations can diverge from the empirical-distribution reference. Practitioners should treat single-run SHAP outputs as realizations of a distribution rather than as authoritative artifacts.

%% file: introduction.tex
\section{Introduction}
\label{sec:intro}
In lending, healthcare, and public services, explanations of algorithmic decisions are treated as authoritative: they justify outcomes to affected individuals, guide human review, and demonstrate regulatory compliance~\citep{zejnilovic2021machine,chaddad2023survey,Shaikhina2021EffectsOU,fresz2024how}. These explanations are typically consumed as ranked lists of the features that mattered most~\citep{bhatt2020explainable,bennetot2024practical}. Prior work has shown that different explanation methods can disagree substantially when applied to the same model and instance~\citep{agarwal2022openxai,krishna2022disagreement}; in this work, we show that substantial disagreement arises even within a single method across repeated runs.

\begin{figure}[t]
    \centering
    \includegraphics[width=0.60\linewidth]{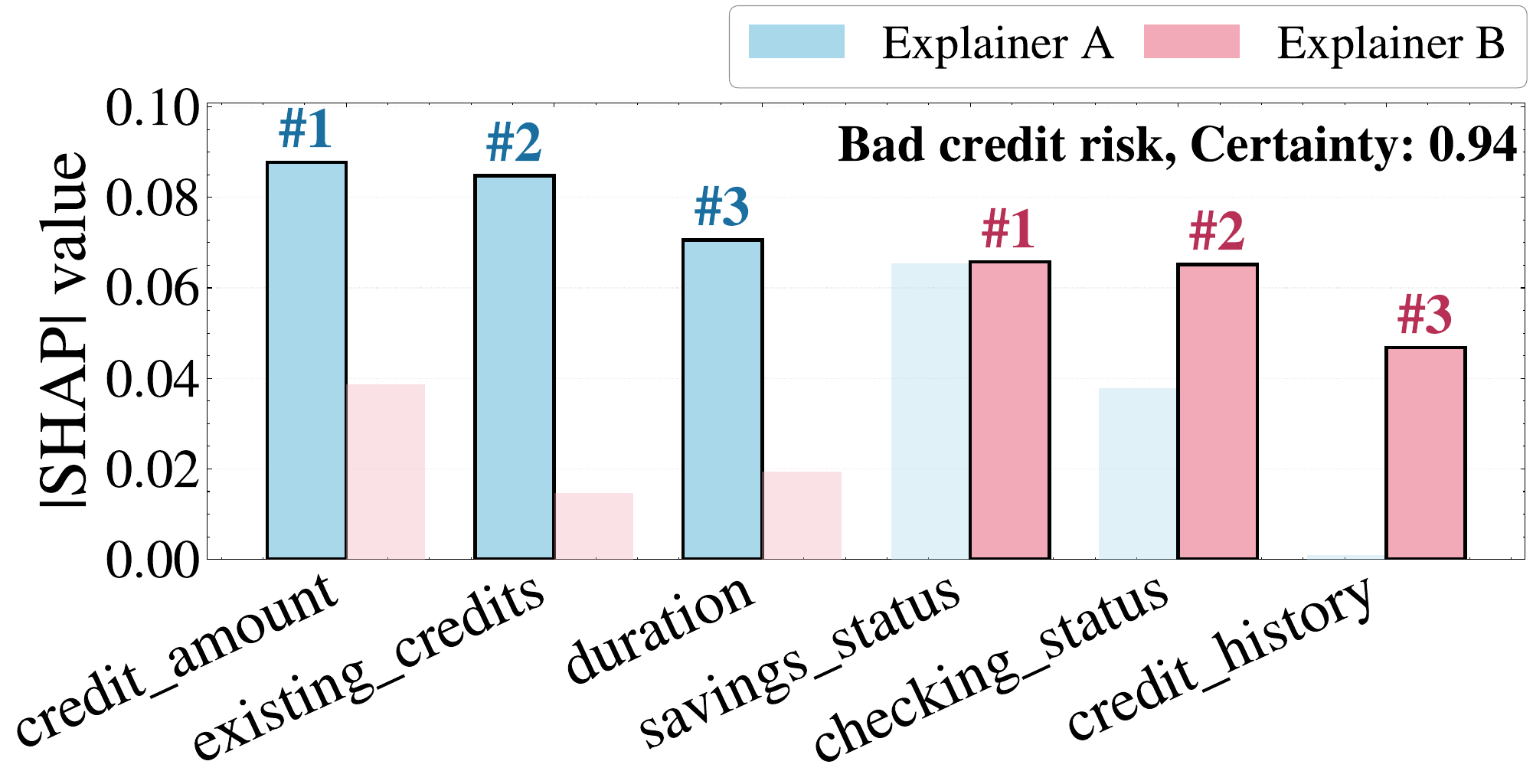}
    \caption{Feature-based explanations for a \textbf{high-confidence adverse credit prediction} vary substantially across reruns of the same explanation pipeline. Top-ranked features differ with no overlap at top-$3$. A and B denote two runs of the same estimator. Example based on lending (\german), with SHAP for an FT-Transformer classifier.}
    \label{fig:motivation1}
\end{figure}

\textbf{Motivating example: explanation disagreement without model disagreement.} A credit-risk classifier denies Alice a loan. Its explanation highlights her credit amount, existing credit lines, and loan duration (Explainer A in Figure~\ref{fig:motivation1}). Rerunning the same pipeline with the same trained model and input instead highlights savings status, checking account status, and credit history (Explainer B). Stochastic sampling changes the dominant factors while the predicted outcome remains unchanged.

 Which explanation should Alice trust, and which factors should she act on when seeking recourse~\citep{fonseca2023setting}? We refer to the existence of multiple, internally valid but substantively different explanations for a single instance as \textbf{explanation multiplicity}, which can arise even when the trained model and its prediction are held fixed. It arises because the SHAP estimators studied here rely on sampling-based approximations of intractable Shapley values~\citep{lundberg2017unified,shapley1953,muscoprovably}, and because different training runs may yield models with comparable accuracy but different internal representations~\citep{black2022model}. In practice these sources are entangled (see Appendix~\ref{app:sources}); Section~\ref{sec:multiplicity} introduces a dual-seed protocol that compares them under separate seed settings. Even when interfaces report standard deviations across runs, these summaries alone do not establish whether explanation variability is substantial. A calibrated baseline is needed to interpret this variation and support explanation-based transparency and accountability~\citep{passi2019problem,black2022model}.

\textbf{From single-run reporting to repeated-run characterization.} Characterizing an explanation distribution requires a feasible computational budget, metrics relevant to downstream use, and a reference for interpreting disagreement. Prior evaluations study magnitude-based robustness and coefficient agreement~\citep{alvarez2018robustness,visani2020statistical}, but practitioners consume ranked feature lists~\citep{krishna2022disagreement}. We show that magnitude-based metrics can suggest stability while masking substantial disagreement in top-ranked features. This paper makes the following contributions:

\stepcircled{1} \textbf{We formalize explanation multiplicity} as a distinct phenomenon, separate from model multiplicity~\citep{black2022model} and inter-method disagreement~\citep{krishna2022disagreement}. Our evaluation methodology includes a dual-seed protocol that compares model-induced and explainer-induced variability under separate seed settings.

\stepcircled{2} \textbf{We develop metrics and calibrated baselines} that surface explanation multiplicity masked by standard evaluation. A hierarchy of magnitude-, rank-, and set-based measures, calibrated against Dirichlet and Mallows null models under a realistic budget of 100 background samples, matching the sample size widely used in the official SHAP tutorials/documentation, provides reference scales for interpreting observed disagreement.

\stepcircled{3} \textbf{We demonstrate empirically} that explanation multiplicity is widespread across multiple datasets, model classes, and sampling strategies, and that it persists even for high-confidence predictions. The relative source contribution depends on the dataset and model: model-induced disagreement is generally greater on smaller datasets, and explainer-induced disagreement on larger datasets. This instability is robust to method-level innovation: state-of-the-art estimators such as CTE can reduce magnitude differences but exhibit rank-level instability comparable to random sampling under our calibrated metrics. K-Means reduces run-to-run variation, but its compressed-background explanations can diverge sharply from the empirical-distribution reference.

%% file: related.tex
\section{Related Work}
\label{sec:relatedwork}

\textbf{Stochasticity in post-hoc explanations.} SHAP~\citep{lundberg2017unified} is sensitive to background size and composition~\citep{yuan2023empiricalstudyeffectbackground,aivodji2023fooling}. \citet{yuan2023empiricalstudyeffectbackground} report improved stability as background size increases from 100 to 1,000 in their study, at greater computational cost. We complement this by characterizing variability at smaller background sizes, with metrics and baselines for interpreting repeated runs. Prior studies connect predictive uncertainty with explanation unreliability~\citep{Shaikhina2021EffectsOU,zhu2025robustexplanationsuncertaintydecomposition}; our confidence-stratified analysis shows that high confidence does not guarantee consistent explanations.

\textbf{Explanation disagreement and improved estimation.} \citet{krishna2022disagreement} introduce the \emph{disagreement problem}, measuring discrepancies between different explanation methods. While their work focuses on \emph{inter-method} disagreement, ours characterizes \emph{intra-method} multiplicity across repeated runs of the same method. On the estimation side, \citet{muscoprovably} propose leverage score sampling and \citet{baniecki2025cte} propose \emph{compress then explain} (CTE) to reduce SHAP approximation error. Our contribution is to characterize the disagreement that persists under standard and improved estimation alike. Explanation variability can also arise from \emph{model multiplicity}~\citep{breiman2001statistical, marx2020predictive}, where retraining yields models with different internal representations~\citep{black2022model}. Our work builds on this by comparing model-induced and explainer-induced variability under a dual-seed protocol.

\textbf{Metrics for explanation consistency.} Prior measures assess selected-feature overlap and coefficient agreement~\citep{visani2020statistical}; Jaccard also measures feature-set stability~\citep{nogueira2017stability}. \citet{slack2021reliable} propose a Bayesian framework for credible intervals, though their evaluation does not address rank disagreement. Large benchmarks like OpenXAI~\citep{agarwal2022openxai} report stability metrics. We complement these evaluations by comparing repeated-run disagreement with reference ranges from specified randomized null models. This comparison shows how metric choice shapes whether explanation multiplicity is surfaced or masked.

%% file: prelims.tex
\section{Preliminaries}
\label{sec:preliminaries}

\textbf{Shapley values and SHAP.} Let $d$ be the number of features, $[d]=\{1,\ldots,d\}$ their index set, $\mathcal{X}\subseteq\mathbb{R}^d$ the input space, $x\in\mathcal{X}$ an input, and $f:\mathcal{X}\to\mathbb{R}$ a trained predictor. For probabilistic classification, $f(x)\in[0,1]$ is the target-class probability. An explainer $\mathcal{E}$ produces a feature-attribution vector $\phi(x)=\mathcal{E}(f,x)\in\mathbb{R}^d$; its coordinate $\phi_m(x)$ is the contribution of feature $m\in[d]$.
The Shapley value~\citep{shapley1953} assigns each feature its marginal contribution averaged over feature coalitions (Appendix~\ref{app:background}). A specified value function yields a unique attribution vector, but exact computation requires $2^{d-1}$ subsets per feature. The model-agnostic SHAP estimators studied here use sampling-based approximations~\citep{lundberg2017unified}.

We focus on \emph{background sampling}: absent features are filled from a finite background dataset $D_{bg}$ of size $K$, using Random sampling or kernel-thinning compression (CTE~\citep{baniecki2025cte}), among other strategies. The induced background distribution $\mathcal{D}_{bg}$ is uniform for row-sampled backgrounds and cluster-weighted for Kernel SHAP K-Means. For a feature coalition $S\subseteq[d]$ and background row $Z\sim\mathcal{D}_{bg}$, the completed input $(x_S,Z_{[d]\setminus S})$ retains $x$ on $S$ and fills the complementary features from $Z$. The coalition value $v_x(S;\mathcal{D}_{bg})$ is
\begin{equation}
v_x(S;\mathcal{D}_{bg}) = \mathbb{E}_{Z\sim\mathcal{D}_{bg}}
\!\left[\, f\!\left(x_S, Z_{[d] \setminus S}\right) \right],
\label{eq:shap_estimate}
\end{equation}
Resampling $D_{bg}$ can change the explanation for fixed $(x,f)$; internal coalition randomness is discussed in Section~\ref{sec:experiments}.

\paragraph{Background distribution as part of the estimand.}
Shapley attributions depend on the coalition value in Eq.~\eqref{eq:shap_estimate}; how to handle ``absent'' features in $[d]\setminus S$ is a modeling decision rather than a purely numerical approximation choice~\citep{janzing2020feature,sundararajan2020many,chen2020true}.
Different background distributions $\mathcal{D}_{bg}$ implicitly define different value functions, and therefore different target Shapley values.
Under this view, sampling-based estimators (e.g., Random, CTE) and
clustering-based summarizers (e.g., K-Means) are not necessarily
approximating the same quantity: the former approximate Shapley values defined with respect to the empirical training distribution $\mathcal{D}_{\text{train}}$,
with value function $v_x(S;\mathcal{D}_{\text{train}})$,
whereas the latter target Shapley values defined with respect to a
compressed background distribution. Throughout this paper, the \emph{empirical-distribution reference} uses the full training background with the same coalition budget; it need not be an exact Shapley computation. We return to this distinction in
Section~\ref{sec:stabilityvsaccuracy} when interpreting the deviation of K-Means from this reference.

%% file: multiplicity.tex
\section{Defining and Assessing Explanation Multiplicity}
\label{sec:multiplicity}

Let $D$ denote a dataset, $\mathcal{T}$ a prediction task, and $\mathcal{M}$ a model class. Together with the explainer family $\mathcal{E}$ and target instance $x$, these specify an \emph{explanation query} $Q = (D, \mathcal{T}, \mathcal{M}, \mathcal{E}, x)$. Executing~$Q$ draws a pipeline realization: training $f \in \mathcal{M}$ and computing $\phi(x) = \mathcal{E}(f,x)$. Because the pipeline may involve stochastic components (random initialization, data shuffling, and background resampling; see Section~\ref{sec:preliminaries}), repeated executions of the same query~$Q$ induce a \emph{distribution} over explanations rather than a single deterministic output.

\begin{definition}[Explanation multiplicity]
\label{def:multiplicity}
For query $Q$, let $\phi,\phi'$ be independent draws from its explanation distribution, $\delta$ a disagreement measure (Section~\ref{sec:metrics}), $\tau$ a threshold specified against the randomized baselines (Section~\ref{sec:preliminaries:baseline}), and $\alpha\in(0,1]$ the required prevalence. We say that \emph{explanation multiplicity} is present at level~$(\tau,\alpha)$ if the pair-exceedance probability satisfies
\begin{equation}
  p_Q(\tau) := \Pr\bigl[\,\delta(\phi, \phi') > \tau \mid Q\,\bigr] \ge \alpha,
  \label{eq:multiplicity_def}
\end{equation}
\end{definition}

Multiplicity depends on the query, measure~$\delta$, threshold~$\tau$, and prevalence level~$\alpha$. Our source-specific analyses estimate the pair-exceedance probability $p_Q(\tau)$ under the seed protocols below and average these estimates across test instances (Eq.~\eqref{eq:pooled_exceedance}). They do not apply $\alpha$ to classify instances as multiplicity-positive. Exceeding a randomized reference is descriptive, not a guarantee of statistical significance or practical harm.

Explanation multiplicity is distinct from two related phenomena, summarized in Table~\ref{tab:concepts}. \emph{Model multiplicity} (the Rashomon effect)~\citep{black2022model,marx2020predictive} concerns multiple models with comparable performance but different instance-level behavior. \emph{Inter-method disagreement}~\citep{krishna2022disagreement} concerns discrepancies between different explanation methods applied to the same model and instance. Explanation multiplicity, by contrast, concerns \emph{intra-method, intra-query} variation: a single method, applied to the same query, produces different outputs across pipeline realizations. Our dual-seed protocol, described below, compares model-induced and explainer-induced variation.

\begin{table}[b]
\centering
\caption{Comparison of explanation multiplicity to related concepts.}
\label{tab:concepts}
\small
\setlength{\tabcolsep}{4pt}
\begin{tabular}{>{\raggedright\arraybackslash}p{.21\linewidth}>{\raggedright\arraybackslash}p{.17\linewidth}>{\raggedright\arraybackslash}p{.17\linewidth}>{\raggedright\arraybackslash}p{.12\linewidth}>{\raggedright\arraybackslash}p{.18\linewidth}}
\toprule
\textbf{Concept} & \textbf{What varies} & \textbf{What is fixed} & \textbf{Level} & \textbf{Reference} \\
\midrule
Model multiplicity & Trained model & Dataset, task & Model class & \citep{black2022model,marx2020predictive} \\
Inter-method disagreement & Method & Model, instance & Explainer & \citep{krishna2022disagreement} \\
Explanation multiplicity & Pipeline run & Query $Q$ & Query & This work \\
\bottomrule
\end{tabular}
\end{table}

To assess explanation multiplicity for a given query~$Q$, we introduce a methodology based on controlled reruns of the same modeling-and-explanation workflow. As discussed in Section~\ref{sec:preliminaries}, randomness may enter at two stages: \emph{model construction} and \emph{explanation} (stochastic approximation of the explainer). Our methodology compares model-induced and explainer-induced variability, clarifying whether practitioners should invest in stabilizing the model or the explainer. Figure~\ref{fig:diagram_pipeline} illustrates the pipeline and the points where multiplicity can arise.

\textbf{Dual-seed protocol.} A \emph{model seed} $s_m$ controls training and model selection, and an \emph{explainer seed} $s_e$ indexes background sampling. We write $\phi(x;s_m,s_e)\in\mathbb{R}^d$ for the resulting explanation. Holding the query fixed while varying these seeds yields repeated explanations. Internal coalition sampling is not necessarily controlled by $s_e$; we qualify this limitation in Section~\ref{sec:experiments}.

\textbf{Multiplicity settings.} We compare \textbf{(i) overall variability}, pooling the available source-specific distance collections; \textbf{(ii) model-induced variability}, varying training seeds with a fixed explainer setting; and \textbf{(iii) explainer-induced variability}, holding the trained model fixed and varying background sampling. The overall collection is not a full factorial experiment: Random/K-Means include both source pools; PredStrat/CTE include only the available background pool. Within~(iii), we compare Random, K-Means, PredStrat, and CTE, evaluating both variability and divergence from a common reference (Section~\ref{sec:stabilityvsaccuracy}).

\input{figures/pipeline}

%% file: figures/pipeline.tex
\begin{figure}
\centering
\begin{tikzpicture}[
    node distance=0.9cm and 0.7cm,
    box/.style={rectangle, draw, rounded corners=2pt, minimum height=0.7cm, 
                minimum width=1.4cm, font=\small, align=center},
    seed/.style={rectangle, draw, dashed, rounded corners=2pt, minimum height=0.7cm,
                 minimum width=1.4cm, font=\small, align=center},
    arr/.style={->, >=stealth, thick},
]
% Main pipeline
\node[box] (train) {Train Set};
\node[box, right=of train, fill=orange!15] (model) {Model};
\node[box, right=of model] (pred) {Prediction};
\node[box, right=of pred, fill=cyan!15] (expl) {Explainer};
\node[box, right=of expl] (out) {Explanation};

% Arrows
\draw[arr] (train) -- (model);
\draw[arr] (model) -- (pred);
\draw[arr] (pred) -- (expl);
\draw[arr] (expl) -- (out);

% Sources of randomness
\node[seed, above=0.6cm of model] (sm) {Model\\[-2pt] Training};
\node[seed, above=0.6cm of expl] (se) {Background\\[-2pt] Sampling};

\draw[arr, dashed] (sm) -- (model);
\draw[arr, dashed] (se) -- (expl);

% Labels
\node[font=\scriptsize\bfseries, below=0.1cm of sm, yshift=0.05cm, xshift=-0.7cm] {(a)};
\node[font=\scriptsize\bfseries, below=0.1cm of se, yshift=0.05cm, xshift=-0.8cm] {(b)};
\end{tikzpicture}
\caption{\textbf{Explanation pipeline.} 
Randomness can enter at (a)~model construction (training and selection) and (b)~background sampling (how absent features are instantiated). Source (b) is specific to the explainer; different estimators handle it differently.}
\label{fig:diagram_pipeline}
\end{figure}

%% file: metrics_short.tex
\section{Metrics and baselines for surfacing explanation multiplicity}
\label{sec:metrics}

Some variation across explanation runs is inevitable. Whether multiplicity is surfaced or masked depends on two choices: the evaluation metric and the reference baseline. We develop both below.

\subsection{Evaluation Metrics}

We evaluate disagreement using a hierarchy of metrics that emphasize different aspects of feature-importance variation. Our choice is informed by practitioner evidence on how explanations are consumed~\citep{krishna2022disagreement}. For a selected depth $k\in\{1,\ldots,d\}$, top-$k$ Jaccard distance relates to their \emph{feature agreement} metric, and rank-biased overlap (RBO) captures a finer-grained version of their \emph{rank agreement}.

\textbf{1. $\ell_2$ Distance.}
For two explanation vectors $\phi,\phi'\in\mathbb{R}^d$ of the same instance,
\begin{equation}
    D_{\ell_2}(\phi,\phi') = ||\phi - \phi'||_2
\end{equation}
While $\ell_2$ captures magnitude of change, it does not directly measure rank changes: a small $D_{\ell_2}$ can still correspond to a different set of top features.

\textbf{2. Top-$k$ Jaccard Distance.}
Let $\pi$ order feature indices by decreasing $|\phi|$, and let $\pi^{1:k}$ be the set of its first $k$ indices; define $\pi'$ analogously for $\phi'$. Tie conventions are given in Appendix~\ref{app:metric_ties}. The top-$k$ Jaccard distance is:
\begin{equation}
D_{\mathrm{Jac}}(\pi,\pi')
=
1 - \frac{|\pi^{1:k} \cap \pi'^{1:k}|}
     {|\pi^{1:k} \cup \pi'^{1:k}|}
\end{equation}
A value of 0 indicates identical top-$k$ sets; 1 indicates disjoint sets. This measures membership churn, not ordering within the set.

\textbf{3. Rank-Biased Overlap (RBO).}
Top-$k$ Jaccard ignores ordering within the top set. RBO~\citep{webber2010similarity} is a top-weighted similarity between ranked lists. For depth $\ell\in\{1,\ldots,d\}$, let
$\mathrm{A}_\ell = |\pi^{1:\ell}\cap \pi'^{1:\ell}|/\ell$ be the prefix overlap. With $p=1-1/d$ setting the prefix weights, we define:
\begin{equation}
S_{\mathrm{RBO}}(\pi,\pi')
=
(1-p)\sum_{\ell=1}^{d} p^{\ell-1}\mathrm{A}_\ell + p^d \mathrm{A}_d,
\end{equation}
At full depth, both prefixes contain $[d]$, so $\mathrm{A}_d=1$ and the final term is $p^d$. We report $1-S_{\mathrm{RBO}}$, which is zero for identical rankings.

\textbf{4. Feature-wise Sensitivity (diagnostic view).}
For $R\geq2$ runs on one instance, let $\phi_m^{(r)}$ be feature $m\in[d]$'s attribution in run $r\in\{1,\ldots,R\}$. Its mean pairwise change is:
\begin{equation}
    S_{feat}^{(m)} = \frac{1}{R(R-1)} \sum_{1\leq r\neq s\leq R} \left| \phi_m^{(r)} - \phi_m^{(s)} \right|
\label{eq:feature_wise_sensitivity}
\end{equation}
This identifies which features drive instability.

\textbf{Aggregation.} For $N$ evaluated instances and $R=5$ runs, write $\phi^{(r)}(x_i)$ for run $r$'s explanation of instance $x_i$. Within each source-specific collection, each instance contributes $\binom{R}{2}=10$ pair distances. Using the indicator $\mathbf{1}\{\cdot\}$, we report the percentage of these $10N$ run-pair--instance comparisons exceeding $\tau$:
\begin{equation}
 B_{\delta}(\tau)=\frac{100}{10N}\sum_{i=1}^{N}\sum_{1\leq r<s\leq R}
 \mathbf{1}\{\delta(\phi^{(r)}(x_i),\phi^{(s)}(x_i))>\tau\}.
 \label{eq:pooled_exceedance}
\end{equation}
Let $\widehat p_i(\tau)$ be instance $i$'s fraction of pairs exceeding $\tau$, estimating its query's probability in Definition~\ref{def:multiplicity}. Equivalently, $B_\delta(\tau)=\frac{100}{N}\sum_i\widehat p_i(\tau)$. We average exceedance indicators, not distances, and do not threshold $\widehat p_i$ at $\alpha$. Distribution plots pool unaveraged pair distances unless stated otherwise. The ten pairs share runs and are not independent observations.

\subsection{Calibration via Randomized Baselines}
\label{sec:preliminaries:baseline}

We introduce randomized null models that provide calibrated reference ranges for each metric.  For magnitude-based disagreement, a Dirichlet model allocates a common nonnegative mass $T=0.4$ across features. It supplies a reference scale for signed attribution distances, rather than matching each instance's mass.

\begin{proposition}[Expected squared $\ell_2$ distance]
\label{prop:l2_baseline}
Let $1\leq r<d$ be the dominant-group size, $\rho\in(0,1)$ its mass fraction, and $\kappa>0$ the concentration in the Dirichlet model of Appendix~\ref{app:proof_l2}. For i.i.d.\ mass-$T$ vectors $X,Y$ from this model,
\begin{equation}
    \mathbb{E}\,\|X - Y\|_2^2
    =
    \frac{2T^2}{\kappa + 1}
    \left(
    1 - \frac{\rho^2}{r} - \frac{(1-\rho)^2}{d-r}
    \right).
    \label{eq:l2_baseline_final}
\end{equation}
We compare $\ell_2$ distances to the square root of this expectation, an RMS reference. Proof in Appendix~\ref{app:proof_l2}.
\end{proposition}

For rank-based metrics, a Mallows model under Kendall-tau distance~\citep{mallows1957non} perturbs a shared ordering with dispersion $q$. Monte Carlo mean distances over the parameter sweep form the baseline band (Appendix~\ref{app:mallows_def}); its minimum defines $\tau$. These bands are parameter ranges, not confidence intervals.

%% file: experiments_abbrev.tex
\section{Empirical Evaluation}
\label{sec:eval}

We empirically investigate explanation multiplicity in SHAP guided by four key questions.
\textbf{Source of multiplicity:} When explanations disagree, is this primarily driven by the trained model or by stochasticity in the explainer under varying background samples? (Section~\ref{subsec:source_dissection})
\textbf{The illusion of stability:} Under which evaluation metrics is explanation multiplicity surfaced versus masked? (Section~\ref{sec:illusion_of_stability})
\textbf{Stability vs.\ reference agreement:} Do stable sampling methods also agree with the empirical-distribution reference? (Section~\ref{sec:stabilityvsaccuracy})
\textbf{Role of confidence:} Is high prediction confidence sufficient to guarantee stable explanations? (Section~\ref{subsec:certainty})

\subsection{Experimental Setup}
\label{sec:experiments}

\textbf{Datasets.}
We evaluate tabular datasets spanning different scales:
\acs (Virginia, 2018)~\citep{ding2021retiring} ($|D|{=}45{,}960, d{=}8$),
\german~\citep{german_credit_1994} ($|D|{=}988, d{=}16$),
\gmsc (\gmscabbr)~\citep{kaggle2011givemesomecredit} ($|D|=45,000$, $d{=}10$), \diabetes~\citep{smith1988using} ($|D|{=}392, d{=}8$), with the corresponding classification tasks. See Table~\ref{tab:datasets} for details. We use stratified five-fold partitioning; the reported figures analyze the first held-out fold of size $N$.

\textbf{Models.}
To avoid model-specific conclusions, we consider diverse predictors:
tree-based (Decision Tree (DT), Random Forest (RF)~\citep{breiman2001random}, XGBoost (XGB)~\citep{chen2016xgboost}), neural (MLP, FT-Transformer (FTT)~\citep{gorishniy2021revisiting}).
Models are tuned via grid search within the training fold (Appendix~\ref{app:hyperparameters}).

\textbf{Evaluation protocol.} We use $K=100$ background instances and five background seeds, comparing Kernel SHAP and Permutation SHAP with Random, K-Means, PredStrat, and CTE. A single GMSC configuration (9,000 test instances, $K=100$, 1,024 coalitions) took approximately 24 hours on one NVIDIA L40S, including model queries; five runs incur five such computations. Kernel SHAP uses up to $\min(2^d,1024)$ coalitions; Permutation SHAP uses the same numerical \texttt{max\_evals} budget, with different evaluation units. Neither budget guarantees internal-seed control: the Kernel SHAP coalition seed is ineffective in the subsampled setting. We therefore do not attribute all explainer variability to background sampling alone. The \emph{empirical-distribution reference} uses the full training background at the same coalition budget; for German Credit it is itself approximate.

\subsection{Identifying the Source of Explanation Multiplicity}
\label{subsec:source_dissection}

\begin{figure}
    \centering
    \begin{subfigure}[b]{0.95\textwidth}
        \centering
        \includegraphics[width=\textwidth]{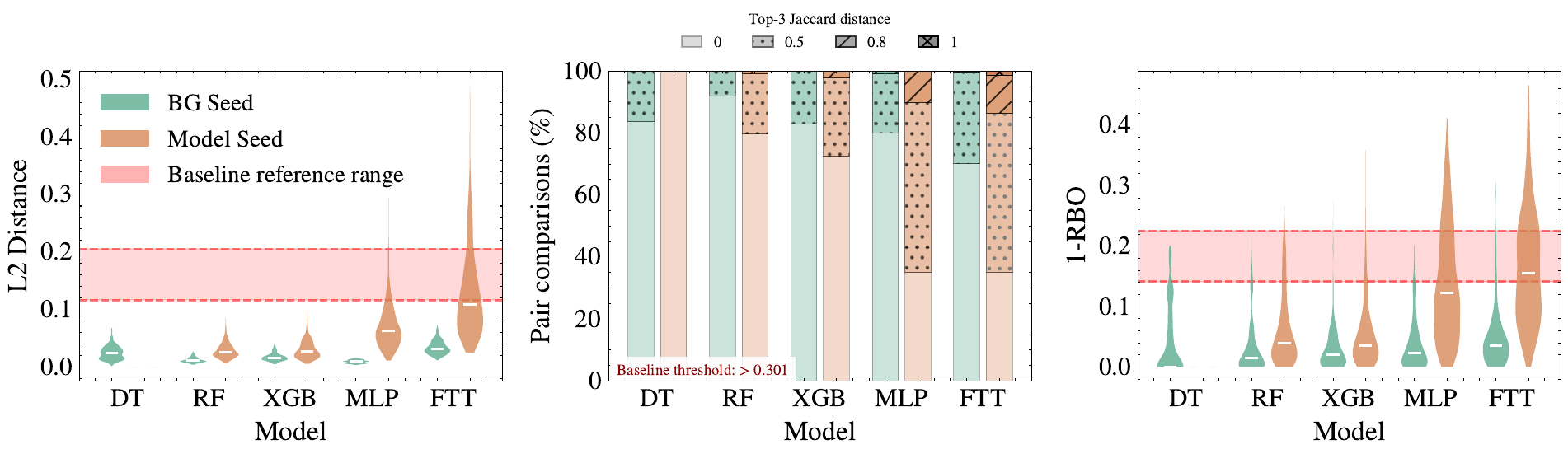}
        \caption{\diabetes}
        \label{fig:diabetes_violin_main}
    \end{subfigure}
    \hfill
    \begin{subfigure}[b]{0.95\textwidth}
        \centering
        \includegraphics[width=\textwidth]{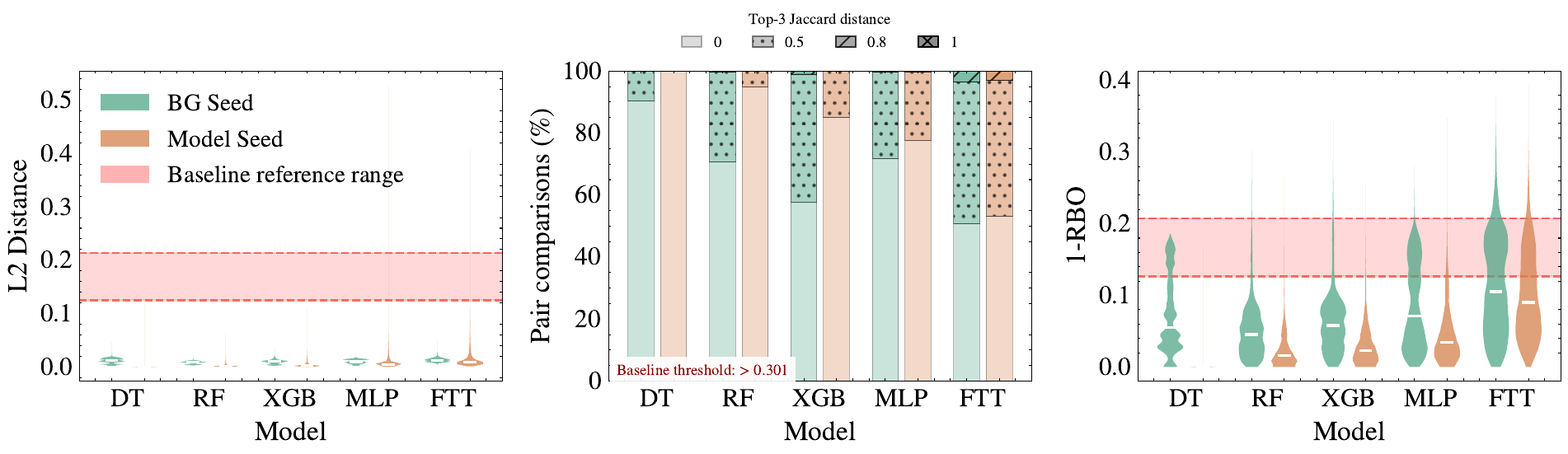}
        \caption{\gmsc}
        \label{fig:gmsc_violin_main}
    \end{subfigure}
    \hfill
    \centering

\caption{\textbf{Model vs.\ explainer multiplicity (dissection).}
Kernel SHAP with Random background sampling. Pair-distance distributions by source ($\ell_2$, Top-3 Jaccard, 1-RBO); Jaccard uses four-value frequency bars. Green: explainer seed varied with model fixed; orange: model seed varied with explainer fixed. Shaded bands denote randomized baseline ranges. Model-induced disagreement is generally greater on smaller datasets, with model-dependent exceptions; explainer-induced disagreement is greater on larger datasets. Full results appear in Figure~\ref{fig:seed_sensitivity}.}
    \label{fig:seed_sensitivity_small}
    \vspace{-0.5cm}
\end{figure}

To identify the source of explanation multiplicity, we compare two sources: \textit{model-induced} and \textit{explainer-induced}.
Using the dual-seed protocol, we measure \textit{model-induced multiplicity} by varying the model random seed while fixing the explainer, and \textit{explainer-induced multiplicity} by varying the explainer seed while fixing the trained model.
Figure~\ref{fig:seed_sensitivity_small} shows this comparison for \diabetes and \gmscabbr, see Figure~\ref{fig:seed_sensitivity} for complete details.

For smaller datasets such as \german and \diabetes, model-induced disagreement is generally greater, with model-dependent exceptions. This is consistent with greater variability in learned models when training data is limited. On the larger \acs and \gmscabbr datasets, explainer-induced disagreement is greater in these comparisons; the protocol does not isolate all internal sampling effects. Neural models often exhibit greater variability, but this association depends on dataset and source of randomness and does not establish a causal effect of parameter count.

\subsection{The Illusion of Stability under Metric Divergence}
\label{sec:illusion_of_stability}

We report the percentage of run-pair--instance comparisons exceeding $\tau$, the minimum randomized baseline (Eq.~\eqref{eq:pooled_exceedance}). In the displayed matrices, $\ell_2$ exceedance rounds to $0\%$ across datasets and models. This does not establish that the explanations are identical or robust.
However, rank-based metrics reveal a different picture for the same comparisons. Under Random sampling, the proportion of run-pair--instance comparisons exceeding the Top-3 Jaccard baseline reaches 28\% for \german, 30\% for \diabetes, 24\% for \acs, and 49\% for \gmscabbr, all on FT-Transformer (Figure~\ref{fig:heatmap_full_kernel}). Across Random, CTE, and PredStrat, $\ell_2$ exceedance rounds to $0\%$, while Top-3 Jaccard exceedance spans 10--65\% of these comparisons for \gmsc (Figure~\ref{fig:bg_heatmap}; other datasets in Appendix~\ref{app:exp}). Thus, rank-level disagreement persists under improved background sampling.

This divergence is consistent with the underlying \emph{feature-importance landscape} (Figure~\ref{fig:feature_wise} in Appendix~\ref{app:exp}): modest attribution changes can alter membership and ordering when feature magnitudes are close. The figure summarizes mean absolute attributions and pairwise changes for the six most sensitive features, pooled across instances and sampling methods. It provides an aggregate view of this pattern, rather than measuring adjacent importance gaps within individual explanations or establishing a causal effect of class imbalance.

Top-3 membership churn can arise from exchanges near the selection boundary without changing the leading feature. We therefore examine how far disagreement extends into the ranking. For FT-Transformer with Kernel SHAP at $K=100$, we select a consensus leading feature for each instance using its mean absolute attribution across five runs. Under Random and CTE, this feature retains rank~1 in approximately 96--97\% of the $5N$ run--instance observations on \german and \diabetes, compared with 70.2\% under Random and 80.3\% under CTE on \gmscabbr. On \gmscabbr, the leading feature also differs in 42.9\% and 32.0\% of the $10N$ run-pair--instance comparisons under Random and CTE, respectively. Thus, the observed instability extends to the leading feature in this setting and cannot be explained solely by rank-3/rank-4 exchanges. Appendix~\ref{app:revision_evidence} reports the full rank-position distributions (Figure~\ref{fig:appendix_churn_depth}) and churn summaries (Table~\ref{tab:top1_churn}).

RBO also detects disagreement (23--42\% pair--instance exceedance for GMSC DT/MLP/FTT under Random; Figure~\ref{fig:heatmap_full_kernel}(c)). Additional $k$- and $K$-sensitivity results appear in Appendix~\ref{app:revision_evidence}.

\begin{figure}
    \centering
    \begin{subfigure}[b]{0.49\textwidth}
        \centering
        \includegraphics[width=\textwidth]{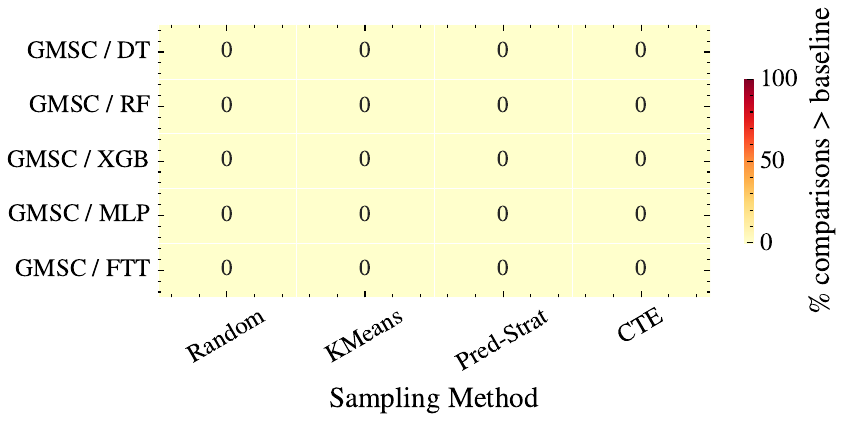}
        \caption{$\ell_2$ Distance}
        \label{fig:heatmap_l2}
    \end{subfigure}\hfill
    \begin{subfigure}[b]{0.49\textwidth}
        \centering
        \includegraphics[width=\textwidth]{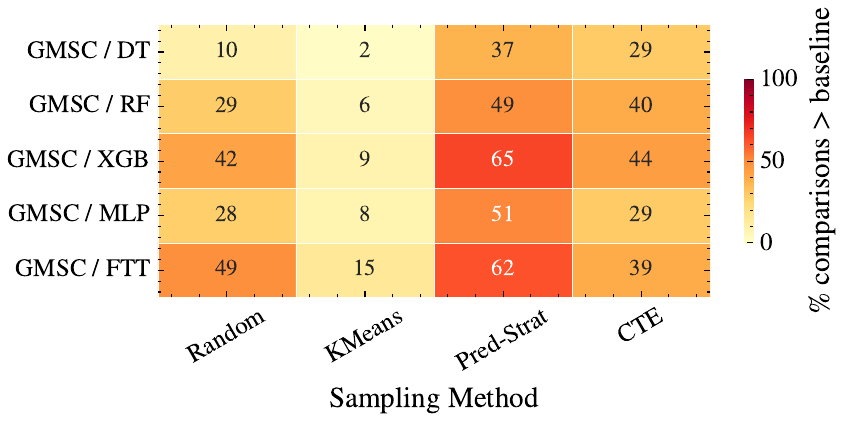}
        \caption{Top-3 Jaccard Distance}
        \label{fig:heatmap_jaccard}
    \end{subfigure}
    \hfill
    \par\medskip
    \begin{subfigure}[b]{0.49\textwidth}
        \centering
        \includegraphics[width=\textwidth]{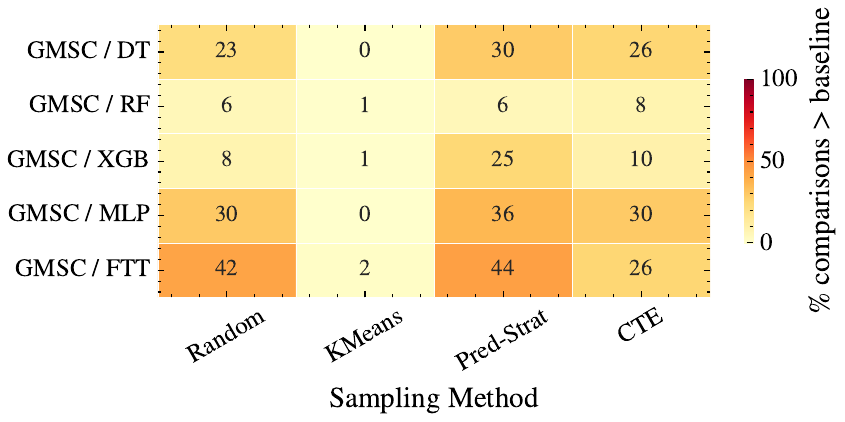}
        \caption{$1-$RBO}
        \label{fig:heatmap_rbo_main}
    \end{subfigure}
\caption{\textbf{Metric divergence for Kernel SHAP on \gmscabbr.} Percentage of run-pair--instance comparisons exceeding the randomized baseline, by method and model (Eq.~\eqref{eq:pooled_exceedance}). Each cell uses $10N$ comparisons. Low $\ell_2$ exceedance coexists with Top-3 membership churn and RBO disagreement.}
    \label{fig:bg_heatmap}
    \vspace{-0.5cm}
\end{figure}

\subsection{Stability vs. Reference Agreement}
\label{sec:stabilityvsaccuracy}

A natural objection is to use a stable summarizer such as K-Means. However, agreement across reruns does not imply agreement with the empirical-distribution reference: a compressed background changes the value function.

K-Means greatly reduces run-to-run variability, but stability should not be conflated with reference agreement. Our reference comparisons cover Diabetes and GMSC (Figure~\ref{fig:heatmap_true_distance}(a)). K-Means produces larger reference distances than Random and CTE in many of these configurations. On \gmsc, K-Means yields mean $\ell_2$ distances of $0.026$--$0.051$ across model classes, roughly $3$--$8\times$ higher than Random ($0.006$--$0.010$) or CTE ($0.006$--$0.009$). These mean $\ell_2$ distances remain below the randomized baseline, yet $78\%$ to $90\%$ of the $5N$ seed--instance comparisons exceed the Top-3 Jaccard baseline (Figure~\ref{fig:heatmap_true_baseline}(b)). On Decision Tree, the corresponding percentages are $89\%$ for K-Means, $6\%$ for Random, and $16\%$ for CTE.

\paragraph{Bias or different estimand?}
A compressed background defines a different target~\citep{janzing2020feature,chen2020true}. Our comparisons measure divergence from a same-budget empirical-distribution reference, not error against universally exact Shapley values. For reference comparisons, each of five explanations is compared with one full-background explanation; thresholds are applied before pooling the $5N$ distances. Some neural sampled and reference runs differ in reconstructed model output, so their distances cannot be attributed solely to the background (Appendix~\ref{app:reference_caveat}).

\subsection{Confidence Does Not Guarantee Stability}
\label{subsec:certainty}
Finally, we examine whether prediction confidence provides a reliable signal of explanation multiplicity. We stratify instances by predicted probability $P=f(x)$ into \emph{Certain} ($P\ge0.9$ or $P\le0.1$) and \emph{Uncertain} ($0.4\le P\le 0.6$). With the trained model fixed, a group of $N_g$ instances contributes all $10N_g$ run-pair distances, without averaging within instances.
Intuitively, one might expect explanations for high-confidence predictions to be more stable than those near the decision boundary. Figure~\ref{fig:certainty} partially supports this intuition: uncertain predictions generally exhibit greater distances. However, high-confidence predictions also produce rank distances approaching or exceeding the randomized baseline, and on \texttt{GMSC} their distances can surpass those of uncertain predictions for several model classes.
This pattern is consistent with the flat-landscape explanation above: reduced attribution magnitudes can leave small adjacent gaps, so modest fluctuations change feature ranks even at high prediction confidence.

%% file: disc_conc.tex
\section{Conclusions, Limitations, and Future Work}
\label{sec:discussion}

This work highlights a broader implication for responsible AI practice: \emph{the evaluation of explanations is not a neutral technical exercise, but a normative choice that shapes what kinds of failures become visible}. Across datasets, models, and evaluation lenses, we find that explanation multiplicity is often present even when conventional stability metrics suggest robustness.

\textbf{From point estimates to distributional reporting.}
Our findings argue against reporting a single SHAP run as if it were \emph{the} explanation. Under realistic computational budgets, repeated runs exhibit variability large enough to reorder top-ranked features in a substantial fraction of run-pair--instance comparisons, even when magnitude-based metrics suggest stability. We advocate distributional reporting: computing explanations across multiple pipeline realizations, summarizing the resulting distribution using rank-aware metrics, and calibrating observed variability against principled null models. Table~\ref{tab:instance187_report} illustrates this reporting. While our empirical analysis focuses on SHAP, the broader lesson is method-agnostic: many post-hoc explanation techniques rely on approximation or sampling, and the same case for distributional reporting applies wherever such variability is present.

\textbf{Limitations.} Our main analysis examines $K=100$ in depth, complemented by a background-size sweep up to $K=1600$ in Appendix~\ref{app:revision_evidence}. These experiments characterize explanation multiplicity under realistic but not exhaustive sampling regimes. The randomized baselines rely on specified distributional assumptions and parameter choices rather than ground-truth generative models; we view them as pragmatic yardsticks that improve upon interpreting stability metrics in isolation. The framework is modular and could incorporate richer distributional assumptions in future work.

\textbf{Future work.} Extending this evaluation framework to explanation methods beyond SHAP, exploring adaptive background sampling that mitigates explainer-induced multiplicity while preserving agreement with the intended reference, investigating how explanation variability should be communicated to non-expert stakeholders, and integrating calibrated explanation evaluation into deployment and governance workflows are all important directions for future work.

%% file: acknowledgments.tex
\section*{Acknowledgments}
This work was supported in part by the NYU-KAIST Partnership and by the Institute of Information \& Communications Technology Planning \& Evaluation (IITP) with a grant funded by the Ministry of Science and ICT (MSIT) of the Republic of Korea (No. RS-2024-00469482, RS-2024-00509258, and RS-2022-II220157). This work was also supported in part by US National Science Foundation (NSF) Awards No. 2312930, 2326193, and NSF GRFP DGE-2234660, and by the National Research Foundation of Korea (NRF) grant funded by the Korea government (MSIT) (No.\@ RS-2022-NR070121). This work was supported in part through the NYU IT High Performance Computing resources, services, and staff expertise.

%% file: appendix.tex
\appendix
\section*{Appendix Roadmap}
This appendix is organized as follows. Appendix~\ref{app:background} formalizes Shapley values and the sources of stochasticity in SHAP. Appendix~\ref{app:proof_l2} provides the proof of Proposition~\ref{prop:l2_baseline} and details the Mallows-based ranking baseline. Appendix~\ref{app:pipeline} describes our feature-level explanation interface. Appendix~\ref{app:exp_settings} lists dataset statistics, baseline hyperparameter sweeps, background sampling strategies, and model tuning grids. Appendix~\ref{app:exp} reports extended empirical results, including (i) Kernel vs.\ Permutation SHAP comparisons, (ii) the full source-decomposition across datasets, (iii) feature-wise sensitivity, (iv) sampling-method comparisons, (v) divergence from the empirical-distribution reference, and (vi) confidence stratification. Appendix~\ref{app:revision_evidence} adds churn-depth, background-size, conditional Shapley, and high-dimensional dataset results.

\paragraph{Notation.}
\begin{center}
\begin{tabular}{ll}
\toprule
Symbol & Meaning \\
\midrule
$d$, $[d]$ & Feature count and index set \\
$x$, $f$, $S$ & Input instance, predictor, and feature coalition $S\subseteq[d]$ \\
$\mathcal{E}$, $\phi(x)$, $\phi_m(x)$ & Explainer, attribution vector, and feature $m$'s contribution \\
$v_x(S;\mathcal{D}_{bg})$ & Coalition value for input $x$ and background distribution \\
$N$, $R$ & Evaluated test instances and explanation runs ($R=5$) \\
$\phi^{(r)}(x_i)$, $\phi_m^{(r)}$ & Run $r$'s vector at $x_i$; feature $m$'s value at fixed input \\
$D_{bg}$, $\mathcal{D}_{bg}$ & Sampled background dataset and background distribution \\
$K$, $k$, $r$ & Background size, Jaccard depth, Dirichlet dominant-group size \\
$\delta$, $\tau$, $\alpha$ & Disagreement measure, reference threshold, probability level \\
$p_Q(\tau)$, $B_\delta(\tau)$ & Query exceedance probability and pooled percentage \\
$\pi$, $\pi^{1:k}$ & Feature ranking and set of its first $k$ feature indices \\
$\mathrm{A}_\ell$, $p$ & Prefix overlap at depth $\ell$ and RBO persistence parameter \\
$T$, $\rho$, $\kappa$, $q$ & Null mass, mass fraction, Dirichlet concentration, Mallows dispersion \\
\bottomrule
\end{tabular}
\end{center}

\section{Background on Feature-Attribution Methods}
\label{app:background}
\noindent This appendix complements Section~\ref{sec:preliminaries} by formalizing Shapley values, detailing how SHAP approximates them via background sampling, and elaborating on the entangled sources of stochasticity---coalition sampling, background completion, and model training---that jointly give rise to explanation multiplicity.

\subsection{SHAP and Shapley Values}
\label{app:background:shap}

For a fixed predictor $f$, input $x$, and background distribution $\mathcal{D}_{bg}$, write $v_x(S)=v_x(S;\mathcal{D}_{bg})$ for the coalition value in Eq.~\eqref{eq:shap_estimate}. The Shapley value~\citep{shapley1953} of feature $i\in[d]$ is
\begin{equation}
    \phi_i(x;f,\mathcal{D}_{bg}) = \sum_{S \subseteq [d] \setminus \{i\}}
    \frac{|S|!(d-|S|-1)!}{d!}
    \left( v_x(S \cup \{i\}) - v_x(S) \right).
\label{eq:shapley_value_def}
\end{equation}

The magnitude of a Shapley value reflects how much a feature influences the prediction, while its sign indicates the direction of that influence. High Shapley values assigned to protected attributes such as age or race indicate that these features play a substantial role in the decision, which may raise ethical or legal concerns. We refer to the assignment of Shapley values (equivalently, ``weights'' or ``contributions'') to a sample's features as an \emph{explanation}.

In principle, Shapley values are uniquely defined for a given model, input instance, and specified value function. The Shapley value of a feature is obtained by averaging its marginal contribution to the model output over all possible subsets of the remaining features, weighted by subset size, as formalized in~\eqref{eq:shapley_value_def}. This definition yields a single, well-defined attribution vector over features that satisfies desirable axioms such as efficiency, symmetry, linearity, and the null-player
property~\citep{shapley1953,lundberg2017unified}.

\paragraph{Background sampling.} Practical explanation methods like SHAP~\citep{lundberg2017unified}
approximate true Shapley values using \emph{background sampling}. They estimate marginal contributions with respect to a background distribution, so changing the background dataset changes the distribution being approximated.

To illustrate this mechanism, consider three features: \textit{payment\_history} (\textit{ph}), \textit{savings\_amount} (\textit{sa}), and \textit{loan\_size} (\textit{ls}). Let $x$ describe
an applicant with $\textit{ph}=\texttt{good}$, $\textit{sa}=\texttt{low}$, and $\textit{ls}=\texttt{high}$. When estimating the contribution of \textit{ph}, SHAP compares the model's
prediction with \textit{ph} present to predictions in which \textit{ph} is treated as absent. Since the model requires all features to be specified, absent values are filled in by drawing from a background dataset.

Concretely, for a coalition $S \subseteq [d] \setminus \{i\}$, SHAP estimates
\[
v_x(S;\mathcal{D}_{bg}) = \mathbb{E}_{Z\sim\mathcal{D}_{bg}}
\!\left[\, f\!\left(x_S, Z_{[d] \setminus S}\right) \right],
\]
where $x_S$ supplies values from $x$ and $Z_{[d]\setminus S}$ supplies the remaining values from a background row $Z$. A finite dataset $D_{bg}$ defines an empirical distribution $\mathcal{D}_{bg}$, uniform for row-sampled backgrounds and cluster-weighted for Kernel SHAP K-Means. Repeated runs with different $D_{bg}$ can therefore produce different explanations for the same $(x,f)$.

\subsection{Sources of Explanation Multiplicity}
\label{app:sources}

At a high level, explanation multiplicity arises because widely used post-hoc explanation methods are implemented as \emph{stochastic approximation pipelines} rather than deterministic mappings from a model and an input to a unique explanation. While Shapley values are uniquely defined in theory~\citep{shapley1953}, computing them exactly is intractable for all but the simplest models, as it requires evaluating the model on all possible coalitions of input features. As a result, practical implementations rely on sampling-based approximations that introduce randomness into the explanation process.

In particular, popular estimators such as Kernel SHAP~\citep{lundberg2017unified} and Leverage SHAP~\citep{muscoprovably} estimate feature attributions through sampling-based approximations. Even when the trained model and the input instance are held fixed, different random samples of background data or different internal random seeds can yield different attribution vectors. Consequently, a feature-attribution explanation should be understood not as a single deterministic object, but as a realization of a random variable induced by the explainer's approximation procedure.

Explanation multiplicity can also arise from stochasticity elsewhere in the ML pipeline. In many practical settings, models are trained through procedures that involve randomness, such as random initialization, data shuffling, or hyperparameter selection. Different training runs may yield models with comparable predictive performance but different internal representations and decision boundaries, a phenomenon often described as \emph{model multiplicity}~\citep{black2022model}. When explanations are generated for such models, differences in feature attributions may reflect genuine differences in model behavior rather than noise introduced by the explainer.

In practice, these sources of variability are typically entangled. Standard stability analyses often rerun the full explanation pipeline and report aggregate dispersion statistics, without distinguishing whether observed explanation disagreement is driven by differences between trained models or by stochasticity intrinsic to the explainer itself. As a result, explanation multiplicity can be observed even under highly controlled conditions, including settings where the trained model is held fixed and the prediction remains unchanged across runs. This entanglement makes it difficult to attribute observed explanation disagreement to a specific component of the pipeline, motivating the dual-seed protocol introduced in Section~\ref{sec:multiplicity}.

\section{Proof of Proposition~\ref{prop:l2_baseline}}
\label{app:proof_l2}
\noindent This appendix provides the full proof of
Proposition~\ref{prop:l2_baseline} on the expected squared
$\ell_2$ distance under the Dirichlet null model, and specifies
the Mallows-based ranking null model
(Section~\ref{sec:preliminaries:baseline}) used to calibrate
rank-based metrics via Monte Carlo estimation.

These metrics and baselines complement the separate model-seed and explainer-seed protocols introduced in Section~\ref{sec:multiplicity} by providing reference scales for interpreting disagreement in the source-specific comparisons, and are used throughout the empirical evaluation in Section~\ref{sec:eval}.

\paragraph{Metric hierarchy.} We evaluate disagreement using a hierarchy of metrics that ranges from magnitude-based ($\ell_2$) to rank-based (top-$k$ Jaccard, RBO) and finally to feature-level granularity (feature-wise sensitivity). Each metric captures a distinct notion of disagreement, and as we show empirically, some lenses systematically mask explanation multiplicity while others reveal it. Our metric hierarchy extends the practitioner-oriented perspective of~\citep{krishna2022disagreement} by adding randomized reference scales for interpreting observed disagreement.

\paragraph{Interpretation of the baselines.}
We treat these randomized null models not as ground truth, but
as interpretive yardsticks against which observed explanation
disagreement can be evaluated. Without a reference point,
isolated metric values---e.g., a top-$k$ Jaccard distance of
$0.4$---are difficult to interpret: they may reflect substantial
multiplicity or merely match what random variation alone would
produce. The square root of the expected squared $\ell_2$ distance derived below provides an RMS reference scale for magnitude-based disagreement under random
allocation of attribution mass, while the Mallows baseline
(Appendix~\ref{app:mallows_def}) provides the analogous reference
for rank-based metrics.

\paragraph{Stakeholder relevance of top-$k$ metrics.} Practitioner evidence emphasizes agreement among the highest-ranked features~\citep{krishna2022disagreement}, motivating top-$k$ Jaccard distance and 1-RBO as complements to magnitude-based measures.

\paragraph{Why $\ell_2$ alone is insufficient.} While $\ell_2$ captures the magnitude of change between explanation vectors, it does not directly measure rank changes: a small $\ell_2$ distance can still correspond to a different set of top features. Thus, $\ell_2$ distance can mask explanation multiplicity that is salient for stakeholders who rely on ranked feature importance rather than absolute attribution magnitudes.

\paragraph{Role of feature-wise sensitivity.} Beyond aggregate metrics, the feature-wise sensitivity $S_{feat}^{(m)}$ defined in Eq.~\eqref{eq:feature_wise_sensitivity} identifies which features drive instability by quantifying their fluctuation magnitude across runs. This feature-wise view helps explain why aggregate metrics may disagree by revealing which features drive explanation multiplicity and where small fluctuations can reorder rankings.

\paragraph{Choice of the Dirichlet null.}
Given two explanation vectors $\phi, \phi' \in \mathbb{R}^d$,
fix a nonnegative magnitude scale $T>0$. It can represent total attribution mass; our comparisons use a common $T=0.4$. We model a randomized
attribution magnitude vector as $X := TM$, where
$M \sim \mathrm{Dirichlet}(\alpha)$ is drawn over the probability
simplex $\Delta^{d-1}$, and define $Y := TM'$ analogously. The
Dirichlet distribution allocates a fixed total mass across
nonnegative components under a unit-sum constraint, making it a
natural random model for distributing attribution mass across
feature dimensions while preserving the overall scale of the
explanation vector.

\paragraph{Proof.}

To define a baseline structure, we fix an index set $S \subset [d]=\{1, 2, \dots, d\}$ with $1\leq |S|=r<d$ containing top-$r$ features. Let $\rho \in (0,1)$ denote the fraction of total mass assigned to the top-$r$ features, i.e., the cumulative importance carried by indices in $S$.
We define a baseline mean composition $m \in \Delta^{d-1}$ by allocating a total mass of $\rho$ to the top-$r$ features and $1-\rho$ to the remaining $d-r$ features, and distributing the mass uniformly within each group.
Formally, the baseline mean vector $m$ is defined component-wise as:
\begin{equation}
m_i =
\begin{cases}
\rho / r, & i \in S, \\
(1-\rho)/(d-r), & i \notin S .
\end{cases}
\label{eq:l2_baseline_mi}
\end{equation}

We model $M$ as a $\mathrm{Dirichlet}$ random vector,
\[
M \sim \mathrm{Dirichlet}(\alpha),
\qquad
\alpha_i = \kappa m_i >0 \quad \forall i,
\]
where $\kappa > 0$ is a concentration parameter.
Smaller values of $\kappa$ induce heavier-tailed and sparser realizations, while larger values concentrate mass more tightly around the mean $m$. We adopt a Dirichlet distribution to model the allocation of nonnegative attribution mass for the reference scale. Observed $\ell_2$ distances are computed from raw signed SHAP vectors, without normalization or an absolute-value transformation.

Since $X$ and $Y$ are i.i.d.,
\[\mathbb{E}\,\|X - Y\|_2^2
= 2\,\mathrm{tr}(\mathrm{Cov}(X)).\]
We have
\[\mathrm{Cov}(X) = T^2 \mathrm{Cov}(M)\] as $X = TM$
, and thus \[\mathbb{E}\,\|X - Y\|_2^2
= 2T^2\,\mathrm{tr}(\mathrm{Cov}(M)).\]

For a $\mathrm{Dirichlet}(\alpha)$ distribution with $\alpha_0 = \sum_i \alpha_i = \kappa$ and $m_i = \alpha_i/\alpha_0$,
\begin{align}
\mathrm{Var}(M_i) &= \frac{m_i(1-m_i)}{\kappa + 1}, \\
\mathrm{Cov}(M_i,M_j) &= -\frac{m_i m_j}{\kappa + 1}, \quad i \neq j .
\end{align}
Therefore,
\[
\mathrm{tr}(\mathrm{Cov}(M))
= \sum_{i=1}^d \mathrm{Var}(M_i)
= \frac{1 - \sum_{i=1}^d m_i^2}{\kappa + 1}.
\]

From equation~\eqref{eq:l2_baseline_mi},
\[
\sum_{i=1}^d m_i^2
=
r\left(\frac{\rho}{r}\right)^2
+
(d-r)\left(\frac{1-\rho}{d-r}\right)^2
=
\frac{\rho^2}{r}
+
\frac{(1-\rho)^2}{d-r}.
\]

With $\alpha_0 = \kappa$, we obtain
\[
\mathrm{tr}(\mathrm{Cov}(M))
=
\frac{1}{\kappa + 1}
\left(
1 - \frac{\rho^2}{r} - \frac{(1-\rho)^2}{d-r}
\right).
\]
Multiplying by $2T^2$ completes the proof.
\hfill$\square$

It is noteworthy that while $d$, $r$, and $T$ are determined by the explanation setting (with $r=\lfloor d/3\rfloor$ and fixed $T=0.4$ in our main analysis), the parameters $\rho$ and $\kappa$ are hyperparameters required to instantiate a baseline distribution.
Empirically, we find that $\rho \approx 0.7$ and $\kappa \approx 10$ roughly match the scale of observed SHAP value distributions. We treat this as a central reference point while varying $\rho \in \{0.6, 0.7, 0.8\}$ and $\kappa \in \{5, 6, \dots, 15\}$ to show how the reference scale varies with these choices.
Importantly, we use these parameters only to define a plausible perturbation regime. The baseline remains an independent stochastic process that reflects how much explanations could vary.

\subsection{Mallows model under Kendall-Tau distance}
\label{app:mallows_def}

\paragraph{Choice of the Mallows null.}
For rank-based metrics, we require a null model that captures
plausible variability in ranked feature importance while
retaining a shared latent ordering. The Mallows model generates
rankings by perturbing a common central ranking with controlled
dispersion, making it a principled reference distribution for
rank-based stability metrics such as top-$k$ Jaccard and RBO.
\begin{definition}[Mallows model under Kendall-Tau distance~\citep{mallows1957non}] Let $\mathcal{S}_d$ denote the set of all permutations of $[d]$, representing rankings of $d$ features, and let $d_{\mathrm{KT}}$ be the Kendall-Tau distance (i.e., the number of pairwise inversions between two permutations) on $\mathcal{S}_d$. For a central ranking $\pi_0 \in \mathcal{S}_d$ and a dispersion parameter $q \in [0,1)$, the \textit{Mallows model} on $\mathcal{S}_d$ is defined as:
\begin{equation}
\mathbb{P}(\pi \mid \pi_0, q)
=
\frac{q^{d_{\mathrm{KT}}(\pi,\pi_0)}}{Z_d(q)},
\qquad
Z_d(q)
=
\sum_{\sigma \in \mathcal{S}_d}
q^{d_{\mathrm{KT}}(\sigma,e)},
\label{eq:mallows_def}
\end{equation}
where $e=(1,2,\dots,d)$ denotes the identity permutation. The normalizing constant $Z_d(q)$ is the same for any center $\pi_0$ due to the right-invariance of the Kendall-Tau distance.
\end{definition}

We first sample a central ranking $\pi_0 \sim \mathrm{Unif}(\mathcal{S}_d).$ Conditioned on $\pi_0$, we then generate two ranking permutations centered at $\pi_0$ as follows:
\[
\pi, \pi'
\;\overset{\mathrm{i.i.d.}}{\sim}\;
\mathrm{Mallows}(\pi_0, q),
\]
where $q$ controls the magnitude of the perturbations. This construction defines a \textit{shared-center null model}: $\pi$ and $\pi'$ are conditionally independent given $\pi_0$, but marginally dependent through the common latent ranking. Compared to the hypergeometric distribution, which assumes uniformly random top-$r$ sets, the shared-center Mallows model provides a controlled and more realistic stochastic variability.

For any ranking-based functional $h: \mathcal{S}_d \times \mathcal{S}_d \to \mathbb{R}$ (e.g., Top-$k$ Jaccard distance or $1-\mathrm{RBO}$), we define the Mallows baseline as:
\begin{equation}
b_h(q)
:=
\mathbb{E}_{\pi_0} \mathbb{E}_{\pi,\pi' \mid \pi_0} \bigl[h(\pi,\pi')\bigr].
\label{eq:mallows_baseline_def}
\end{equation}

Since all quantities of interest are invariant to the choice of the central ranking $\pi_0$,
the baseline simplifies to

\begin{equation}
b_h(q)
=
\mathbb{E}_{\pi,\pi' \sim \mathrm{Mallows}(e,q)}
\bigl[h(\pi,\pi')\bigr].
\label{eq:mallows_baseline_simplified}
\end{equation}
In practice, the expectation in~\eqref{eq:mallows_baseline_simplified} is approximated using $N_{\mathrm{MC}}$ Monte Carlo samples: $\widehat{b}_h(q) = \frac{1}{N_{\mathrm{MC}}}\sum_{t=1}^{N_{\mathrm{MC}}} h\!\left(\pi^{(t)},\pi'^{(t)}\right)$, where $\pi^{(t)},\pi'^{(t)} \sim \mathrm{Mallows}(e,q)$.

For the Top-3 Jaccard baseline, the dimensionality $d$ and $k=3$ are fixed by the experimental settings.
For the RBO-based ranking baseline, we set $p = 1 - 1/d$, where $d$ is the number of features,
so that the expected effective comparison depth, $1/(1-p)$, spans the full feature ranking.
The ranking baselines use $N_{\mathrm{MC}} = 20{,}000$ Monte Carlo samples; $N$ denotes evaluated test instances.

The hyperparameter used to instantiate the ranking-based baseline is the Mallows dispersion parameter $q$.
Empirically, we find that $q \approx 0.4$ matches the scale of observed variability, and we therefore consider $q \in \{0.3, 0.4, 0.5\}$ to capture a range of plausible randomness levels.
In contrast to the $\ell_2$ baseline~\eqref{eq:l2_baseline_final}, which requires calibrating multiple parameters, the ranking-based baseline depends on a single dispersion parameter.
This results in a more constrained and tractable calibration space, and consequently a narrower and more realistic range of randomized baselines.

\section{Feature-level SHAP under preprocessing pipelines}
\label{app:pipeline}
\noindent This appendix describes how our pipeline-based explanation interface ensures that SHAP attributions are reported at the level of semantic input features rather than post-encoding dimensions, enabling consistent feature-level auditing across heterogeneous model classes (trees, ensembles, and neural networks) under a single model-agnostic explainer.

\paragraph{Motivation.}
Our goal is to audit stability at the level of \emph{semantic input features} (e.g., \texttt{race}, \texttt{occupation}), rather than at the level of post-encoding dimensions.
In tabular modeling practice, categorical variables are frequently one-hot encoded prior to training.
Running SHAP directly on one-hot dimensions splits a single semantic feature into many binary indicators, complicating interpretation and downstream auditing.
More importantly for perturbation-based explainers, independently masking or sampling one-hot dimensions may violate mutual-exclusivity constraints (e.g., multiple category indicators active simultaneously), producing coalitions that are not meaningful in the original feature space and potentially distorting attribution comparisons.

\paragraph{Pipeline design.}
To preserve the training-time preprocessing while producing feature-level explanations, we implement the predictor as an \texttt{sklearn} \texttt{Pipeline}:
\[
f(x) \;=\; g(T(x)),
\]
where $x$ is the original (pre-encoded) tabular input, $T$ is a deterministic preprocessing transform (e.g., one-hot encoding and scaling), and $g$ is the learned model.
All models in our study are trained and evaluated through the same interface, so that SHAP queries operate on the same feature schema across heterogeneous predictors.

\paragraph{Why a model-agnostic explainer.}
This pipeline-centric design supports consistent, feature-level auditing across heterogeneous model classes (trees, ensembles, and neural networks) without re-implementing feature-grouping logic separately per model-specific explainer.
While model-specific SHAP implementations can be more computationally efficient under certain assumptions, using a single model-agnostic explainer avoids confounding differences in explanation backends and ensures that all multiplicity comparisons are performed under the same perturbation semantics and interface.
Accordingly, we use \texttt{KernelSHAP} and \texttt{PermutationSHAP} for the main analysis.

\paragraph{Reproducibility details.}
We keep the preprocessing transform $T$ fixed once trained, and we sample the background dataset $D_{bg}$ from the corresponding training split.
Explainer variability can involve both background and internal coalition sampling. In the subsampled-coalition Kernel SHAP setting, the coalition-level seed parameter does not control all internal randomness. Fixing the budget therefore does not isolate background variation. We do not claim verified internal-seed control for Permutation SHAP; this remains a separate implementation diagnostic.

\section{Experiment Setting Specifics}
\label{app:exp_settings}
\noindent This appendix summarizes the datasets used in our
evaluation, including sample sizes, feature counts, class balance,
and the stratified five-fold partition used to select the evaluated test fold, and details the parameter sweeps used to
instantiate the randomized baselines.

\subsection{Datasets}
We evaluate on four tabular benchmarks spanning different scales
and domains. \german~\citep{german_credit_1994} is a
small-scale credit risk dataset where the task is to classify loan
applicants as good or bad credit risks based on financial and
demographic features. \diabetes~\citep{smith1988using} is a
small-scale medical dataset for predicting diabetes onset from
diagnostic measurements (e.g., glucose, BMI, age) on female
patients of Pima Indian heritage. \acs~\citep{ding2021retiring}
is derived from the American Community Survey (Virginia, 2018) and
frames income prediction (above/below \$50K) as a
fairness-relevant classification task with demographic features.
\gmsc~\citep{kaggle2011givemesomecredit}
is a Kaggle credit-scoring dataset for predicting serious
delinquency within two years; it is notably class-imbalanced
($6.7\%$ positive rate), reflecting realistic deployment conditions
in financial risk modeling. Dataset statistics are summarized in
Table~\ref{tab:datasets}. We use the first test fold of a stratified five-fold partition for the reported comparisons. We remove missing values during preprocessing.

\begin{table}[h!]
    \centering
    \begin{tabular}{lrrrrr}
        \toprule
        Dataset & Features & Total & Train & Test & Positive Rate \\
        \midrule
        German Credit & 16 & 988 & 790 & 198 & 29.9\% \\
        Diabetes & 8 & 392 & 313 & 79 & 33.2\% \\
        ACS Income & 8 & 45,960 & 36,768 & 9,192 & 43.7\% \\
        GMSC & 10 & 45,000 & 36,000 & 9,000 & 6.7\% \\
        \bottomrule
    \end{tabular}
    \caption{Dataset statistics and the evaluated train/test fold from a five-fold stratified partition.}
    \label{tab:datasets}
\end{table}

\paragraph{Same-budget empirical-distribution reference.}
The reference uses the full training background and the same coalition budget as the sampled-background estimator. For Kernel SHAP with $d\leq10$ (Diabetes, ACS Income, GMSC), this budget covers all coalitions; for German Credit ($d=16$), $1024\ll2^{16}$ and the reference is approximate. These comparisons quantify divergence from that reference, not universal error against exact Shapley values.
\paragraph{Reference-model consistency.}
\label{app:reference_caveat}
The sampled and full-background query features align. However, signed SHAP sums plus their base values differ for some neural-network sampled and reference results produced in separate runs. The largest absolute reconstructed-output differences for FTT are 0.030864 on Diabetes and 0.215995 on GMSC; tree-model differences are at numerical roundoff. These neural distances are descriptive comparisons of the retained runs, not isolated effects of changing the background. In contrast, the five background runs within each method agree in reconstructed output to numerical precision. Matching that output does not by itself establish identical model functions on all masked inputs.

\subsection{Baseline Parameters}

The shaded $[\min,\max]$ bands span reference values over a null-parameter sweep. Their lower endpoint is the descriptive exceedance threshold. They are neither confidence intervals nor tests with a prescribed false-positive rate.

\paragraph{$\ell_2$ baseline.}
The RMS reference $\sqrt{\mathbb{E}\|X-Y\|_2^2}$ from the Dirichlet null (Proposition~\ref{prop:l2_baseline}) is parameterized by the total
SHAP mass $T$, the top-$r$ concentration $\rho$, and the Dirichlet
concentration $\kappa$. We fix $T = 0.4$ (matching the empirical
average across datasets) and $r = \lfloor d/3 \rfloor$, and sweep
$\rho \in \{0.6, 0.7, 0.8\}$ and $\kappa \in \{5, 6, \ldots, 15\}$
to capture a range of plausible top-feature concentrations and
dispersion levels.

\paragraph{Jaccard and RBO baselines.}
The Mallows null model (Appendix~\ref{app:mallows_def}) is centered
at the identity ranking. We sweep the dispersion parameter
$q \in \{0.3, 0.4, 0.5\}$. For each $q$, we estimate the expected
distance using $20{,}000$ Monte Carlo trials. For Jaccard, we use
$k = 3$; for RBO, we use persistence parameter $p = 1 - 1/d$,
adaptive to the number of features.

\paragraph{Rank construction and ties.}
\label{app:metric_ties}
For Jaccard and RBO, features are ordered by absolute attribution. The implementation sorts $|\phi|$ in ascending order with NumPy's default \texttt{argsort} (quicksort): Jaccard selects the last $k$ indices, and RBO reverses the complete ordering. Exact ties follow this implementation's ordering; they are not assigned average ranks. The separate Top-1 churn diagnostic uses \texttt{argmax}, selecting the first feature index in a tie. Consensus-rank and selection/rank diagnostics instead sort $-|\phi|$ (and the mean absolute attribution for the consensus) using the same default sort. These conventions preserve the reported calculations; tied features need not receive the same ordering across these diagnostics.

\subsection{Sampling Methods}
\noindent This appendix details the four background sampling strategies compared in Section~\ref{sec:experiments} (Random, K-Means, PredStrat, CTE) and the per-model hyperparameter search spaces used during inner-fold tuning.

\paragraph{Random Sampling} We sample $K$ training instances uniformly at random without replacement, using a NumPy random-number generator controlled by the background seed.
\paragraph{K-Means Sampling}
K-Means sampling summarizes the training distribution with $K$ centroids. Each centroid coordinate is rounded to the nearest observed value of that feature; the resulting vector need not be an actual training row. Kernel SHAP weights these representatives by cluster size, while the Permutation SHAP masker uses their unweighted array. The background seed is passed to K-Means initialization, so this method is not assumed deterministic. It nevertheless exhibits low run-to-run variation in the retained comparisons.

\paragraph{PredStrat} (Prediction-Stratified Sampling):
Let $f(x_i) \in [0,1]$ denote the model's predicted probability for training
instance $x_i$. The $\{0, 20, 40, 60, 80, 100\}$-th percentiles of these
probabilities define $m=5$ prediction bins $\mathcal{P}_1,\ldots,\mathcal{P}_m$.
Tied predictions can produce bins of unequal size or empty bins.
From each bin $\mathcal{P}_j$, we sample
$\min\{\lfloor K/m\rfloor, |\mathcal{P}_j|\}$ instances uniformly without replacement.
Any remaining slots are filled by uniform sampling without replacement from
unselected training instances, yielding $K$ background instances.
For $K=100$, the target allocation is 20 instances per bin.

\paragraph{CTE (Compress Then Explain)}
CTE~\citep{baniecki2025cte} constructs a highly representative background dataset by compressing the original training data distribution via kernel thinning. Rather than relying on simple random selection or distance-based clustering, this approach carefully selects a coreset of exactly $K$ instances that is designed to approximate the underlying statistical properties of the full dataset. CTE aims to reduce approximation error and explainer-induced variability; the compression procedure can itself depend on its random seed.

\paragraph{Hyperparameter Optimization}
\label{app:hyperparameters}

For all models, we performed hyperparameter tuning within the training partition of the evaluated fold. To reduce computational cost while maintaining robustness, we employed a validation set strategy using \texttt{ShuffleSplit} (1 split, test size = 0.2, random state = 0) rather than full inner cross-validation. The best hyperparameter configuration was selected based on the \textbf{ROC-AUC} score.
\subsection{Hyper Parameters for Model Training}
Table~\ref{tab:hyperparameters} summarizes the fixed parameters and the grid search space for each model.

\begin{table}[h]
    \centering
    \caption{\textbf{Hyperparameter Search Space.} Fixed parameters are set to ensure fair comparison or convergence, while grid parameters are optimized using a validation set.}
    \label{tab:hyperparameters}
    \renewcommand{\arraystretch}{1.2}
    \small
    \begin{tabular}{C{3.0cm}|C{3.0cm}|C{5cm}}
        \toprule
        \textbf{Model} & \textbf{Fixed Parameters} & \textbf{Grid Search Parameters (Values)} \\
        \midrule

        \multirow{2}{*}{\textbf{Decision Tree}} & \multirow{2}{*}{-}
        & \texttt{max\_depth} $\in \{3, 5, \text{None}\}$ \\
        & & \texttt{min\_samples\_leaf} $\in \{1, 5, 10\}$ \\
        \midrule

        \multirow{2}{*}{\textbf{Random Forest}} & \texttt{n\_estimators} $= 200$
        & \texttt{max\_depth} $\in \{\text{None}, 7, 15\}$ \\
        & & \texttt{min\_samples\_leaf} $\in \{1, 5\}$ \\
        \midrule

        \multirow{3}{*}{\textbf{XGBoost}} & \texttt{n\_estimators} $= 200$
        & \texttt{max\_depth} $\in \{3, 5\}$ \\
        & \texttt{tree\_method} $= \text{`hist'}$
        & \texttt{learning\_rate} $\in \{0.05, 0.1\}$ \\
        & \texttt{eval\_metric} $= \text{`logloss'}$
        & \texttt{subsample} $\in \{0.8, 1.0\}$ \\
        \midrule

        \multirow{3}{*}{\textbf{FT-Transformer}} & \texttt{attn\_heads} $= 8$
        & \texttt{d\_block} $\in \{64, 128\}$ \\
        & \texttt{dropout} $= 0.1$
        & \texttt{n\_blocks} $\in \{2, 3\}$ \\
        & \texttt{optimizer} $= \text{AdamW}$
        & \texttt{lr} $\in \{1\text{e-}4, 3\text{e-}4\}$ \\
        \midrule

        \multirow{4}{*}{\textbf{MLP}} & \texttt{activation} $= \text{ReLU}$
        & \texttt{hidden\_dims} $\in \{(64), (128), (128, 64)\}$ \\
        & \texttt{batch\_size} $= 256$
        & \texttt{lr} $\in \{1\text{e-}3, 3\text{e-}4\}$ \\
        & \texttt{epochs} $= 100$
        & \texttt{weight\_decay} $\in \{1\text{e-}4, 1\text{e-}3\}$ \\
        &
        & \texttt{dropout} $\in \{0.0, 0.1\}$ \\
        \bottomrule
    \end{tabular}
\end{table}

\paragraph{Seed control.}
\label{app:sample_seed}
We index five background runs by seeds $\{0,42,84,126,168\}$. For each evaluated instance, all ten unordered run pairs enter the distance collection. Exceedance is the percentage of $10N$ run-pair--instance comparisons whose distance strictly exceeds the threshold, as in Eq.~\eqref{eq:pooled_exceedance}; distances are not averaged before thresholding. Shared seed labels do not guarantee identical background draws across methods or control all internal sampling. A confidence group $g$ with $N_g$ instances contributes $10N_g$ distances. Reference comparisons retain five run--instance distances per query, hence $5N$ observations.

\FloatBarrier

\section{Additional Experimental Results}
\label{app:exp}
\subsection{Churn depth, background size, and additional analyses}
\label{app:revision_evidence}

The churn-depth, Top-$k$, background-size, and high-dimensional analyses reported below use Kernel SHAP.

\paragraph{Units and churn depth.}
We examine whether disagreement is confined to the Top-3 boundary or also changes the leading features.
Top-1 \emph{any-seed} churn is the fraction of $N$ test instances for which the five runs do not all select the same leading feature. Top-1 \emph{pairwise} churn counts differing leading features over $10N$ run-pair--instance observations. Neither uses a Mallows threshold. For FT-Transformer at $K=100$, any-seed churn is 8.1\%/10.1\% on German Credit, 11.4\%/15.2\% on Diabetes, and 67.7\%/62.0\% on GMSC under Random/CTE, respectively. On GMSC, the corresponding pairwise rates are 42.9\%/32.0\%. Leading-feature churn is lower on the two smaller datasets, while GMSC shows substantial changes even at Top-1.

Raw Top-3 Jaccard distance has values $\{0,0.5,0.8,1\}$, corresponding to zero through three replaced members. Its frequencies pool $10N$ run-pair--instance distances and describe membership changes, not ordering within a fixed set. In a consensus-rank analysis, consensus features are selected by mean absolute SHAP across runs; each selected consensus rank has $5N$ run--instance rank observations. This averaging selects a feature; it is not an average of pair distances.

\begin{figure}[htbp]
\centering
\includegraphics[width=\textwidth]{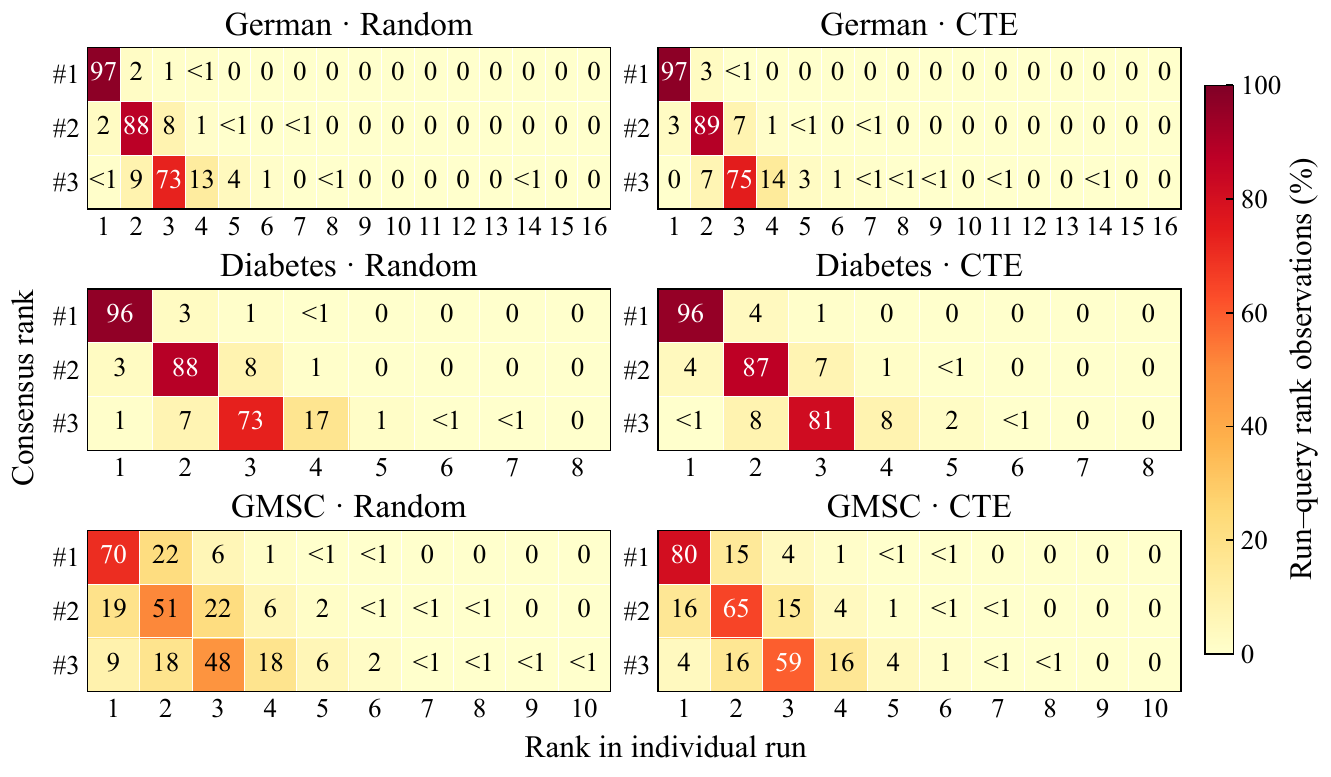}
\caption{\textbf{Rank positions of consensus features.} For each query, the five-run mean absolute SHAP values select the first three consensus features. Each row records the rank of one such feature in the five contributing runs, pooling $5N$ observations; row frequencies sum to $100\%$ before rounding. Columns include every possible rank. Entries are rounded percentages; $<1$ marks positive values that round to zero. Panels use FT-Transformer, $K=100$, and a shared color scale. This within-sample rank summary is distinct from the Top-1 event frequencies in Table~\ref{tab:top1_churn}.}
\label{fig:appendix_churn_depth}
\end{figure}

Figure~\ref{fig:appendix_churn_depth} shows where these consensus features appear in individual runs; Table~\ref{tab:top1_churn} reports the two Top-1 summaries separately.

\begin{table}[htbp]
\centering
\caption{Top-1 churn for FT-Transformer with $K=100$. Any-seed churn uses $N$ queries, whereas pairwise churn uses all $10N$ run-pair--instance comparisons. Both are event frequencies without a randomized threshold.}
\label{tab:top1_churn}
\begin{tabular}{llrrr}
\toprule
Dataset & Method & $N$ & Any-seed & Pairwise \\
\midrule
German Credit & Random & 198 & 8.1\% & 4.4\% \\
              & CTE    & 198 & 10.1\% & 4.9\% \\
Diabetes      & Random & 79 & 11.4\% & 5.9\% \\
              & CTE    & 79 & 15.2\% & 7.3\% \\
GMSC          & Random & 9,000 & 67.7\% & 42.9\% \\
              & CTE    & 9,000 & 62.0\% & 32.0\% \\
\bottomrule
\end{tabular}
\end{table}

\FloatBarrier

\paragraph{Sensitivity to the selected depth.}
We vary the Top-$k$ depth to assess whether the observed disagreement depends on selecting $k=3$.
We recompute the Mallows reference for each $k\in\{2,3,4,5\}$ and report exceedance over $10N$ run-pair--instance comparisons. For German Credit/FTT, averaging the resulting percentages across Random, K-Means, PredStrat, and CTE gives 14.3\%, 25.9\%, 35.8\%, and 45.7\%, respectively. These average four method-level percentages, not per-instance distances. On GMSC/FTT, Top-2 pair--instance exceedance is 52.6\% under Random and 37.8\% under CTE. These changes cannot be explained solely by a rank-3/rank-4 swap. Pairwise Top-1 churn at $k=1$ remains a separate, uncalibrated event frequency.

\begin{figure}[htbp]
\centering
\includegraphics[width=\textwidth]{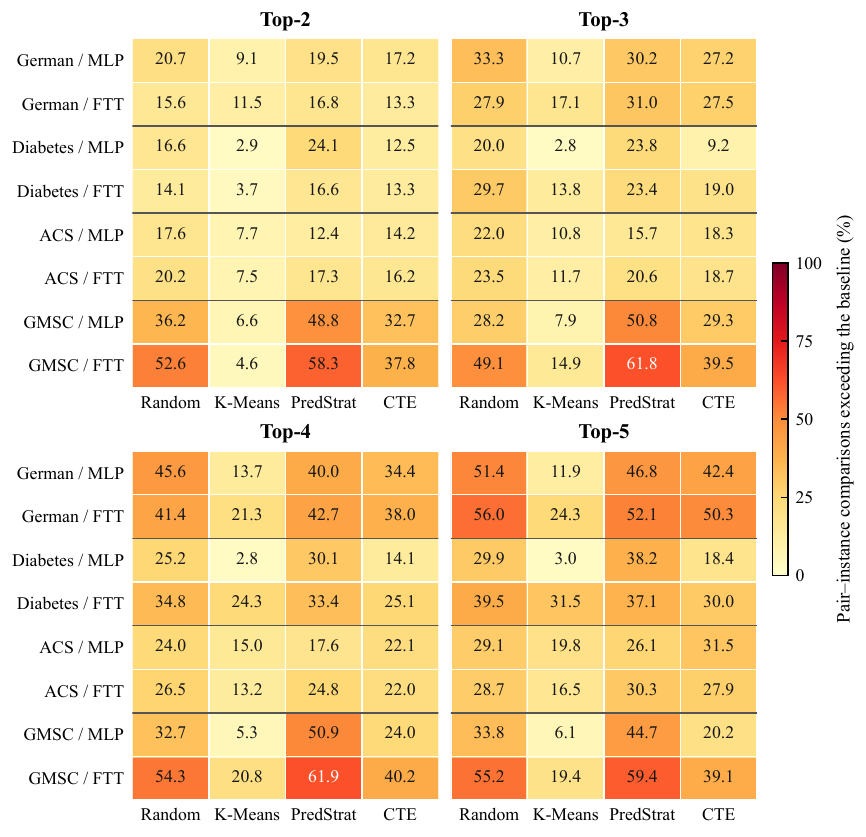}
\caption{\textbf{Sensitivity to Top-$k$ depth.} Each cell is the percentage of $10N$ run-pair--instance Jaccard distances strictly above the dataset- and $k$-specific Mallows baseline, with $K=100$. No pair distances are averaged before thresholding. All four panels share the same color scale; the $k=3$ panel matches the main analysis. In these settings, the threshold is below the smallest positive Jaccard distance, so exceedance equals the frequency of Top-$k$ set changes. The panels need not increase monotonically with $k$.}
\label{fig:appendix_topk_depth}
\end{figure}

Figure~\ref{fig:appendix_topk_depth} reports each method separately using the same $K=100$ explanations as the main analysis.

\FloatBarrier

\paragraph{Background size.}
We vary the number of background points to assess whether disagreement persists beyond the main $K=100$ setting.
The sweep reuses the main $K=100$ results and adds $K\in\{25,50,200,400,800,1600\}$ where supported, with full-training-background endpoints for the smaller datasets. Each size retains the $10N$ run-pair--instance denominator. On GMSC/FTT under Random sampling, Top-3 exceedance is 49\% at $K=100$ and 19\% at $K=1600$. For Diabetes, the full-background percentages round to zero across metrics and methods. On German Credit, residual Top-3 exceedance remains at approximately 9--13\% for Random, PredStrat, and CTE evaluated with MLP and FTT; coalition subsampling prevents a background-only interpretation.

\begin{figure}[htbp]
\centering
\includegraphics[width=\textwidth]{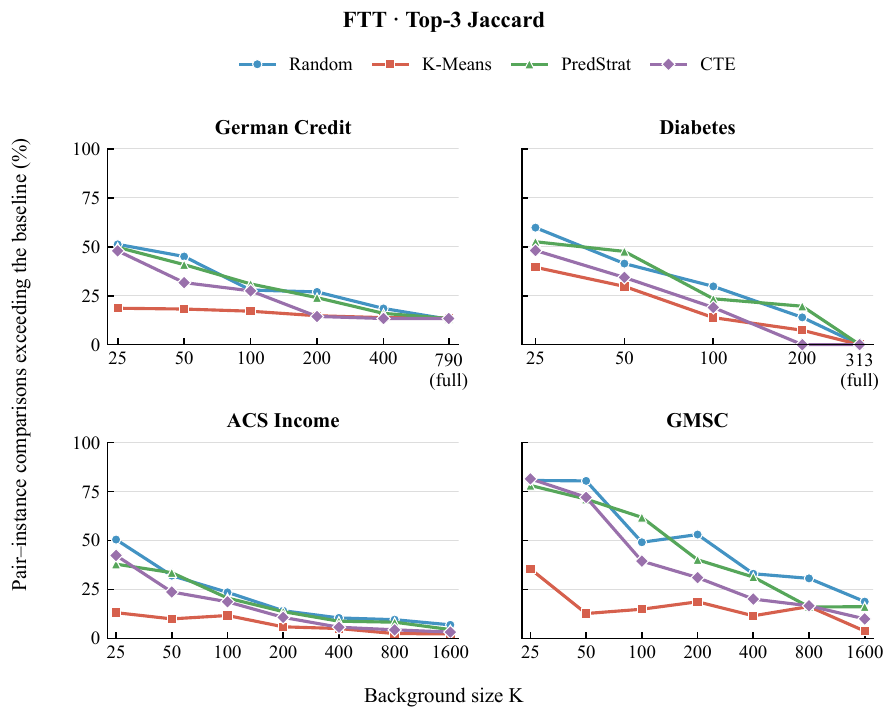}
\caption{\textbf{Background size and Top-3 exceedance: FT-Transformer.} Each point reports the percentage of $10N$ run-pair--instance distances strictly exceeding the dataset's fixed baseline. The horizontal axis is logarithmic and marks the actual evaluated sizes, including the full-training endpoints $K=790$ (German Credit) and $K=313$ (Diabetes). The $K=100$ points reuse the main inputs. Lines connect observed sizes without smoothing; they do not imply monotonicity or isolate background sampling from coalition variability.}
\label{fig:appendix_background_ftt_jac}
\end{figure}

\begin{figure}[htbp]
\centering
\includegraphics[width=\textwidth]{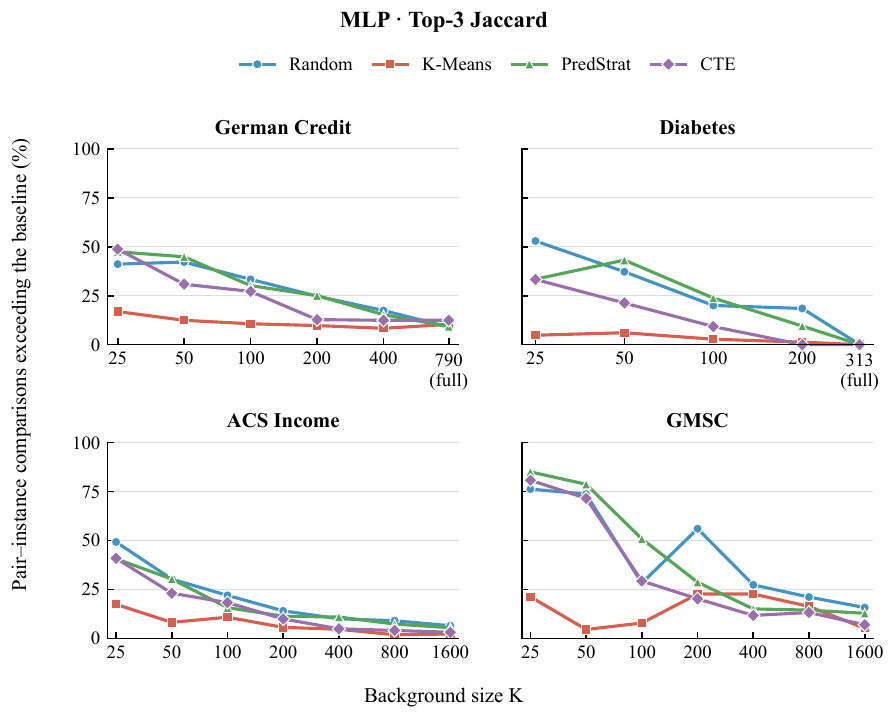}
\caption{\textbf{Background size and Top-3 exceedance: MLP.} The protocol, shared $0$--$100\%$ vertical scale, and strict $10N$ aggregation match Figure~\ref{fig:appendix_background_ftt_jac}. Baselines are fixed across $K$ within each dataset. A larger background does not consistently reduce disagreement at every intermediate size.}
\label{fig:appendix_background_mlp_jac}
\end{figure}

\begin{figure}[htbp]
\centering
\includegraphics[width=\textwidth]{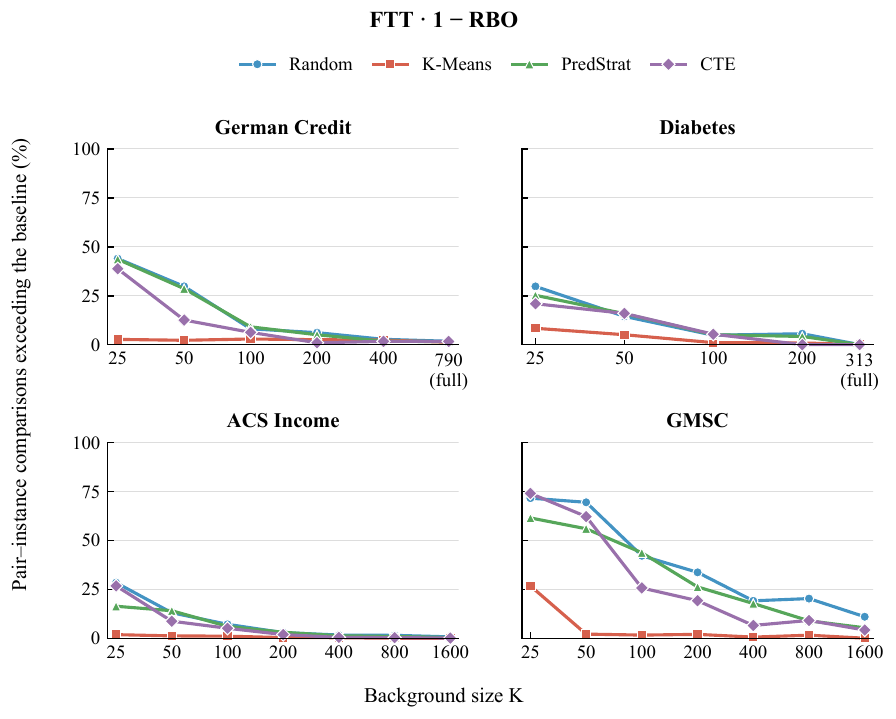}
\caption{\textbf{Background size and $1-$RBO exceedance: FT-Transformer.} Each observed size contributes $10N$ run-pair--instance distances, with strict thresholding before aggregation. The dataset-specific threshold is fixed across sizes. RBO uses the full feature ranking, $p=1-1/d$, and the extrapolated final term. Actual full-training endpoints and canonical $K=100$ reuse follow Figure~\ref{fig:appendix_background_ftt_jac}.}
\label{fig:appendix_background_ftt_rbo}
\end{figure}

\begin{figure}[htbp]
\centering
\includegraphics[width=\textwidth]{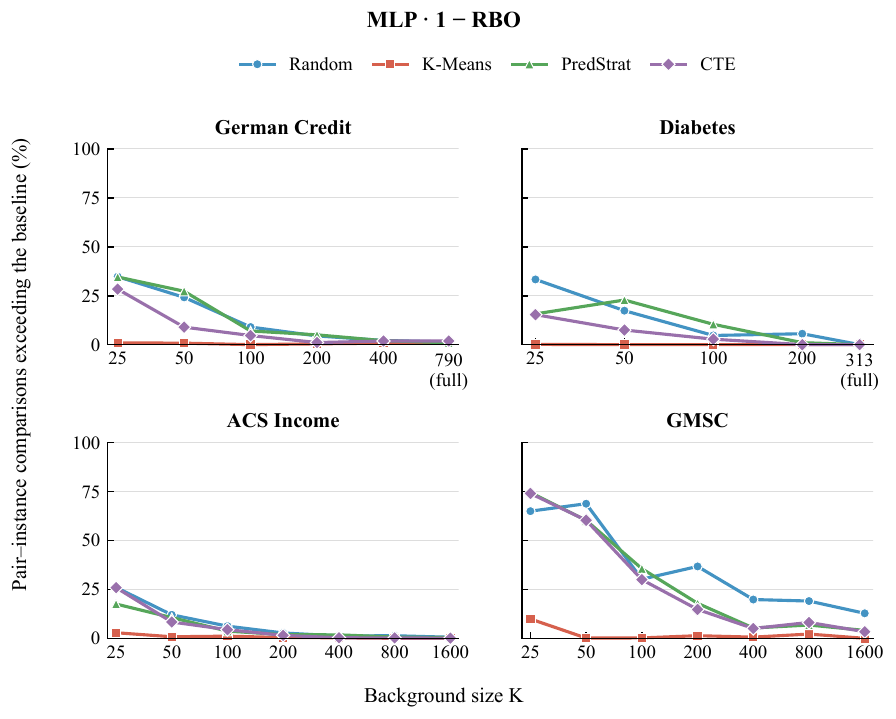}
\caption{\textbf{Background size and $1-$RBO exceedance: MLP.} Each observed size contributes $10N$ run-pair--instance distances, with strict thresholding before aggregation. The dataset-specific threshold is fixed across sizes. RBO uses the full feature ranking, $p=1-1/d$, and the extrapolated final term. Actual full-training endpoints and canonical $K=100$ reuse follow Figure~\ref{fig:appendix_background_ftt_jac}.}
\label{fig:appendix_background_mlp_rbo}
\end{figure}

Figures~\ref{fig:appendix_background_ftt_jac} and~\ref{fig:appendix_background_mlp_jac} show the Top-3 results. The coalition budget is fixed at 256 for Diabetes/ACS and 1,024 for German Credit/GMSC across $K$; this is not a coalition-budget sweep. The larger-background endpoint can improve agreement without a monotone decrease at intermediate sizes. Companion $\ell_2$ results are summarized in Table~\ref{tab:background_l2}; $1-$RBO results appear in Figures~\ref{fig:appendix_background_ftt_rbo}--\ref{fig:appendix_background_mlp_rbo}.

\FloatBarrier

\paragraph{Conditional Shapley values.}
Conditional Shapley methods~\citep{aas2021explaining} account for feature dependence by changing the value function, rather than only the background sampling strategy. We examine whether repeated conditional estimation also produces rank disagreement on the full Diabetes test fold ($N=79$, $d=8$). For each of FTT and MLP, both estimators explain the same fixed CPU-trained checkpoint. Conditional estimation uses \texttt{ctree} with the 313 training observations and 1,000 Monte Carlo samples; marginal Kernel SHAP uses Random backgrounds of size $K=100$ and a coalition budget of 256. We vary the conditional estimator seed or the marginal background seed over five runs, keeping the model fixed.

Table~\ref{tab:conditional_shapley} shows lower Top-1 churn under conditional estimation in both models, with nonzero disagreement remaining. Identical Top-3 sets occur in 83.9\% versus 77.0\% of run-pair--instance comparisons, averaged across models (conditional versus marginal). This comparison concerns the two estimators under their respective settings and does not isolate the effect of conditioning or compare against the original GPU checkpoints.

\begin{table}[htbp]
\centering
\caption{\textbf{Conditional and marginal Shapley estimation on Diabetes.}
Top-1 any-seed churn is the percentage of $N=79$ queries whose leading feature differs across at least two runs. Pairwise Top-1 churn and identical Top-3 sets use all $10N=790$ run-pair--instance comparisons. These are raw event frequencies, without a randomized threshold.}
\label{tab:conditional_shapley}
\begin{tabular}{llrrr}
\toprule
Model & Estimator & Any-seed Top-1 & Pairwise Top-1 & Identical Top-3 \\
\midrule
FTT & Conditional & 10.1\% & 5.2\% & 85.7\% \\
    & Marginal    & 13.9\% & 7.1\% & 73.8\% \\
MLP & Conditional & 12.7\% & 5.8\% & 82.2\% \\
    & Marginal    & 16.5\% & 9.0\% & 80.1\% \\
\bottomrule
\end{tabular}
\end{table}
\FloatBarrier

\paragraph{High-dimensional dataset.}
We examine whether leading-feature and set disagreement also occur in sparse explanations with many input features.
For UCI Arrhythmia~\citep{arrhythmia_1997}, removing one high-missingness column and 32 rows with residual missing values leaves $420$ instances and $278$ features. This experiment uses one test fold ($N=84$), MLP/FTT, Random/CTE, $K=100$, five runs, and 1,024 coalitions. Under \texttt{l1\_reg=num\_features(10)}, each observed explanation has ten nonzero features. Table~\ref{tab:arrhythmia_pilot} shows substantial Top-1 variation and low support-set agreement.

\begin{figure}[htbp]
\centering
\includegraphics[width=\textwidth]{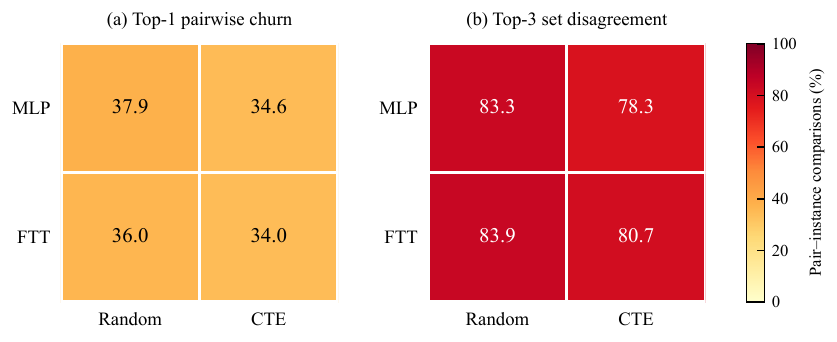}
\caption{\textbf{Pairwise disagreement on Arrhythmia.} Left: the two runs select different leading features. Right: their Top-3 feature sets differ. Both panels use all ten run pairs for each of the 84 test queries ($840$ comparisons per cell) and the same $0$--$100\%$ color scale. These are raw event frequencies, not baseline exceedance rates. Explanations have ten nonzero features out of 278; the low-dimensional Mallows calibration is not applied here.}
\label{fig:appendix_arrhythmia_churn}
\end{figure}

Figure~\ref{fig:appendix_arrhythmia_churn} compares raw Top-1 and Top-3 disagreement on the same $10N=840$ observations. Top-3 sets differ in 78.3--83.9\% of comparisons. This is a different dataset and setting, rather than an ablation isolating the effect of dimensionality.

\begin{table}[htbp]
\centering
\caption{Results on the Arrhythmia dataset. Top-1 any-seed uses $N$ instances; pairwise uses $10N$ run-pair--instance comparisons. Support agreement is mean Jaccard similarity of nonzero supports over those comparisons. Selection/rank churn use only Top-3-disagreeing observations as their denominator and can co-occur.}
\label{tab:arrhythmia_pilot}
\begin{tabular}{llrrrrr}
\toprule
Model & Method & Any-seed & Pairwise & Support & Selection & Rank \\
\midrule
MLP & Random & 54.8\% & 37.9\% & .339 & 77.7\% & 86.3\% \\
MLP & CTE    & 52.4\% & 34.6\% & .358 & 78.4\% & 88.0\% \\
FTT & Random & 54.8\% & 36.0\% & .335 & 70.9\% & 87.2\% \\
FTT & CTE    & 52.4\% & 34.0\% & .362 & 64.2\% & 89.4\% \\
\bottomrule
\end{tabular}
\end{table}

Selection churn involves a differing Top-3 feature absent from the other run's support; rank churn involves a differing member present in both supports. The latter does not count every permutation within an unchanged Top-3 set. With many features tied at zero, we do not apply the low-dimensional Mallows calibration. Background and coalition variability remain mixed; this experiment does not isolate the third seed.

\paragraph{Query-level reporting and rank summaries.}
Table~\ref{tab:instance187_report} retains attribution medians and ranges, rank ranges, and Top-3 inclusion frequencies. Inclusion in at least four of five runs defines an illustrative stable core, not a validated safety threshold.

For a configuration-specific rank summary, let $T_i$ be an instance's mean Kendall distance over the ten pairs and define $C_i=\Pr_{\mathrm{null}}(T\geq T_i)$. The fixed Mallows null uses the same five-ranking, ten-pair average. This score depends on dimension, dispersion, run count, and tie handling; it is not a probability of correctness or a universal cutoff.
For German Credit instance \#187 in Figure~\ref{fig:motivation1}, the FT-Transformer predicts bad credit risk with probability $0.94$. With Kernel SHAP, Random sampling, $K=100$, and five runs, all ten Top-3 Jaccard comparisons exceed $\tau=0.298$, $C=0.002$, and the stable core is empty. The median German Credit instance has $C=0.68$ and a stable core of three features.

\begin{table}[ht]
\centering
\caption{Five-run explanation report for instance \#187. All listed features are in the unstable margin; none meets the illustrative four-of-five Top-3 inclusion rule.}
\label{tab:instance187_report}
\begin{tabular}{lrrrr}
\toprule
Feature & Median & Attribution range & Rank range & Top-3 frequency \\
\midrule
\texttt{savings\_status}   & 0.055 & [0.028, 0.065] & 1--4  & 3/5 \\
\texttt{checking\_status}  & 0.050 & [0.025, 0.065] & 1--5  & 3/5 \\
\texttt{credit\_amount}    & 0.042 & [0.018, 0.088] & 1--6  & 3/5 \\
\texttt{credit\_history}   & 0.038 & [0.005, 0.047] & 3--11 & 3/5 \\
\texttt{duration}          & 0.020 & [0.015, 0.070] & 2--5  & 2/5 \\
\texttt{existing\_credits} & 0.004 & [0.004, 0.084] & 2--11 & 1/5 \\
\bottomrule
\end{tabular}
\end{table}

\FloatBarrier
\paragraph{Background size: additional metrics.}
Table~\ref{tab:background_l2} summarizes $\ell_2$ baseline exceedance across the background-size sweep. For all evaluated configurations with $K\geq100$, exceedance is below $0.21\%$, while rank-based disagreement can persist (Figures~\ref{fig:appendix_background_ftt_jac}--\ref{fig:appendix_background_mlp_jac}). Zero exceedance means that no observed pair distance exceeds the selected threshold; it does not imply zero distances or stable feature rankings. Figures~\ref{fig:appendix_background_ftt_rbo}--\ref{fig:appendix_background_mlp_rbo} show the companion $1-$RBO results under the same protocol.

\begin{table}[htbp]
\centering
\caption{\textbf{Background size and $\ell_2$ baseline exceedance (\%).} At $K=25$ and $K=50$, entries give the minimum--maximum across Random, K-Means, PredStrat, and CTE; a single value denotes an equal rate across methods. The last column is the maximum across all four methods and all evaluated $K\geq100$ for that dataset/model. Each underlying rate uses strict thresholding over $10N$ run-pair--instance comparisons. Ranges are descriptive, not confidence intervals; displayed zeros represent exactly zero exceedances.}
\label{tab:background_l2}
\begin{tabular}{llrrr}
\toprule
Dataset & Model & $K=25$ & $K=50$ & Max., $K\geq100$ \\
\midrule
German Credit & MLP & 0.00 & 0.00 & 0.00 \\
German Credit & FTT & 0.00--1.06 & 0.00--0.20 & 0.20 \\
Diabetes & MLP & 0.00 & 0.00 & 0.00 \\
Diabetes & FTT & 0.63--24.30 & 0.00--1.65 & 0.00 \\
ACS Income & MLP & 0.00--25.44 & 0.00--0.25 & 0.00 \\
ACS Income & FTT & 0.00--31.73 & 0.00--0.61 & 0.00 \\
GMSC & MLP & 0.00 & 0.00 & 0.00 \\
GMSC & FTT & 0.00 & 0.00 & 0.00 \\
\bottomrule
\end{tabular}
\end{table}

\FloatBarrier

\begin{figure}[h]
    \centering
    \begin{subfigure}[b]{0.49\textwidth}
        \centering
        \includegraphics[width=\textwidth]{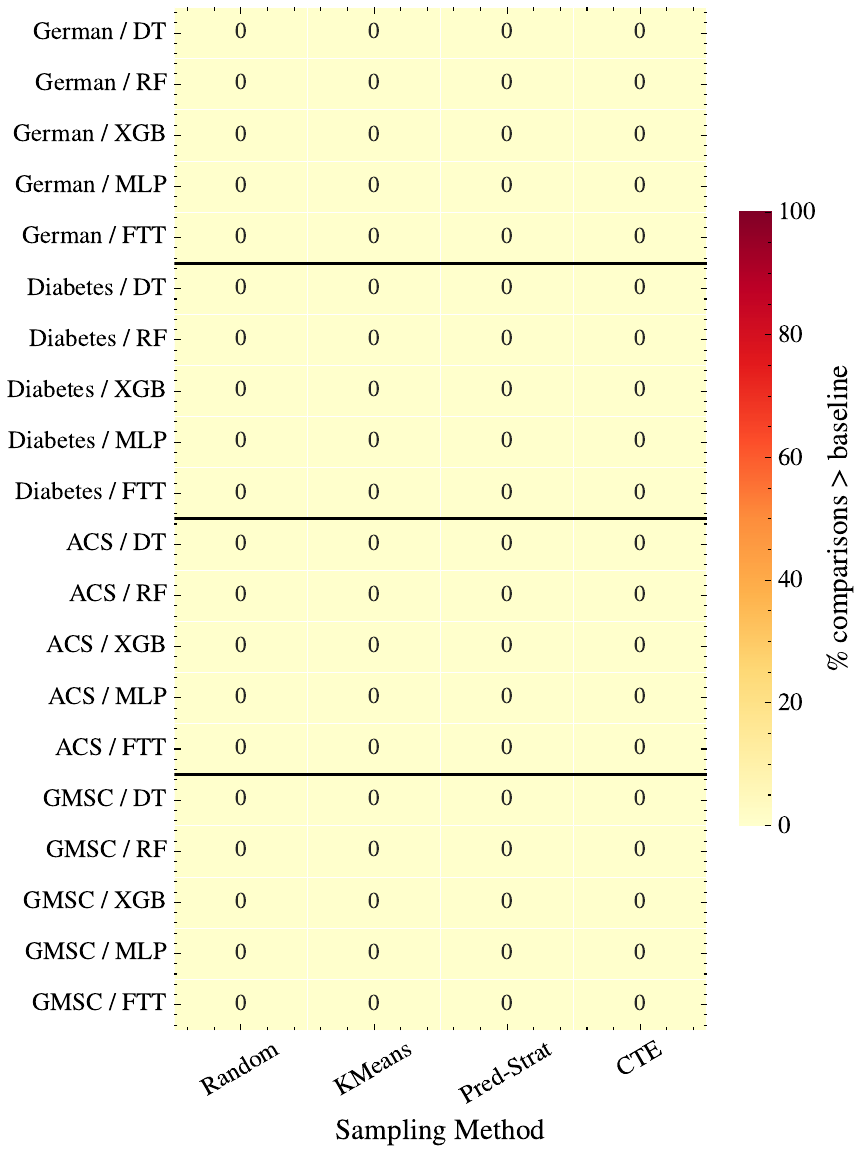}
        \caption{$\ell_2$ Distance}
        \label{fig:heatmap_kernel_l2}
    \end{subfigure}
    \hfill
    \begin{subfigure}[b]{0.49\textwidth}
        \centering
        \includegraphics[width=\textwidth]{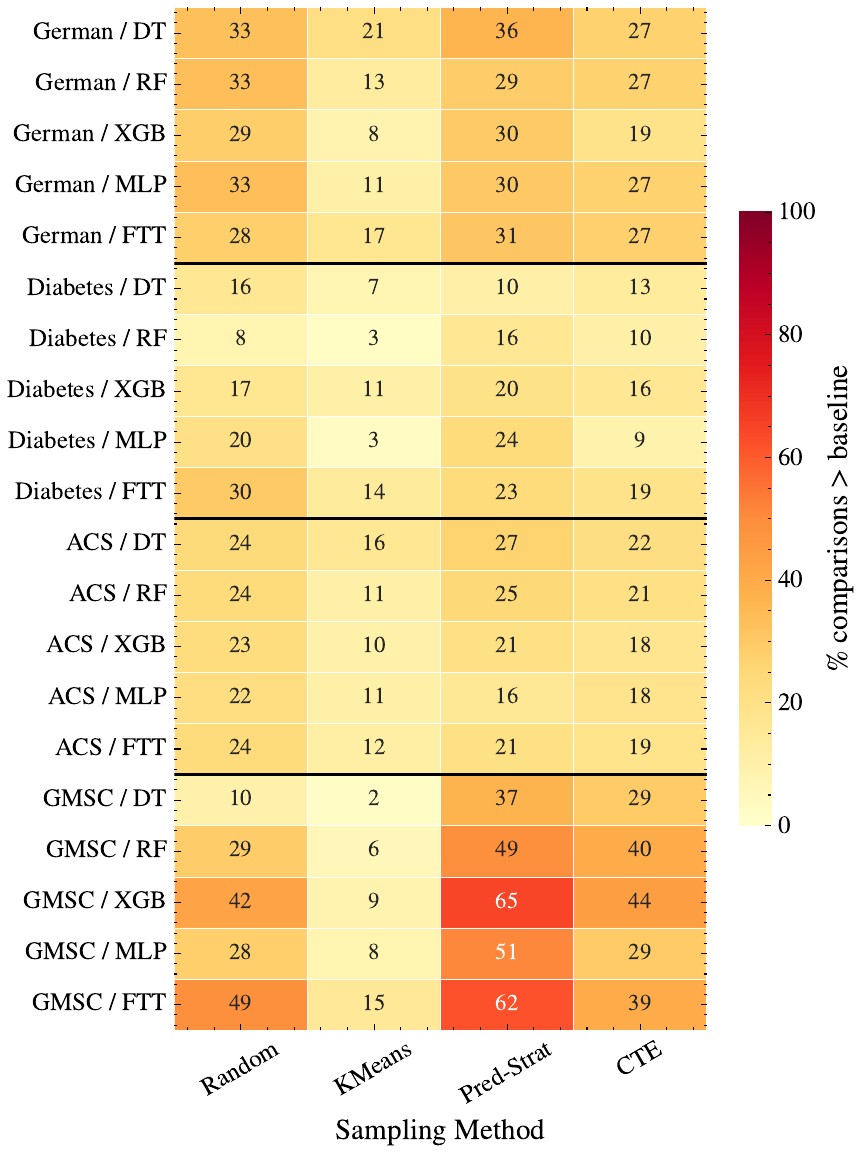}
        \caption{Top-3 Jaccard Distance}
        \label{fig:heatmap_kerneln_jaccard}
    \end{subfigure}
    \hfill
    \begin{subfigure}[b]{0.49\textwidth}
        \centering
        \includegraphics[width=\textwidth]{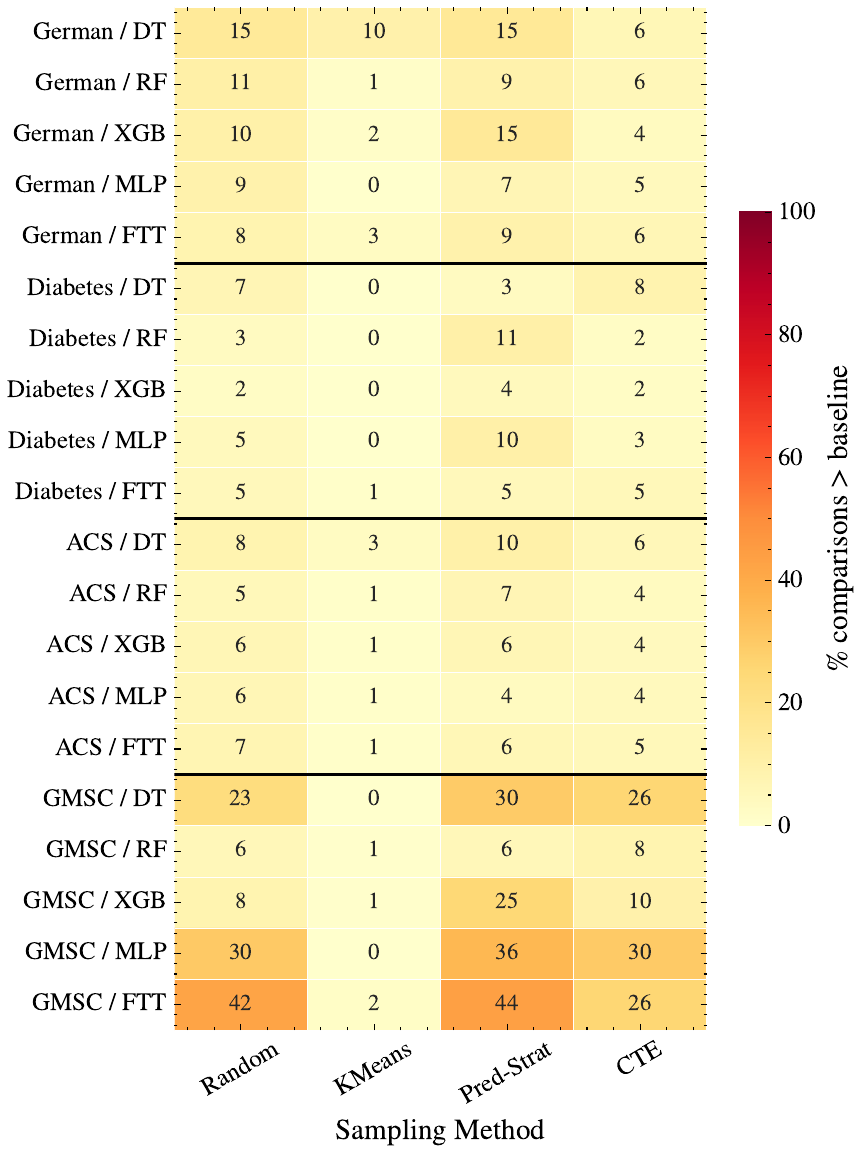}
        \caption{1 - RBO }
        \label{fig:heatmap_kernel_rbo}
    \end{subfigure}
    \hfill
    \caption{\textbf{Pairwise baseline exceedance for Kernel SHAP.}
    For each dataset, model, and method, we report the percentage of $10N$ run-pair--instance distances strictly above the baseline. Sub-panels show results for (a) $\ell_2$ distance, (b) Top-3 Jaccard distance, and (c) $1-$RBO. In the displayed matrices, $\ell_2$ exceedance rounds to $0\%$ while rank-based metrics surface substantial multiplicity, particularly for GMSC and neural architectures.}
    \label{fig:heatmap_full_kernel}
\end{figure}

\begin{figure}[h]
    \centering
    \begin{subfigure}[b]{0.49\textwidth}
        \centering
        \includegraphics[width=\textwidth]{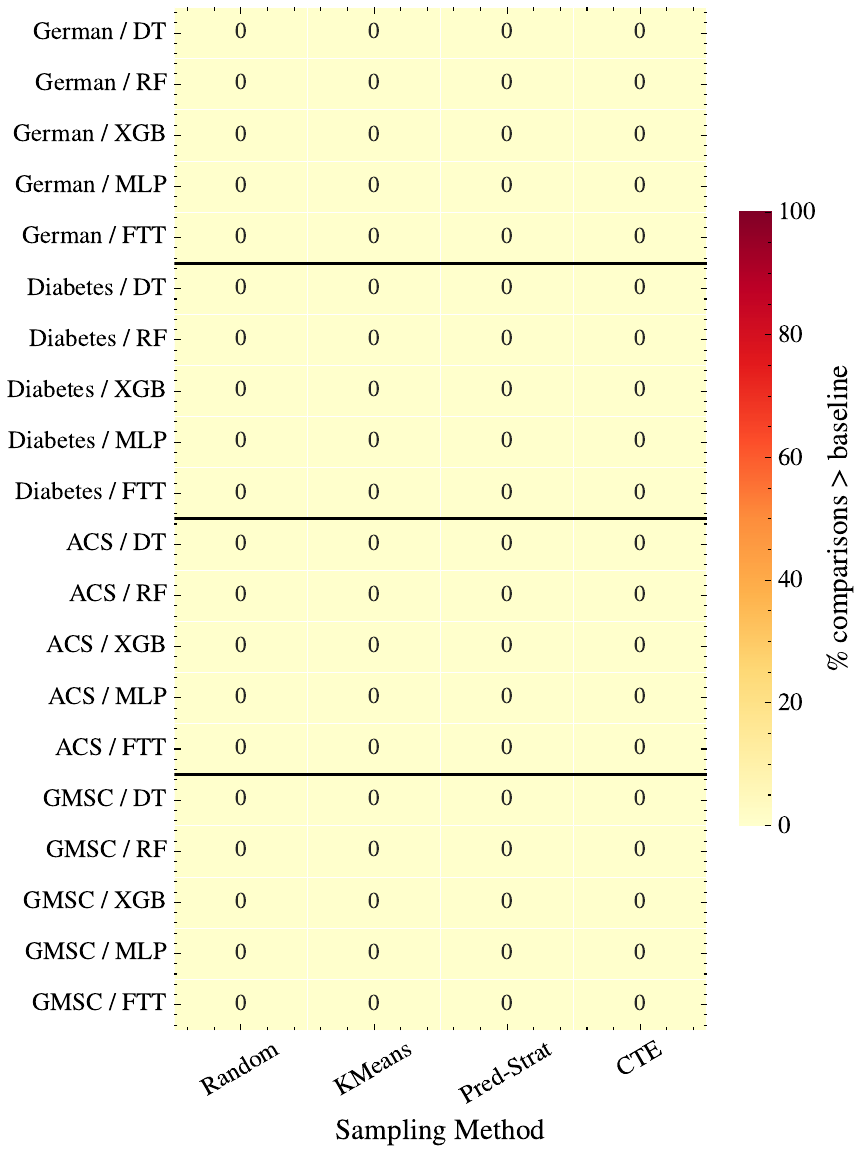}
        \caption{$\ell_2$ Distance}
        \label{fig:heatmap_permutation_l2}
    \end{subfigure}
    \hfill
    \begin{subfigure}[b]{0.49\textwidth}
        \centering
        \includegraphics[width=\textwidth]{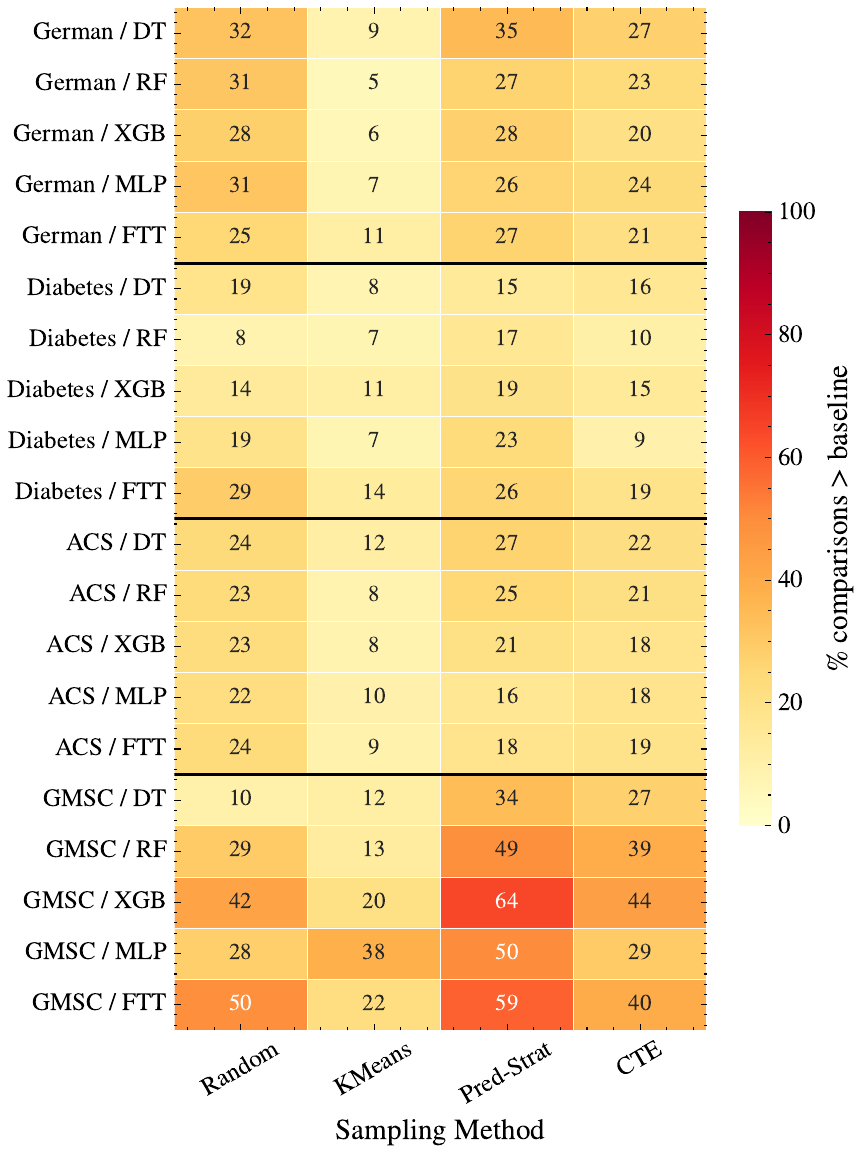}
        \caption{Top-3 Jaccard Distance}
        \label{fig:heatmap_permutation_jaccard}
    \end{subfigure}
    \hfill
    \begin{subfigure}[b]{0.49\textwidth}
        \centering
        \includegraphics[width=\textwidth]{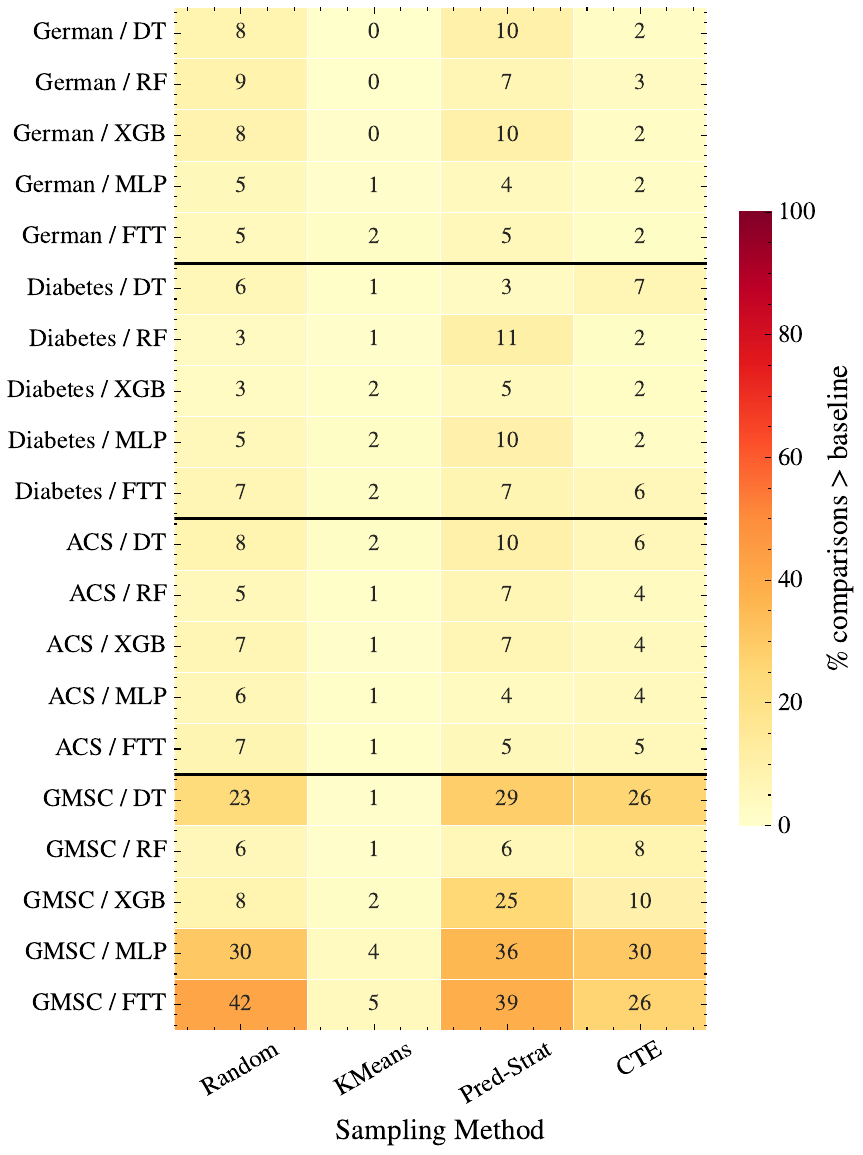}
        \caption{1 - RBO }
        \label{fig:heatmap_permutation_rbo}
    \end{subfigure}
    \hfill
    \caption{\textbf{Pairwise baseline exceedance for Permutation SHAP.}
    Same $10N$ run-pair--instance exceedance protocol as Figure~\ref{fig:heatmap_full_kernel}, using Permutation SHAP. Low $\ell_2$ exceedance coexists with rank-level disagreement. Internal coalition and background randomness are not separately identified by this comparison.}
    \label{fig:heatmap_full_permutation}
\end{figure}

\begin{figure}[H]
    \centering
    \begin{subfigure}[b]{0.99\textwidth}
        \centering
        \includegraphics[width=\textwidth]{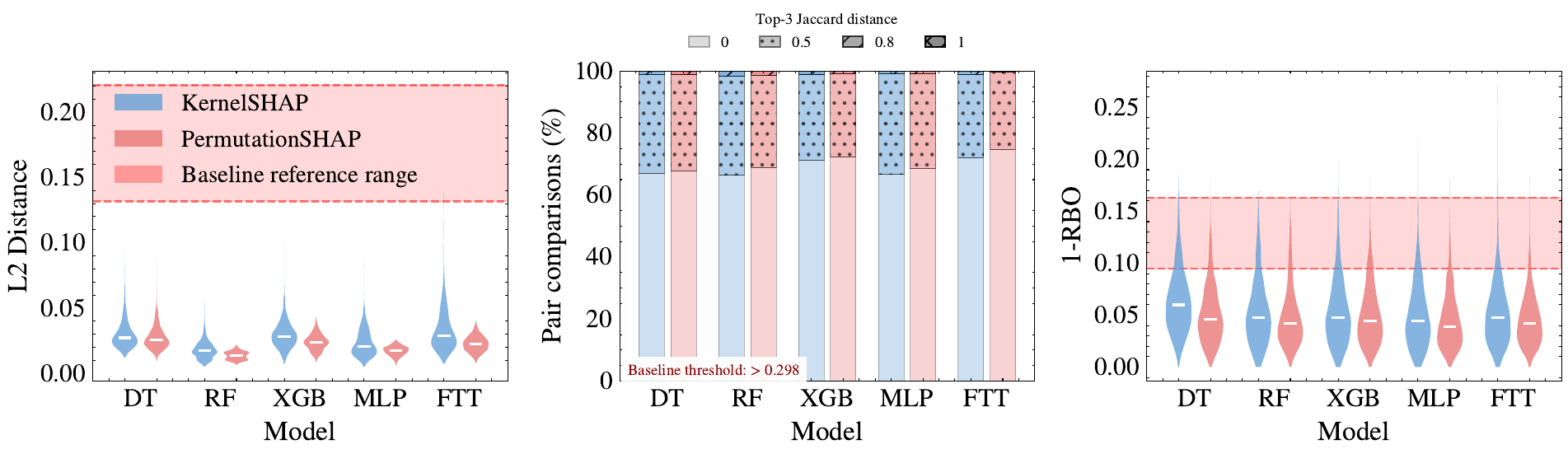}
        \caption{\german}
        \label{fig:german_kernperm}
    \end{subfigure}
    \hfill
    \begin{subfigure}[b]{0.99\textwidth}
        \centering
        \includegraphics[width=\textwidth]{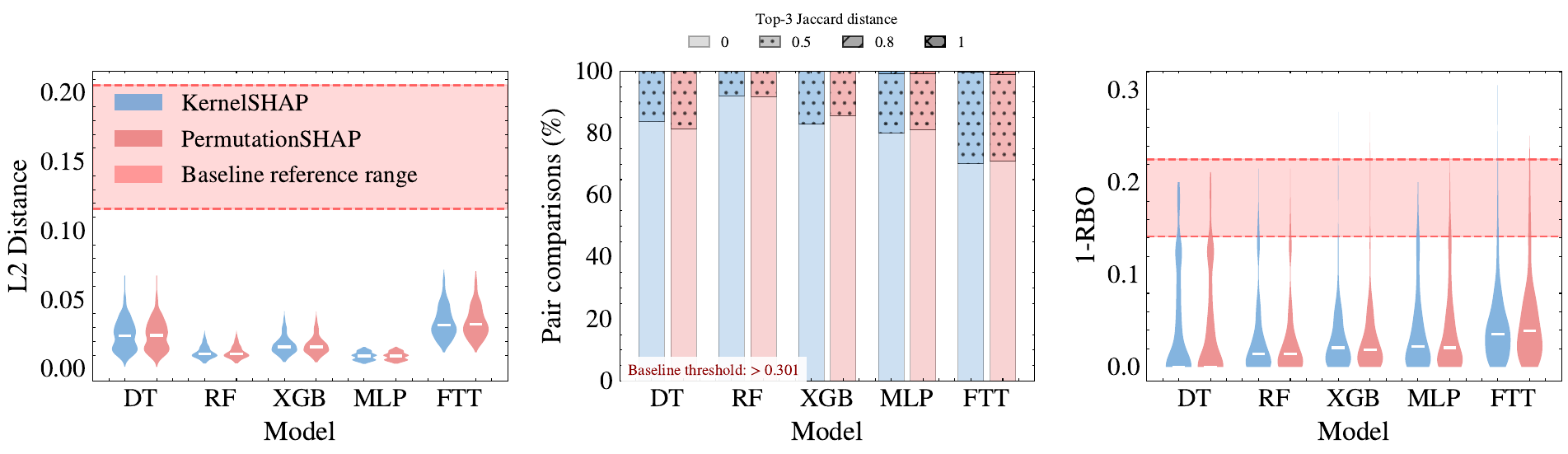}
        \caption{\diabetes}
        \label{fig:diabetes_kernperm}
    \end{subfigure}
    \hfill
    \begin{subfigure}[b]{0.99\textwidth}
        \centering
        \includegraphics[width=\textwidth]{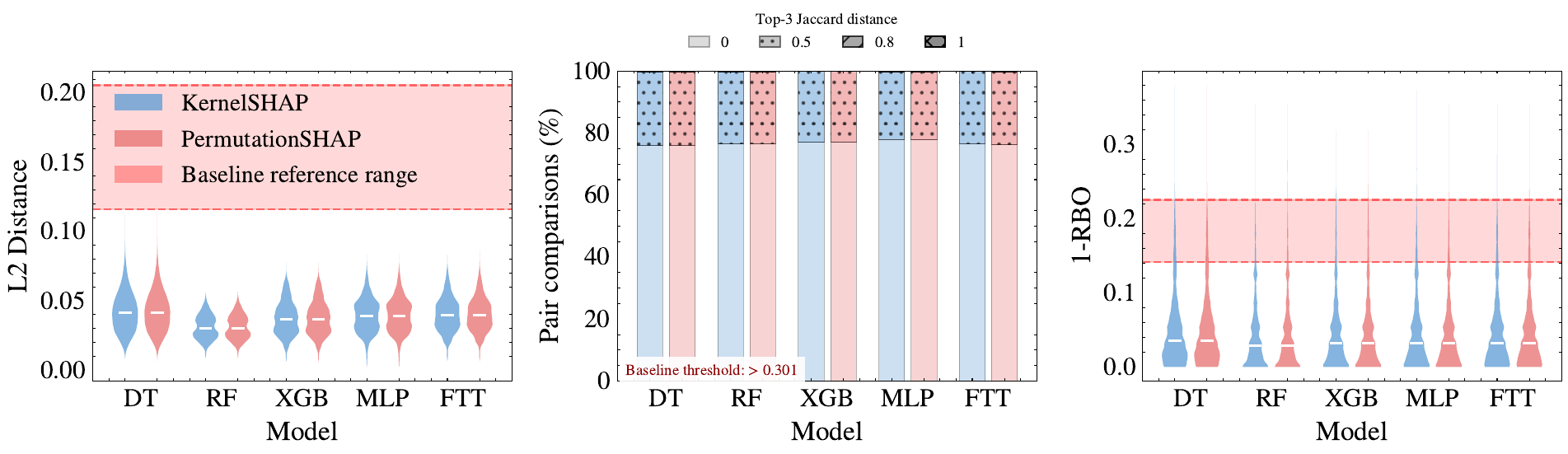}
        \caption{\acs}
        \label{fig:acs_kernperm}
    \end{subfigure}
    \begin{subfigure}[b]{0.99\textwidth}
        \centering
        \includegraphics[width=\textwidth]{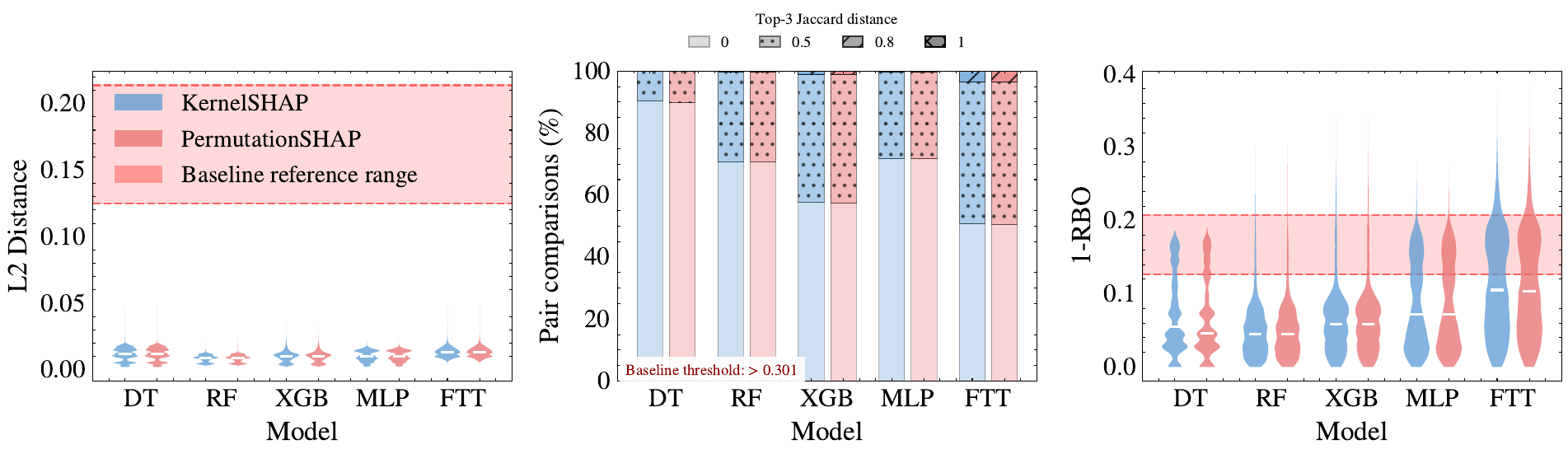}
        \caption{\gmsc}
        \label{fig:gmsc_kernperm}
    \end{subfigure}
    \caption{\textbf{Kernel SHAP vs.\ Permutation SHAP.}
    Distributions of $10N$ run-pair--instance distances across five explainer seeds with Random background sampling, with raw Jaccard shown as frequency bars, comparing Kernel SHAP (blue) and Permutation SHAP (red) across four datasets, five model classes, and three metrics ($\ell_2$, Top-3 Jaccard, $1-$RBO). Shaded bands denote randomized baselines. The two estimators show similar qualitative patterns; this does not establish invariance to all implementation choices or isolate background from coalition randomness.}
    \label{fig:kernel_vs_permutation}
\end{figure}

\paragraph{Kernel vs. Permutation SHAP.}
Figures~\ref{fig:heatmap_full_kernel}--\ref{fig:heatmap_full_permutation} report exceedance; Figure~\ref{fig:kernel_vs_permutation} shows distance distributions. Both explainers exhibit low $\ell_2$ exceedance and rank disagreement; this comparison does not isolate internal sampling.

\begin{figure}[h]
    \centering
    \begin{subfigure}[b]{0.99\textwidth}
        \centering
        \includegraphics[width=\textwidth]{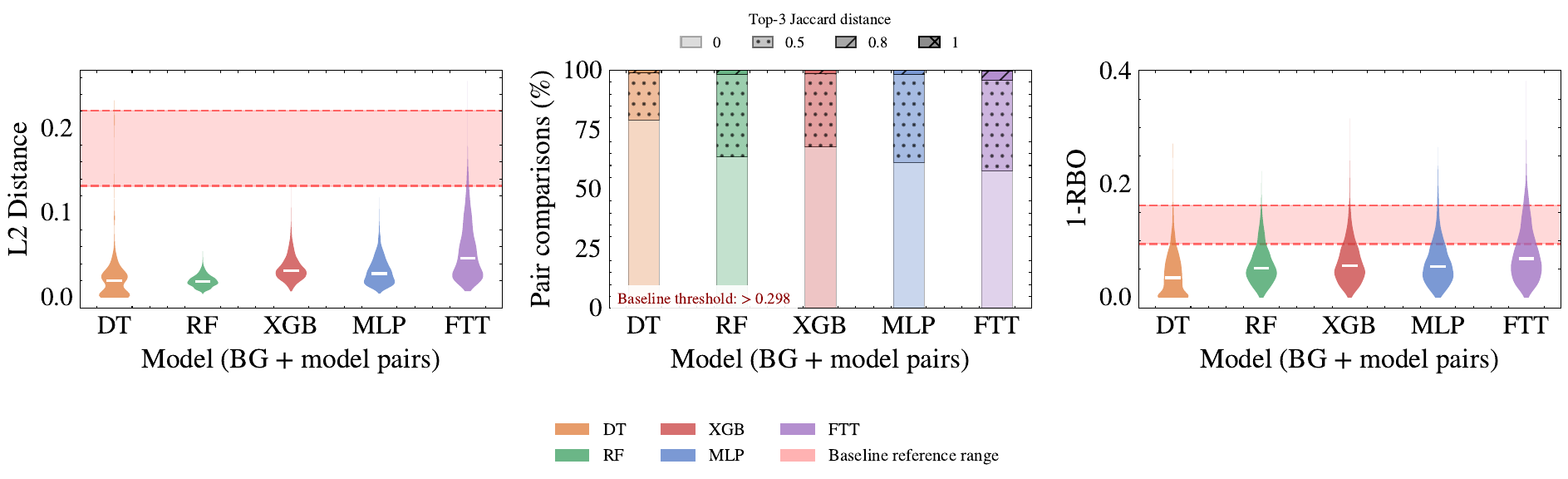}
        \caption{\german}
        \label{fig:german_violin_overall}
    \end{subfigure}
    \hfill
    \begin{subfigure}[b]{0.99\textwidth}
        \centering
        \includegraphics[width=\textwidth]{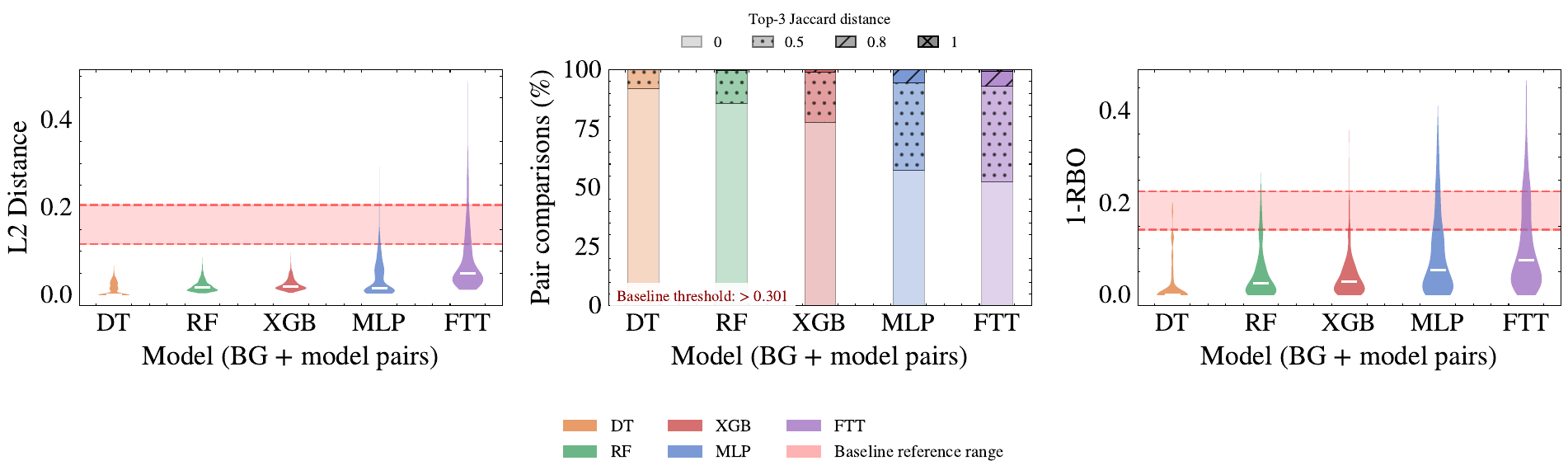}
        \caption{\diabetes}
        \label{fig:diabetes_violin_overall}
    \end{subfigure}
    \hfill
    \begin{subfigure}[b]{0.99\textwidth}
        \centering
        \includegraphics[width=\textwidth]{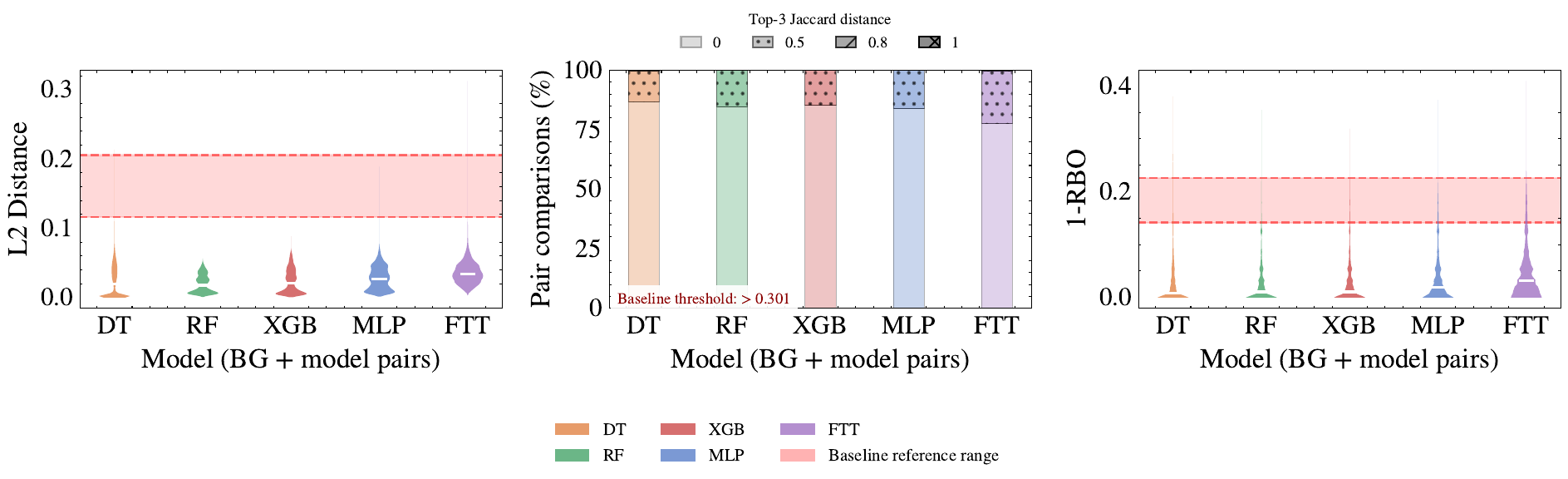}
        \caption{\acs }
        \label{fig:acs_violin_overall}
    \end{subfigure}
    \hfill
    \begin{subfigure}[b]{0.99\textwidth}
        \centering
        \includegraphics[width=\textwidth]{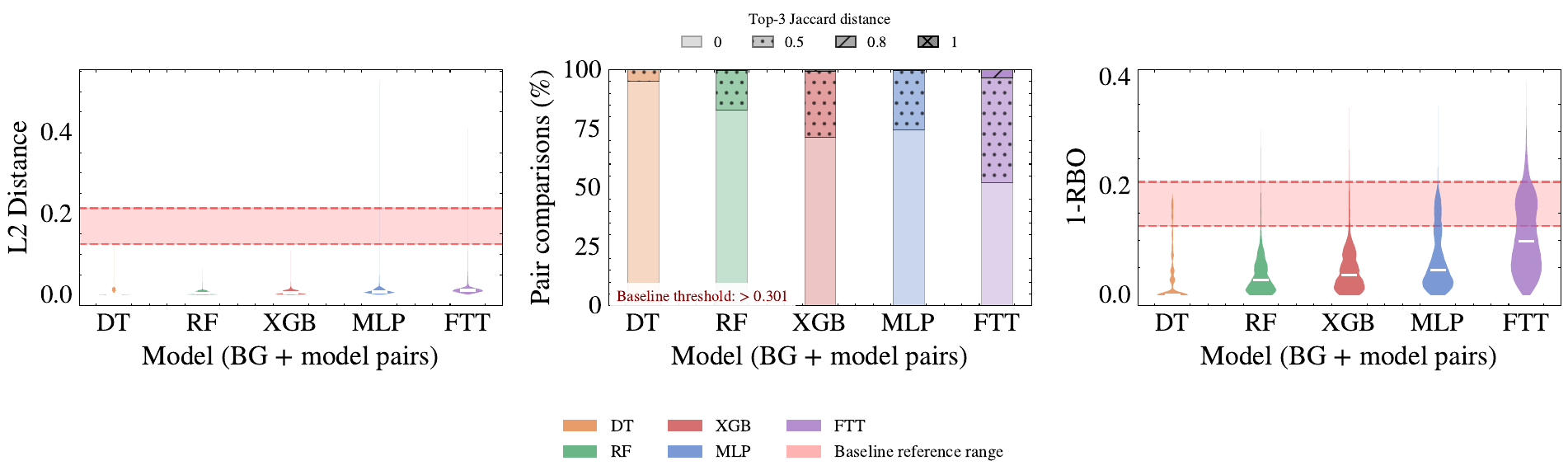}
        \caption{\gmsc}
        \label{fig:gmsc_overall}
    \end{subfigure}
    \hfill
    \caption{\textbf{Overall sensitivity across datasets.} Run-pair--instance distance distributions for Kernel SHAP with Random background sampling on \german, \diabetes, \acs, and \gmsc, measured by $\ell_2$, Top-$k$ Jaccard, and 1-RBO; raw Jaccard uses four-value frequency bars. Model-seed and explainer-seed collections are pooled, each contributing $10N$ distances. Shaded bands indicate randomized baseline ranges. FT-Transformer and MLP exhibit Jaccard multiplicity approaching the baseline, whereas $\ell_2$ remains concentrated near zero across models. On \gmsc and \german, the four Jaccard values carry mass both at zero and at positive membership-change distances.}
    \label{fig:overall_violin}
\end{figure}
\FloatBarrier
\paragraph{Overall sensitivity distributions.}
Figure~\ref{fig:overall_violin} visualizes pooled run-pair--instance distances
for \german, \diabetes, \acs, and \gmsc using three metrics ($\ell_2$, Top-$k$ Jaccard, and 1-RBO).
Consistent with the main text, $\ell_2$ distances often concentrate near zero even when rank-based
metrics show substantial instability, highlighting an \emph{illusion of stability} when only magnitude-based
metrics are reported. These pooled distributions do not by themselves identify recurring modes for an individual query.

\begin{figure}[H]
    \centering
    \begin{subfigure}[b]{0.99\textwidth}
        \centering
        \includegraphics[width=\textwidth]{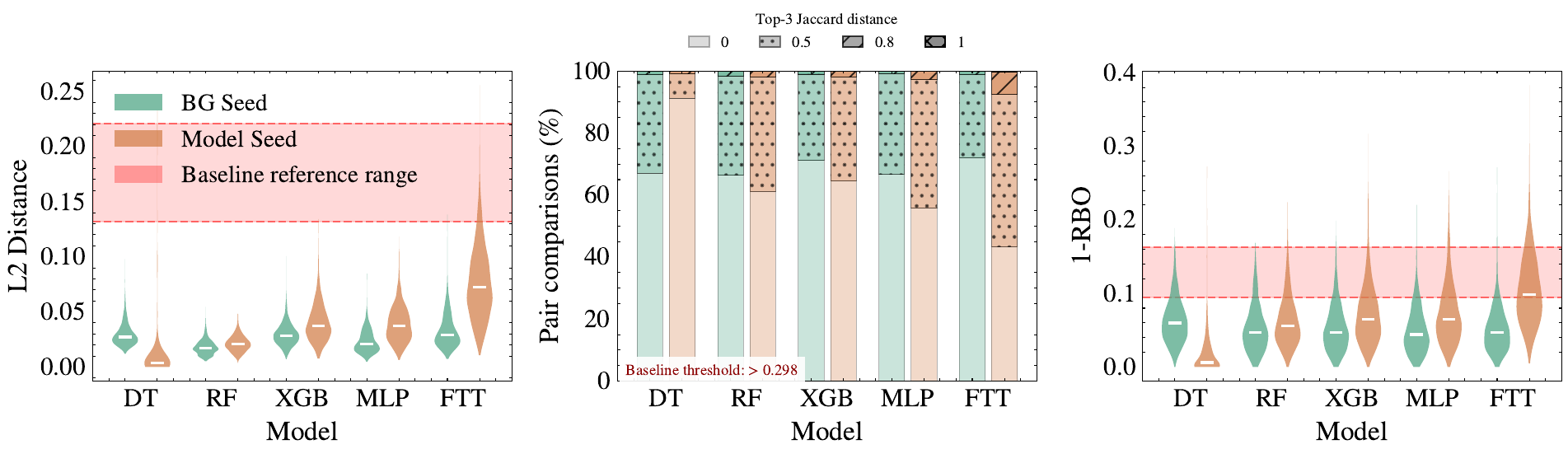}
        \caption{\german}
        \label{fig:german_violin}
    \end{subfigure}
    \hfill
    \begin{subfigure}[b]{0.99\textwidth}
        \centering
        \includegraphics[width=\textwidth]{figures/1b_diabetes_random.pdf}
        \caption{\diabetes}
        \label{fig:diabetes_violin}
    \end{subfigure}
    \hfill
    \begin{subfigure}[b]{0.99\textwidth}
        \centering
        \includegraphics[width=\textwidth]{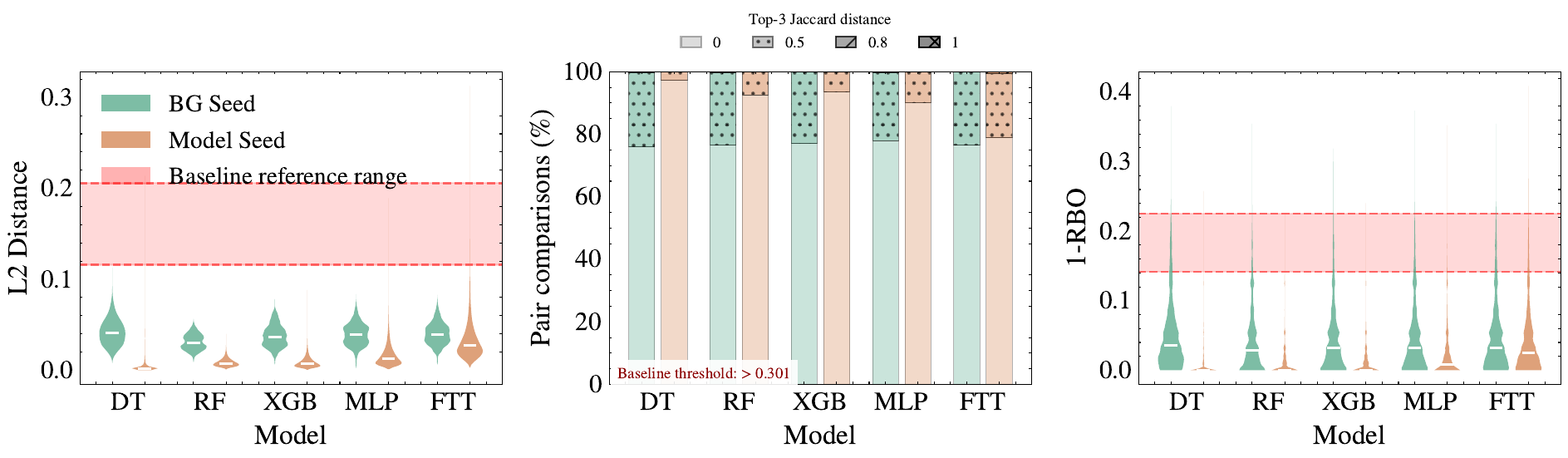}
        \caption{\acs}
        \label{fig:acs_violin}
    \end{subfigure}
    \hfill
    \begin{subfigure}[b]{0.99\textwidth}
        \centering
        \includegraphics[width=\textwidth]{figures/1b_gmsc_random.pdf}
        \caption{\gmsc}
        \label{fig:gmsc_violin}
    \end{subfigure}
    \hfill
    \centering

\caption{
\textbf{Model vs.\ explainer multiplicity (dissection).} Run-pair--instance distance distributions ($10N$ per source) for Kernel SHAP with Random background sampling across four datasets (rows) and three metrics (columns: $\ell_2$, Top-3 Jaccard, and 1-RBO). Jaccard panels use four-value frequency bars.}
    \label{fig:seed_sensitivity}
    \vspace{-0.5cm}
\end{figure}
\paragraph{Extended Dissection of Multiplicity Sources.}
Figure~\ref{fig:seed_sensitivity} extends the source comparison across all four datasets. Model-induced disagreement is generally greater on \german and \diabetes, with model-dependent exceptions. Explainer-induced disagreement is greater on \acs and \gmsc in these comparisons. These distributions do not establish formal variance dominance, training convergence, or a causal effect of dataset size.

\begin{figure}[H]
    \centering
    \begin{subfigure}[b]{0.49\textwidth}
        \centering
        \includegraphics[width=\textwidth]{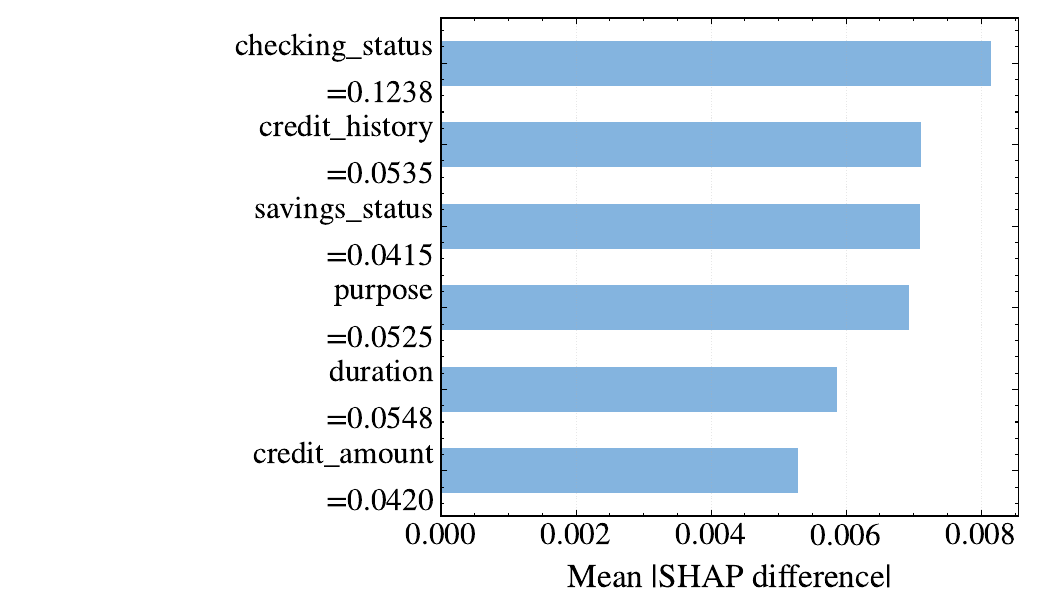}
        \caption{\german}
        \label{fig:german_featurewise}
    \end{subfigure}
    \hfill
    \begin{subfigure}[b]{0.49\textwidth}
        \centering
        \includegraphics[width=\textwidth]{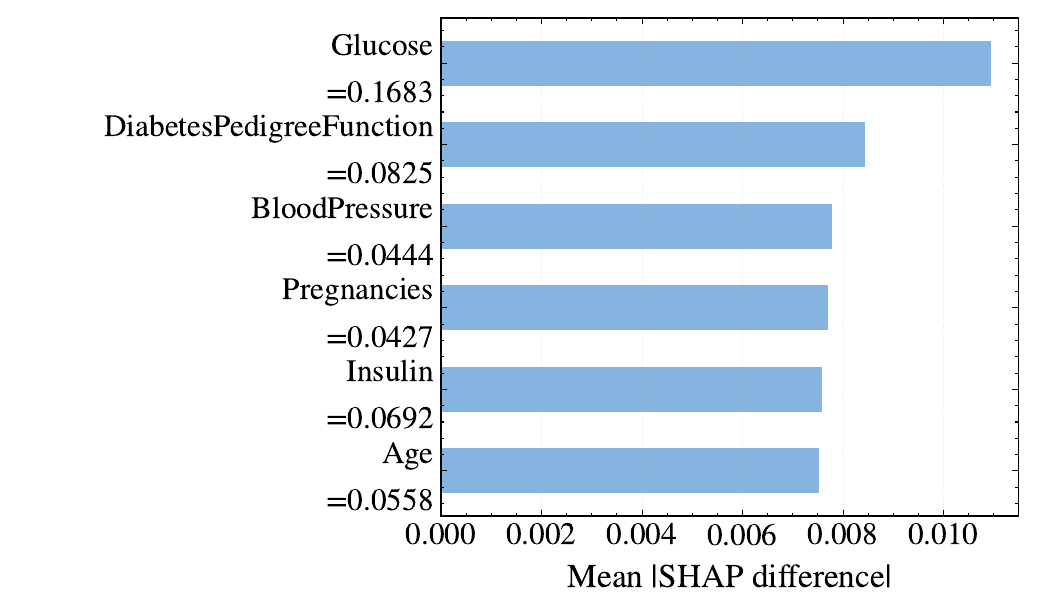}
        \caption{\diabetes}
        \label{fig:diabetes_featurewise}
    \end{subfigure}
    \hfill
    \begin{subfigure}[b]{0.49\textwidth}
        \centering
        \includegraphics[width=\textwidth]{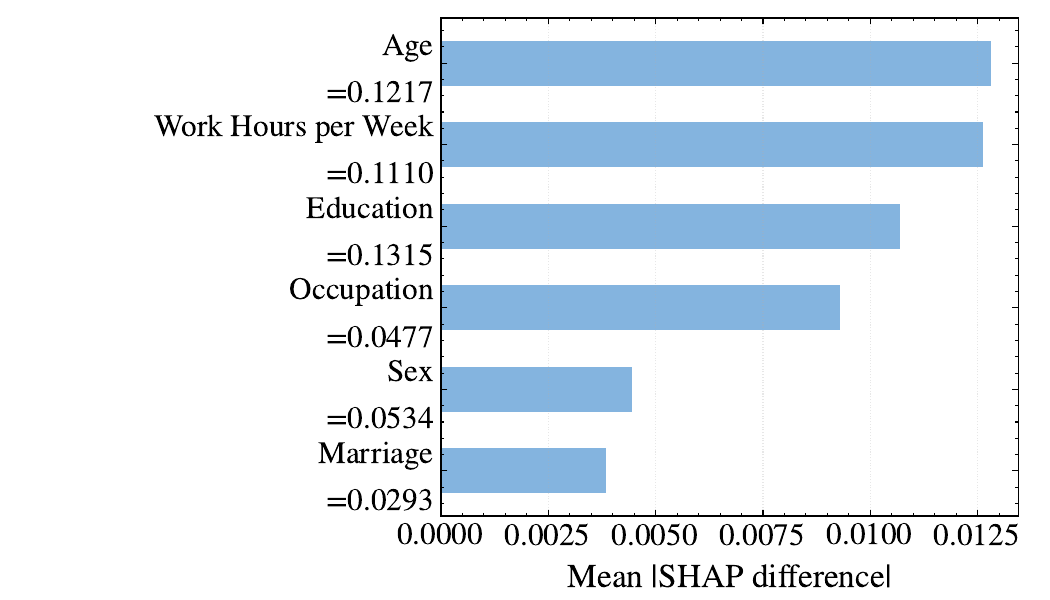}
        \caption{\acs}
        \label{fig:acs_featurewise}
    \end{subfigure}
    \begin{subfigure}[b]{0.49\textwidth}
        \centering
        \includegraphics[width=\textwidth]{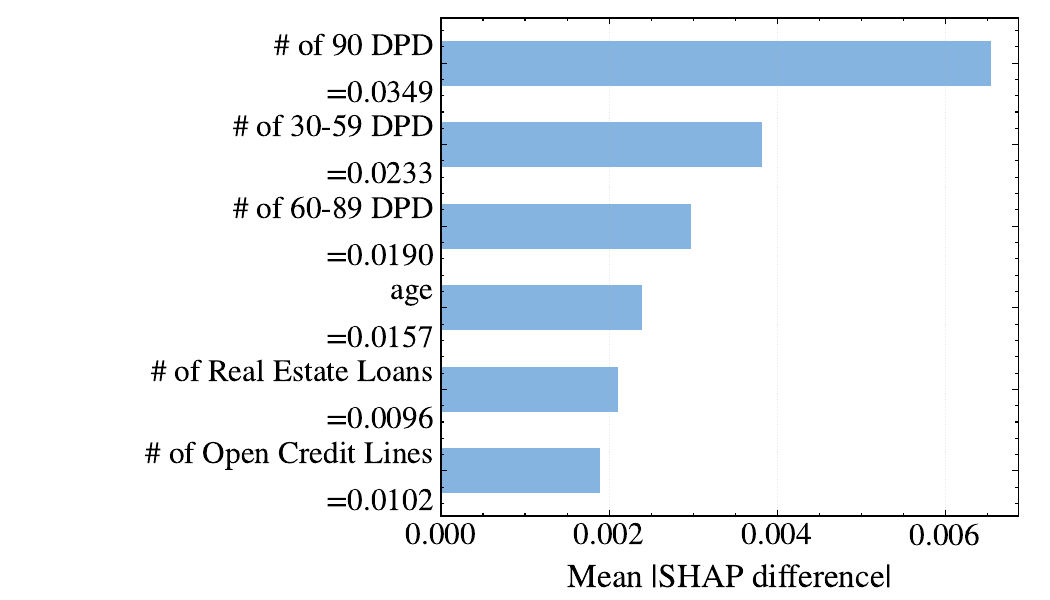}
        \caption{\gmsc}
        \label{fig:gmsc_featurewise}
    \end{subfigure}
    \caption{\textbf{Explainer-seed feature-wise multiplicity for FT-Transformer Model.}
    For Kernel SHAP, the six features with the largest mean absolute pairwise attribution changes are shown, pooling within-method comparisons across Random, K-Means, PredStrat, and CTE.
    Bar length averages these changes over queries and run pairs; labels give mean absolute SHAP values. Features are ordered by sensitivity, not importance.}
    \label{fig:feature_wise}
    \vspace{-0.5cm}
\end{figure}

\paragraph{Feature-wise Sensitivity}
Figure~\ref{fig:feature_wise} shows the six features with the largest mean absolute pairwise attribution changes, pooling within-method comparisons across the four background sampling strategies. Bar length indicates sensitivity, while the labels give mean absolute SHAP values. The small average attribution magnitudes of several displayed features are consistent with rank sensitivity to modest fluctuations. These averages do not directly measure gaps between adjacent feature ranks for individual queries.

Overall, $\ell_2$ distance can understate rank instability because small magnitude differences can change the top-ranked features. In contrast, Jaccard distance surfaces practice-relevant multiplicity by capturing instability in the top-$k$ feature set, whether driven by flat tails or volatile top features. 1-RBO offers a complementary, rank-aware view but is less sensitive to top-$k$ membership changes because it aggregates agreement over ranked prefixes. Since explanations are typically consumed through a small set of top-ranked features, auditing top-$k$ multiplicity is critical.

\paragraph{Kernel SHAP vs Permutation SHAP}
Figure~\ref{fig:kernel_vs_permutation} compares the pooled pair-distance distributions of Kernel SHAP and Permutation SHAP. Similar qualitative patterns do not isolate the respective background and internal sampling contributions.

\begin{figure}[H]
    \centering
    \begin{subfigure}[b]{0.99\textwidth}
        \centering
        \includegraphics[width=\textwidth]{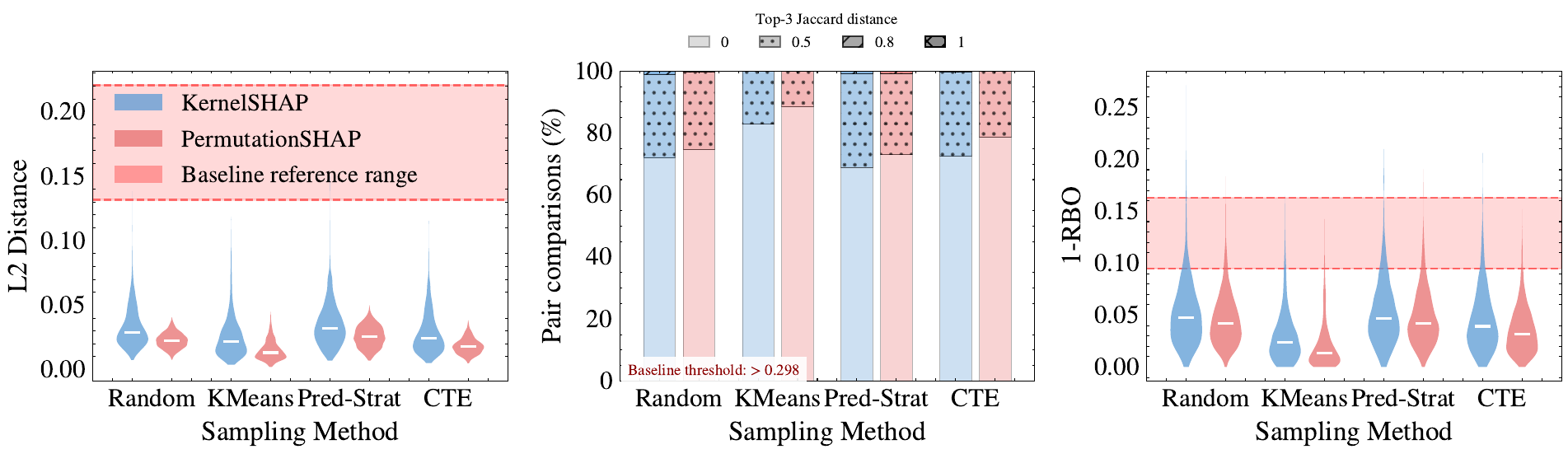}
        \caption{\german}
        \label{fig:german_sampling}
    \end{subfigure}
    \hfill
    \begin{subfigure}[b]{0.99\textwidth}
        \centering
        \includegraphics[width=\textwidth]{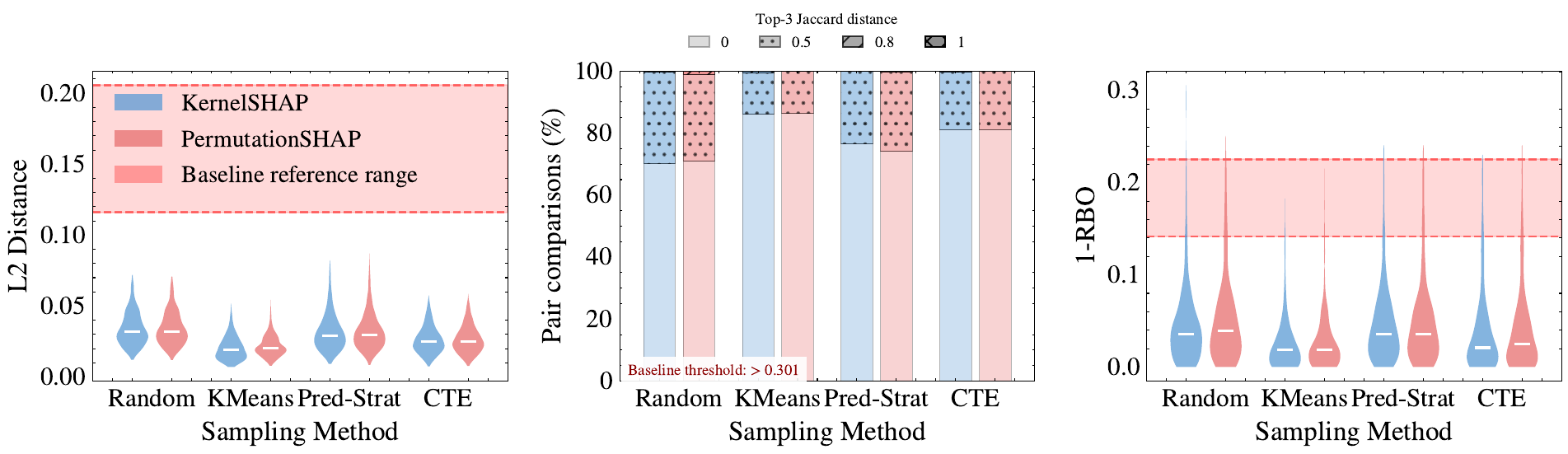}
        \caption{\diabetes}
        \label{fig:diabetes_sampling}
    \end{subfigure}
    \hfill
    \begin{subfigure}[b]{0.99\textwidth}
        \centering
        \includegraphics[width=\textwidth]{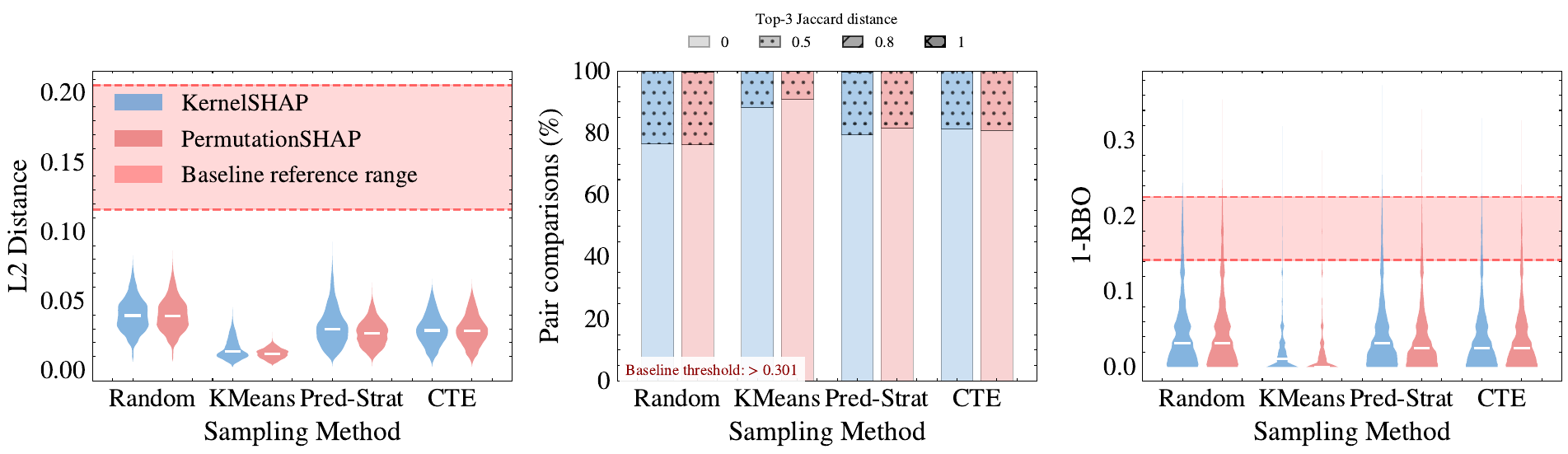}
        \caption{\acs}
        \label{fig:acs_sampling}
    \end{subfigure}
    \begin{subfigure}[b]{0.99\textwidth}
        \centering
        \includegraphics[width=\textwidth]{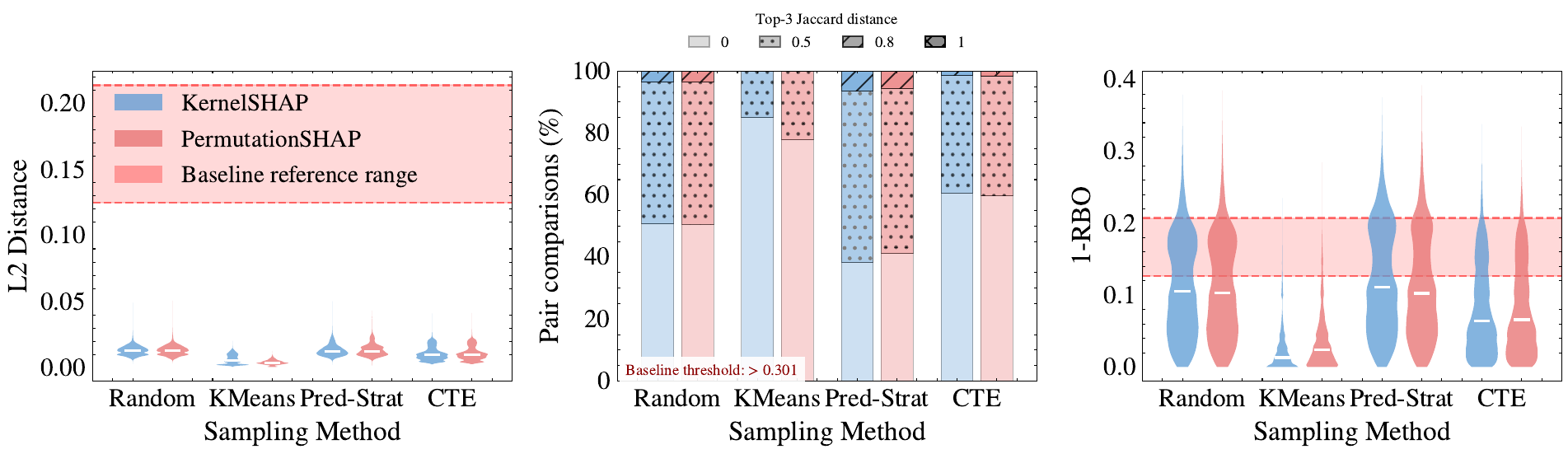}
        \caption{\gmsc}
        \label{fig:gmsc_sampling}
    \end{subfigure}
    \caption{\textbf{Comparing sampling methods.}
    Run-pair--instance distance distributions ($10N$ per configuration), using frequency bars for raw Jaccard, grouped by background sampling strategy (Random, K-Means, Pred-Strat, CTE), shown for Kernel SHAP (blue) and Permutation SHAP (red) across four datasets and three metrics for FT-Transformer. Shaded bands denote randomized baselines. CTE exhibits intra-method sensitivity comparable to Random sampling, while K-Means produces tightly concentrated distributions whose stability does not establish reference agreement (see Figures~\ref{fig:heatmap_true_distance}--\ref{fig:true_violin}).}

    \label{fig:sampling_method}
\end{figure}

\paragraph{Background Sampling Methods and Explainer Multiplicity.}
Figure~\ref{fig:sampling_method} visualizes SHAP multiplicity across diverse sampling strategies. Notably, the state-of-the-art CTE method---designed to optimize distribution compression---exhibits intra-method sensitivity comparable to standard random sampling. K-Means yields tightly concentrated distributions, but reduced variability does not establish agreement with the empirical-distribution reference.

\begin{figure}[H]
    \centering
    \begin{subfigure}[b]{0.49\textwidth}
        \centering
        \includegraphics[width=\textwidth]{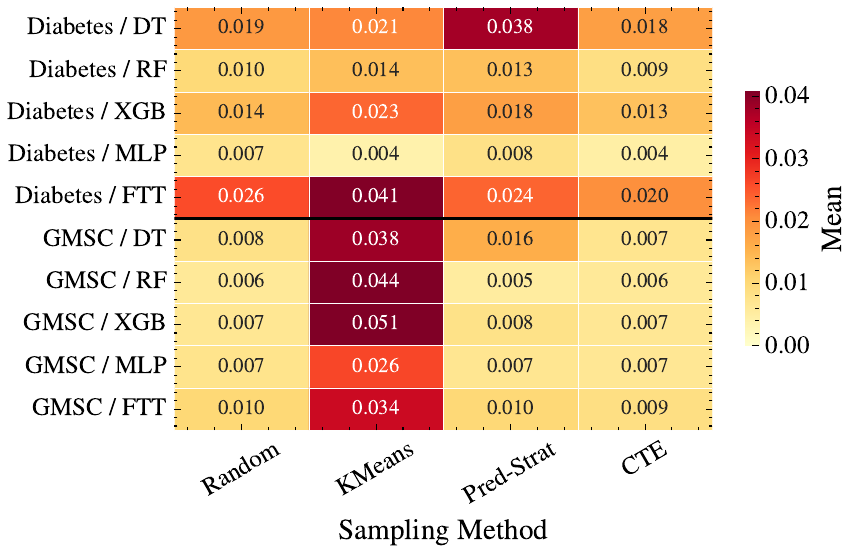}
        \caption{$\ell_2$ Distance}
        \label{fig:heatmap_true_kernel_l2}
    \end{subfigure}
    \hfill
    \begin{subfigure}[b]{0.49\textwidth}
        \centering
        \includegraphics[width=\textwidth]{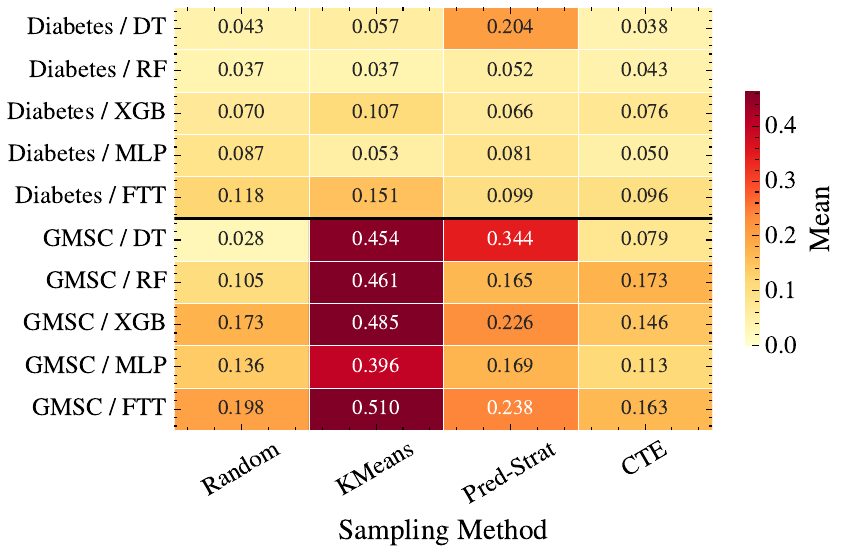}
        \caption{Top-3 Jaccard Distance}
        \label{fig:heatmap_true_kerneln_jaccard}
    \end{subfigure}
    \hfill
    \begin{subfigure}[b]{0.49\textwidth}
        \centering
        \includegraphics[width=\textwidth]{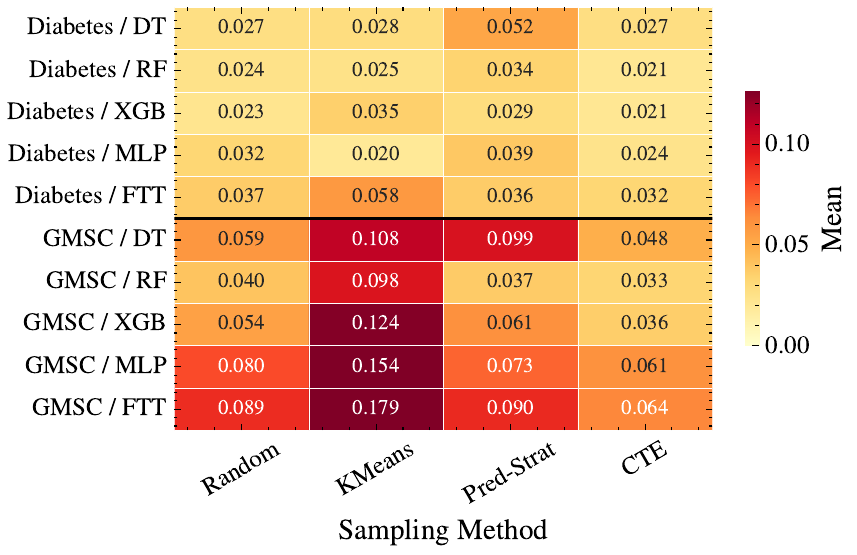}
        \caption{1 - RBO }
        \label{fig:heatmap_true_kernel_rbo}
    \end{subfigure}
    \hfill
    \caption{\textbf{Mean distance to the empirical-distribution reference.} Each cell averages $5N$ run--instance distances to the full-training-background, same-budget reference. With complete equal-sized runs, averaging first within each instance gives the same scalar mean. Panels show (a) $\ell_2$, (b) Top-3 Jaccard, and (c) $1-$RBO. The reported comparisons cover Diabetes and GMSC. These are reference divergences, not universal errors against exact values. Neural comparisons have the limitation in Appendix~\ref{app:reference_caveat}.}
    \label{fig:heatmap_true_distance}
\end{figure}

\begin{figure}[H]
    \centering
    \begin{subfigure}[b]{0.49\textwidth}
        \centering
        \includegraphics[width=\textwidth]{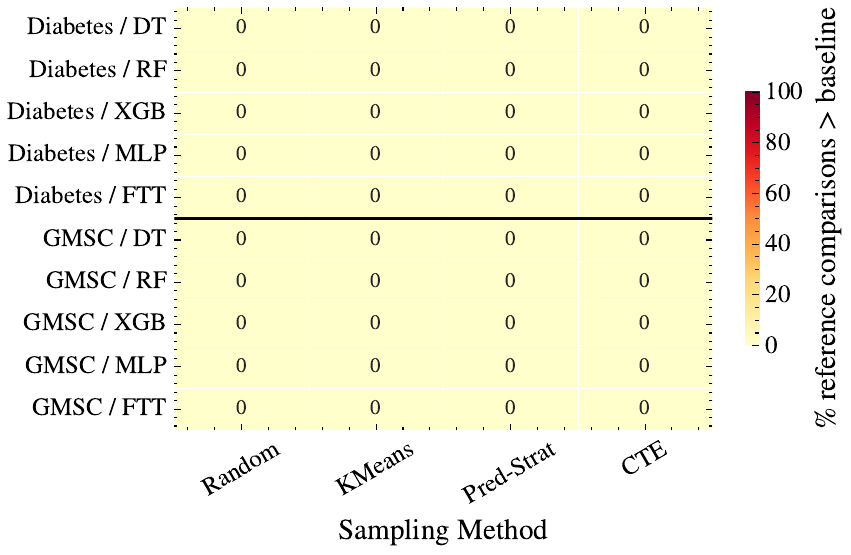}
        \caption{$\ell_2$ Distance}
        \label{fig:heatmap_true_baseline_kernel_l2}
    \end{subfigure}
    \hfill
    \begin{subfigure}[b]{0.49\textwidth}
        \centering
        \includegraphics[width=\textwidth]{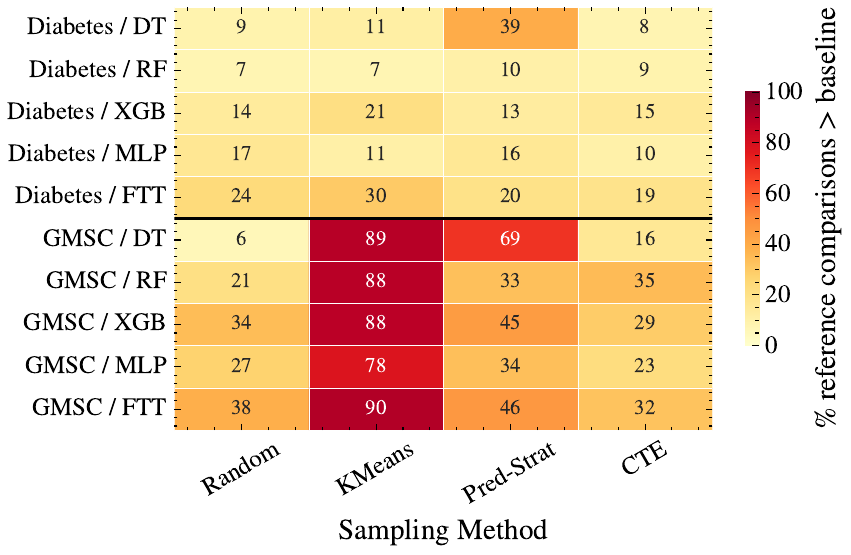}
        \caption{Top-3 Jaccard Distance}
        \label{fig:heatmap_true_baseline_kerneln_jaccard}
    \end{subfigure}
    \hfill
    \begin{subfigure}[b]{0.49\textwidth}
        \centering
        \includegraphics[width=\textwidth]{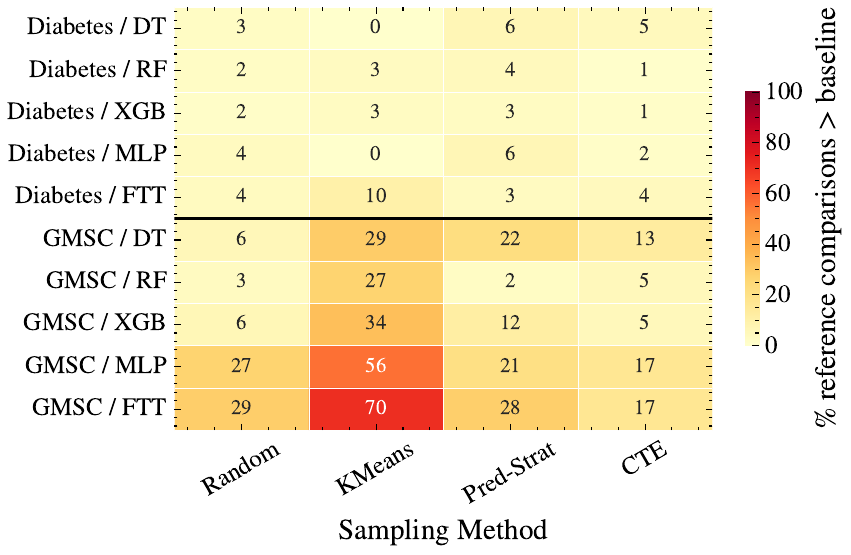}
        \caption{1 - RBO }
        \label{fig:heatmap_true_baseline_kernel_rbo}
    \end{subfigure}
    \hfill
    \caption{\textbf{Baseline exceedance in reference comparisons.} Each cell reports the percentage of $5N$ run--instance distances to the reference strictly above the threshold; no distance averaging precedes thresholding. Panels show (a) $\ell_2$, (b) Top-3 Jaccard, and (c) $1-$RBO, using the same-budget empirical-distribution reference. The reported comparisons cover Diabetes and GMSC; neural comparisons have the limitation in Appendix~\ref{app:reference_caveat}.}
    \label{fig:heatmap_true_baseline}
\end{figure}

\begin{figure}[H]
    \centering

    \hfill
    \begin{subfigure}[b]{0.99\textwidth}
        \centering
        \includegraphics[width=\textwidth]{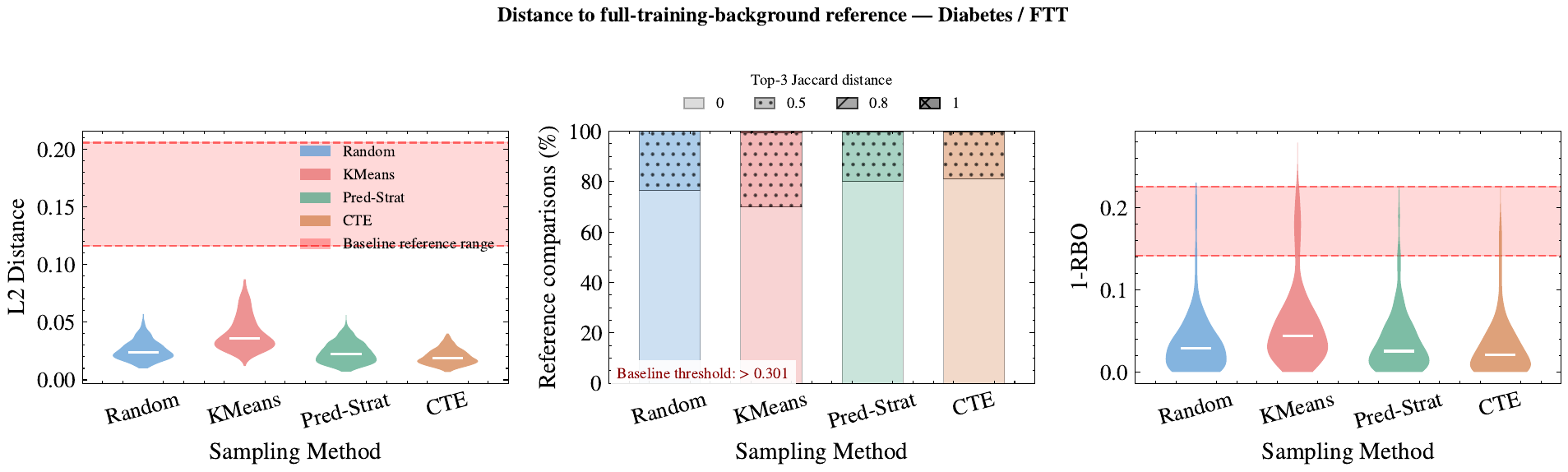}
        \caption{\diabetes}
        \label{fig:diabetes_true_violin}
    \end{subfigure}
    \hfill

    \begin{subfigure}[b]{0.99\textwidth}
        \centering
        \includegraphics[width=\textwidth]{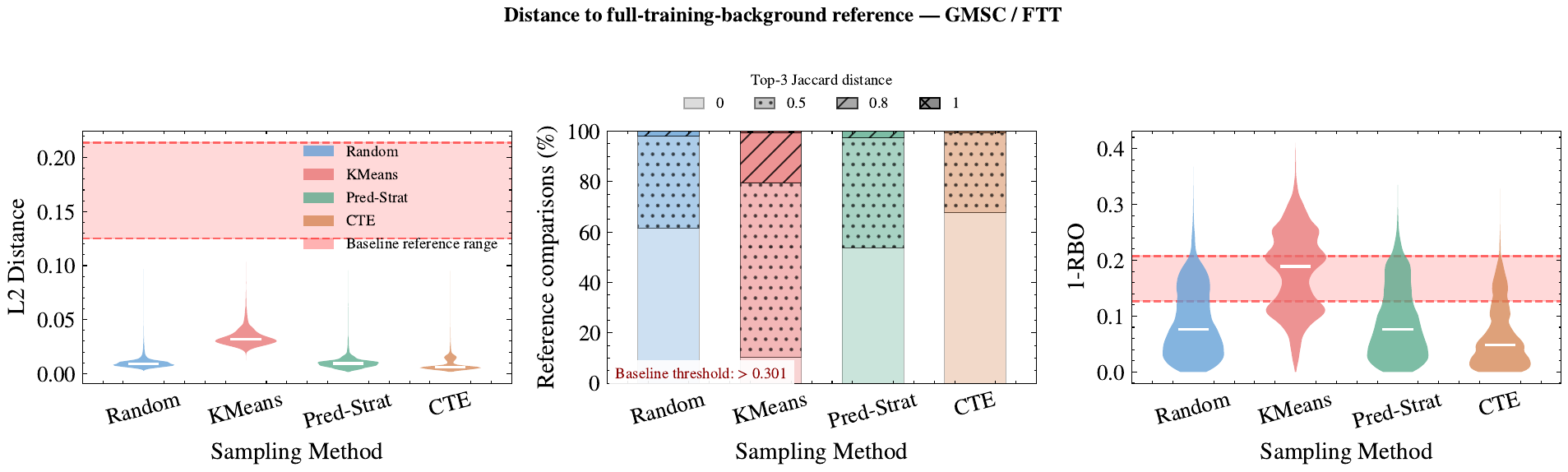}
        \caption{\gmsc}
        \label{fig:gmsc_true_violin}
    \end{subfigure}
    \caption{\textbf{Distribution of reference distances for FT-Transformer.} For Diabetes and GMSC, each distribution pools $5N$ run--instance distances to the same-budget empirical-distribution reference, without averaging within instances. The $\ell_2$/RBO panels use shaded reference ranges; Jaccard shows four-value frequencies. Neural reference comparisons have the limitation in Appendix~\ref{app:reference_caveat}.}
    \label{fig:true_violin}
    \vspace{-0.5cm}
\end{figure}

\paragraph{Stability and reference divergence.}
Figures~\ref{fig:heatmap_true_distance}, \ref{fig:heatmap_true_baseline}, and \ref{fig:true_violin} compare sampled-background explanations with the same-budget empirical-distribution reference. Each test instance contributes five distances, one per background run, giving $5N$ run--instance comparisons to the reference. Figure~\ref{fig:heatmap_true_distance} reports their mean; Figure~\ref{fig:heatmap_true_baseline} reports the percentage strictly above the threshold; Figure~\ref{fig:true_violin} retains their distribution. The reference comparisons cover Diabetes and GMSC. K-Means yields larger reference divergence in many of the retained configurations despite lower run-to-run variation.

\begin{figure}[H]
    \centering
    \begin{subfigure}[b]{0.99\textwidth}
        \centering
        \includegraphics[width=\textwidth]{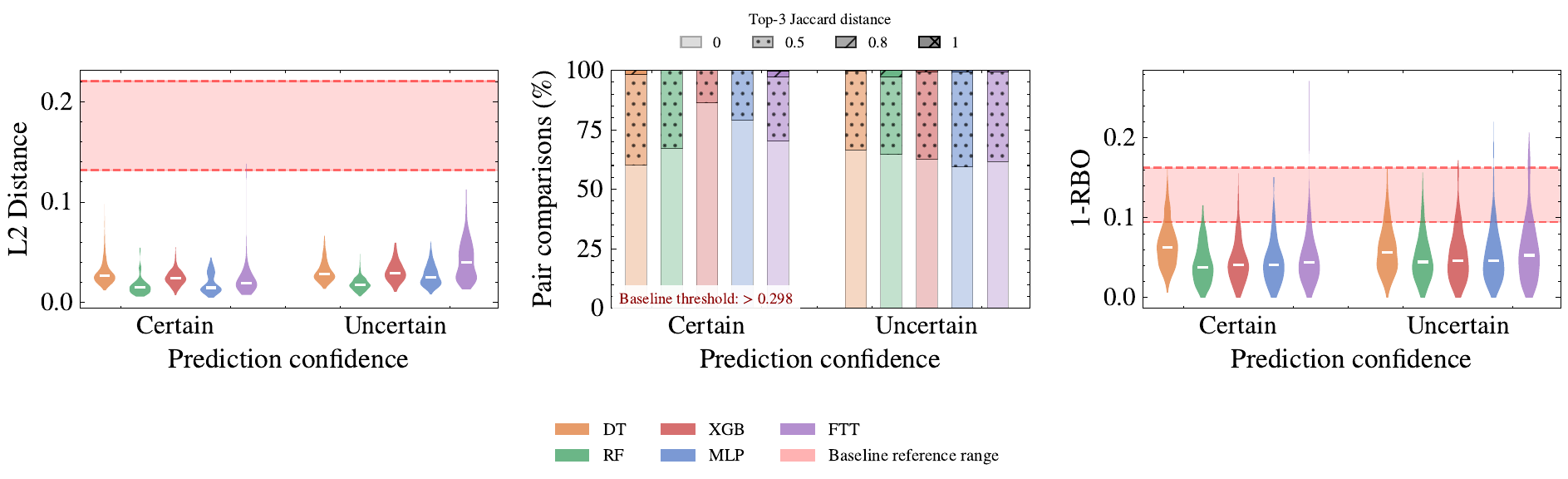}
        \caption{\german}
        \label{fig:german_certainty}
    \end{subfigure}
    \hfill
    \begin{subfigure}[b]{0.99\textwidth}
        \centering
        \includegraphics[width=\textwidth]{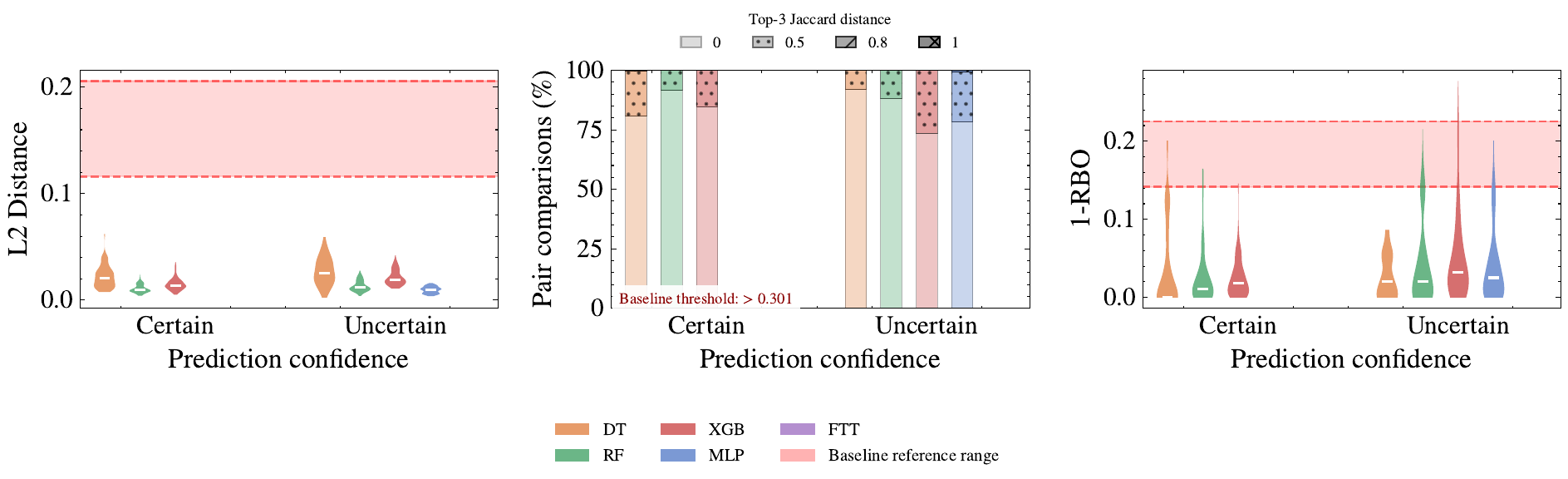}
        \caption{\diabetes}
        \label{fig:diabetes_certainty}
    \end{subfigure}
    \hfill
    \begin{subfigure}[b]{0.99\textwidth}
        \centering
        \includegraphics[width=\textwidth]{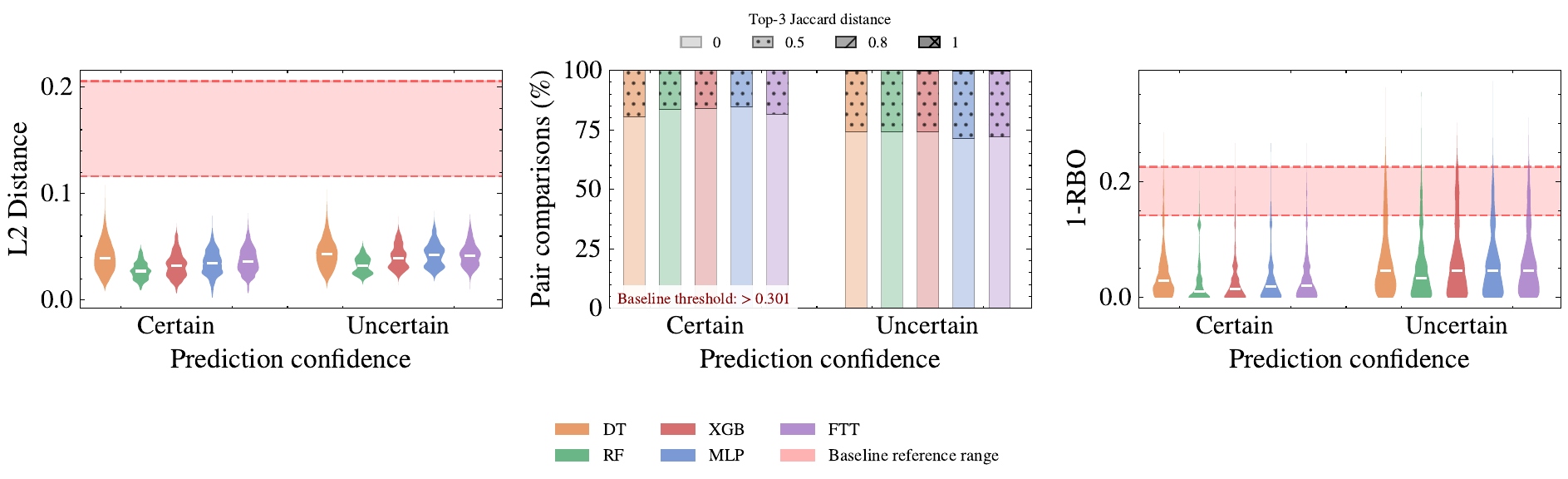}
        \caption{\acs}
        \label{fig:acs_certainty}
    \end{subfigure}
    \begin{subfigure}[b]{0.99\textwidth}
        \centering
        \includegraphics[width=\textwidth]{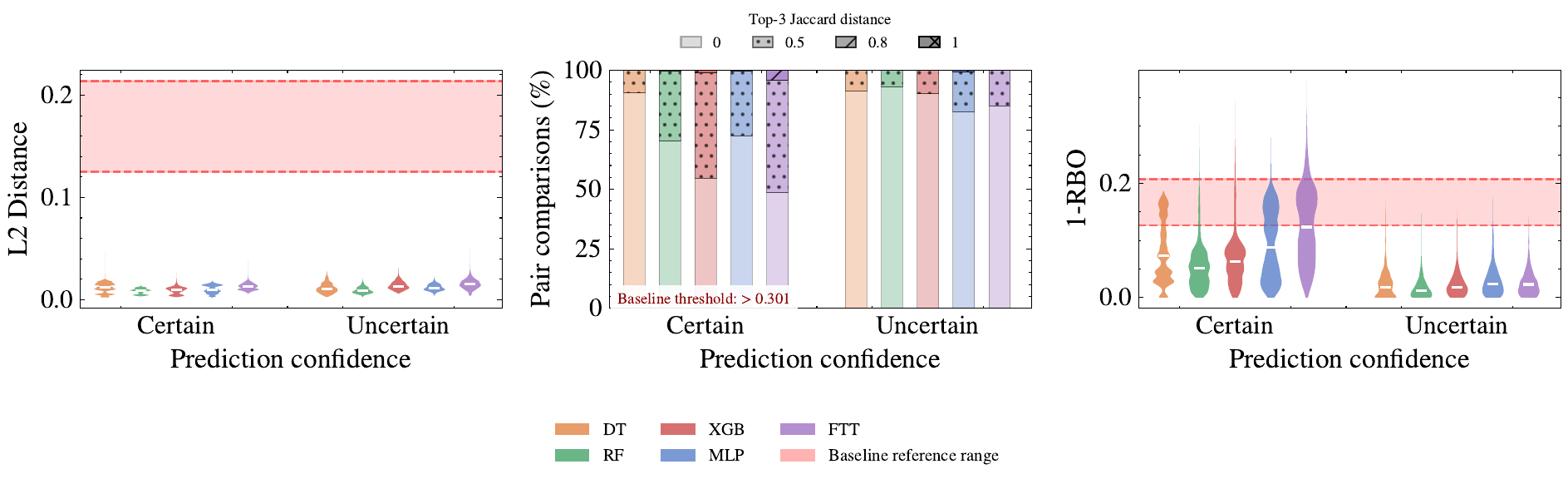}
        \caption{\gmsc}
        \label{fig:gmsc_certainty}
    \end{subfigure}
    \caption{\textbf{Confidence vs.\ stability.} Raw run-pair--instance distances for Kernel SHAP with Random background sampling within \emph{certain} and \emph{uncertain} prediction groups, with the model fixed and explainer seed varied; Top-3 Jaccard uses four-value frequency bars. A group of $N_g$ instances contributes all $10N_g$ distances. The available prediction probabilities cover 19 of 20 dataset--model configurations, excluding Diabetes/FTT; the Diabetes/MLP certain group is empty. Shaded bands denote randomized baselines. High-confidence predictions can still exhibit substantial instability.}
    \label{fig:certainty}
\end{figure}

\paragraph{Explanation Multiplicity Under Certainty Stratification.}
Figure~\ref{fig:certainty} expands on the relationship between model certainty and explanation stability introduced in Section~\ref{subsec:certainty}. We stratify predictions into `Certain' ($P\geq0.9$ or $P\leq0.1$) and `Uncertain' ($0.4\le P \le 0.6$) sets. Each group $g$ pools $10N_g$ run-pair--instance distances, where $N_g$ is its instance count. High-confidence predictions are not immune to explanation multiplicity: Top-3 Jaccard frequencies include membership changes, and RBO distributions can extend beyond the randomized baseline. On \texttt{GMSC}, certain predictions can exhibit comparable or higher rank distances than uncertain ones for several model classes. Flat attribution profiles offer a possible explanation, but this comparison does not identify class imbalance as the cause.

\FloatBarrier